\RequirePackage[2023-11-01]{latexrelease}
\pdfoutput=1

\documentclass[11pt]{article}
\usepackage{arabtex}
\usepackage{utf8}

\usepackage[final]{acl}

\usepackage[utf8]{inputenc}
\usepackage{pifont}

\usepackage{times}
\usepackage{latexsym}
\usepackage{CJKutf8}
\usepackage{CJK}
\usepackage{amssymb}
\usepackage{pifont}
\usepackage{tcolorbox}
\usepackage{colortbl}  
\usepackage{graphicx}  
  \usepackage{arydshln}
\usepackage{tabularx}  
\usepackage{booktabs}  
\usepackage{geometry}  
\usepackage{multirow}
\usepackage{adjustbox}

\usepackage{tabularray}
\usepackage{xcolor}
\usepackage{subcaption}
\usepackage{enumitem}

\usepackage[T1]{fontenc}

\usepackage[utf8]{inputenc}

\usepackage{microtype}

\usepackage{inconsolata}

\usepackage{graphicx}

\usepackage{amsmath}
\usepackage{multirow}
\usepackage{colortbl}
\usepackage{scalerel} 

\usepackage{pdfpages}

\usepackage{ulem} 
\usepackage{tikz}
\usepackage{bbding}
\usepackage{cancel}
\usepackage{ifthen}   
\newcommand{\cmark}{\textcolor{green!70!black}{\ding{51}}}  
\newcommand{\xmark}{\textcolor{red!90!black}{\ding{55}}}    
\newcommand{\hmark}{\includegraphics[width=0.35cm]{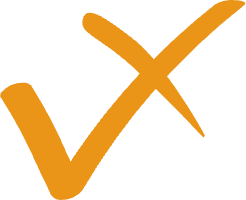}}

\usepackage[textsize=scriptsize]{todonotes}

\title{\includegraphics[width=0.75cm]{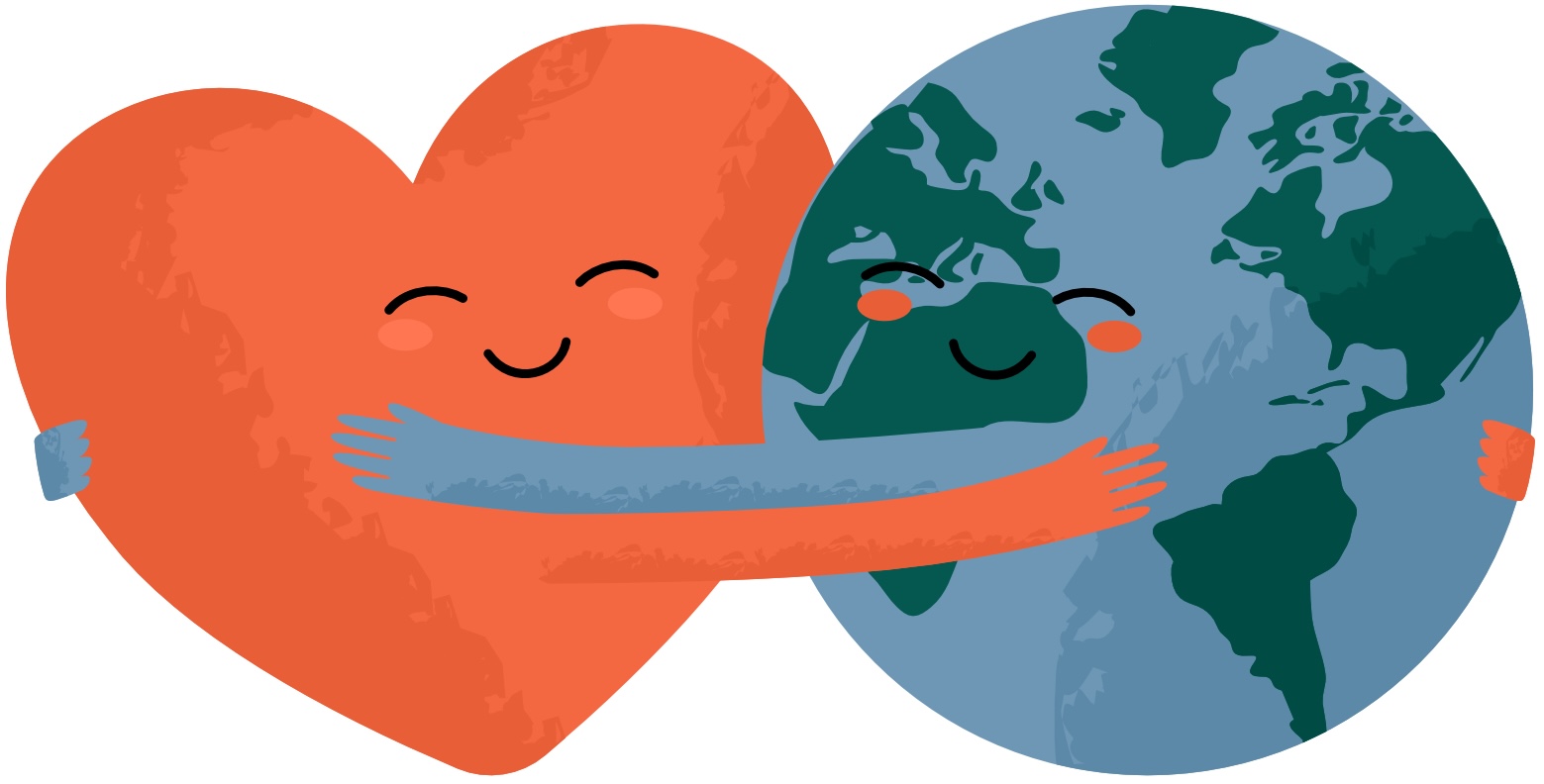}CARE: Multilingual Human Preference Learning for \\ Cultural Awareness}

\author{
 \textbf{Geyang Guo\textsuperscript{$\alpha$}},
 \textbf{Tarek Naous\textsuperscript{$\alpha$}},
 \textbf{Hiromi Wakaki\textsuperscript{$\beta$}},
 \textbf{Yukiko Nishimura\textsuperscript{$\beta$}},
\\
 \textbf{Yuki Mitsufuji\textsuperscript{$\beta,\gamma$}},
 \textbf{Alan Ritter\textsuperscript{$\alpha$}},
 \textbf{Wei Xu\textsuperscript{$\alpha$}}
\\
 \textsuperscript{$\alpha$}Georgia Institute of Technology,
 \textsuperscript{$\beta$}Sony Group Corporation,
 \textsuperscript{$\gamma$}Sony AI
 \\
 \texttt{\{guogeyang, tareknaous\}@gatech.edu} \\
 \texttt{\{hiromi.wakaki, yukiko.b.nishimura, yuhki.mitsufuji\}@sony.com} \\
 \texttt{\{alan.ritter, wei.xu\}@cc.gatech.edu}
}

\definecolor{qwen}{HTML}{44459e}
\definecolor{llama}{HTML}{0064e0}
\definecolor{mistral}{HTML}{f2a63b}
\definecolor{deepseek}{HTML}{4a67f6}

\definecolor{green1}{HTML}{d3e29d}
\definecolor{green2}{HTML}{acc864}
\definecolor{green3}{HTML}{8ab446}
\definecolor{green4}{HTML}{2d6b22}
\definecolor{myred}{HTML}{EFB9AD}

\newcolumntype{Y}{>{\centering\arraybackslash}X}

\begin{document}
\maketitle

\begin{abstract}

Language Models (LMs) are typically tuned with human preferences to produce helpful responses, but the impact of preference tuning on the ability to handle culturally diverse queries remains understudied. In this paper, we systematically analyze how native human cultural preferences can be incorporated into the preference learning process to train more culturally aware LMs. We introduce \textbf{CARE}, a multilingual resource containing 3,490 culturally specific questions and 31.7k responses with human judgments. We demonstrate how a modest amount of high-quality native preferences improves cultural awareness across various LMs, outperforming larger generic preference data. Our analyses reveal that models with stronger initial cultural performance benefit more from alignment, leading to gaps among models developed in different regions with varying access to culturally relevant data.  CARE is publicly available at \url{https://github.com/Guochry/CARE}.

\end{abstract}

\begin{figure}[t]
    \centering
    \includegraphics[width=0.90\linewidth]{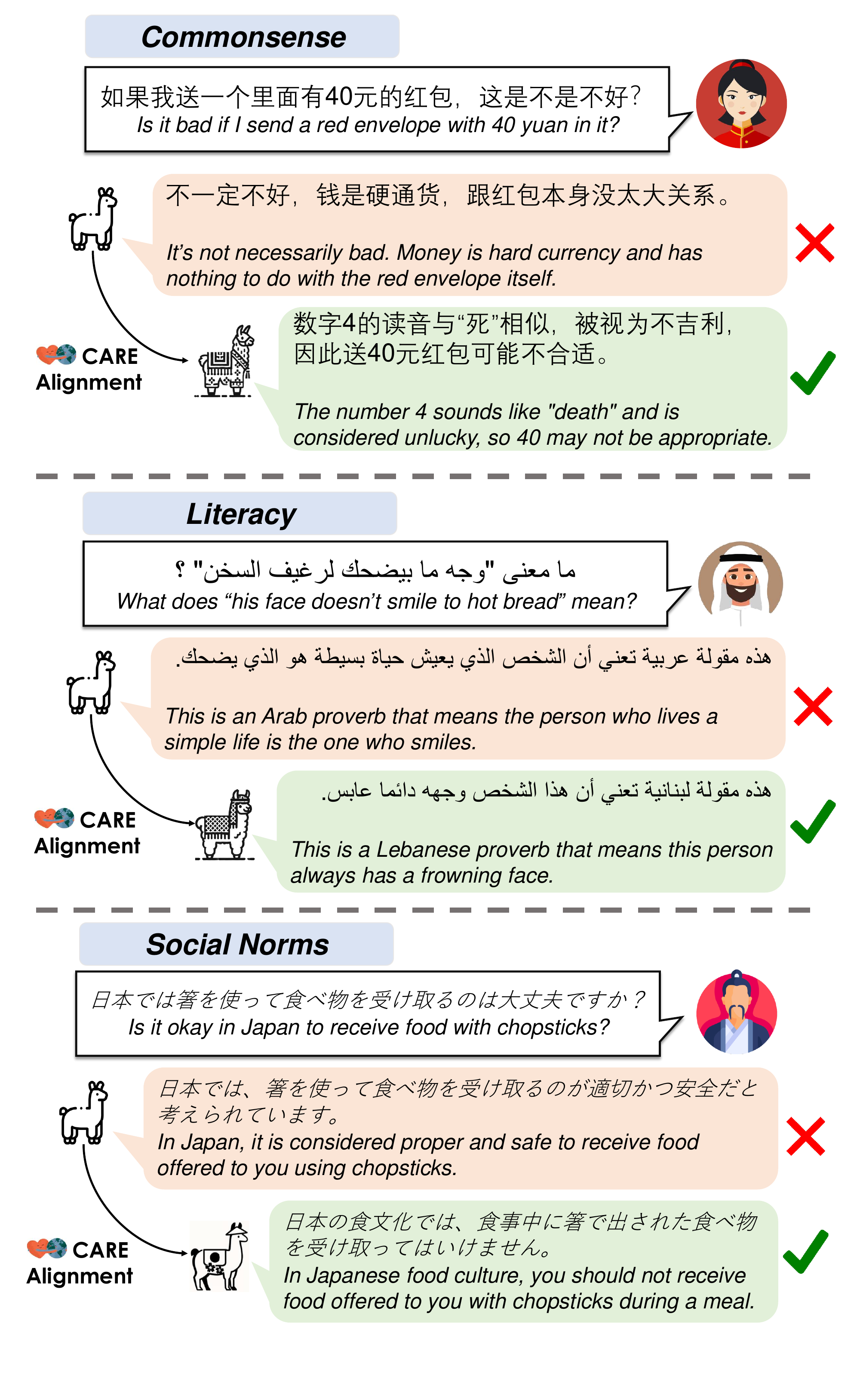}
    \caption{Example LM responses to culture-specific questions in the native languages. The base LM (\includegraphics[height=1em]{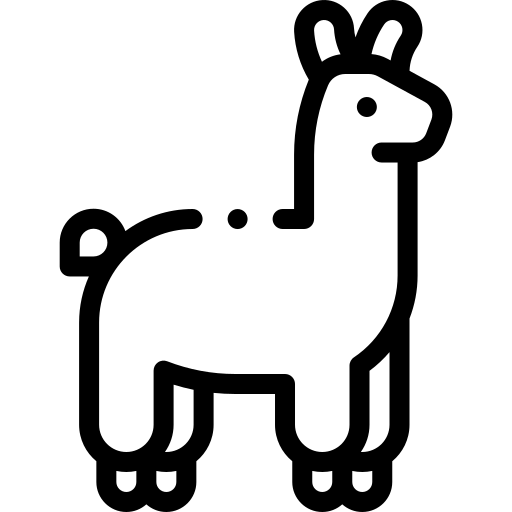}) \texttt{Llama3.1-8B} fails to respond appropriately, while its aligned versions on CARE generate better responses for Chinese (\includegraphics[height=1em]{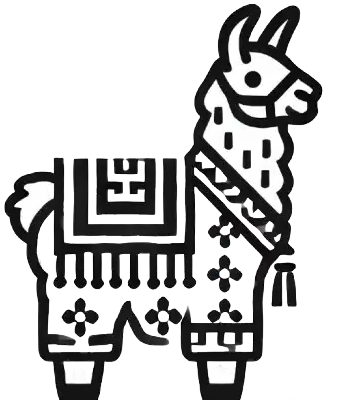}), Arab (\includegraphics[height=1em]{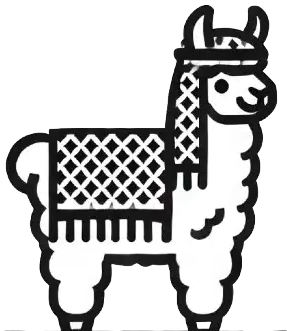}), and Japanese (\includegraphics[height=1em]{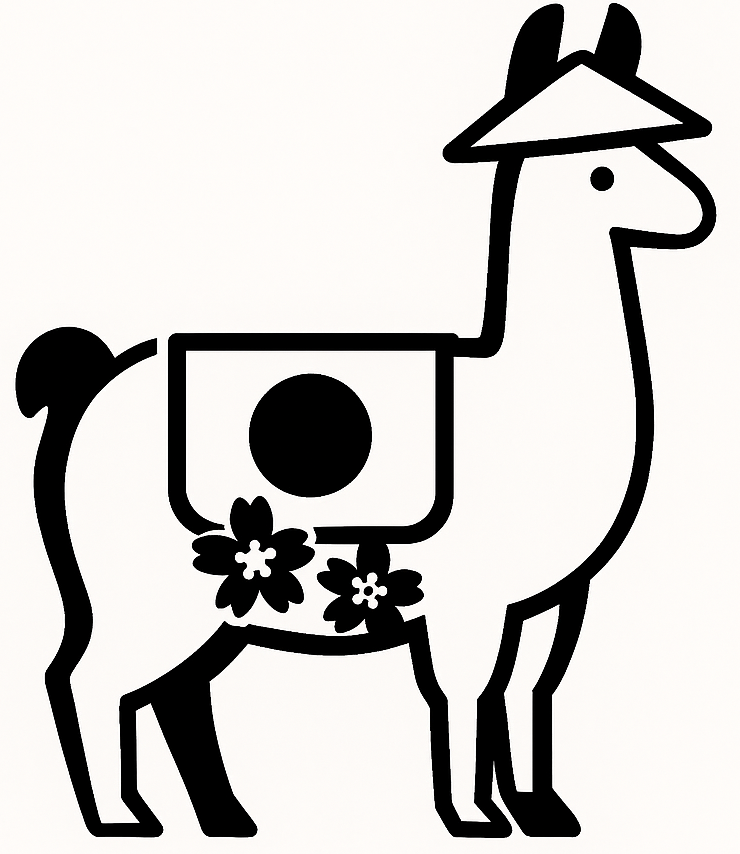})  cultures.}
    \label{fig:figure-intro}
    \vspace{-1em}
\end{figure}

\section{Introduction}
\label{sec:intro}




After large-scale pre-training, Language Models (LMs) typically undergo a crucial post-training phase (aka ``alignment''), which includes supervised fine-tuning and preference tuning, to improve their ability to follow instructions and alignment with human preferences~\cite {ouyang2022training, rafailov2024direct}. 
However, \textit{the effects of alignment on the cultural awareness} of multilingual LMs (i.e., how well they understand and generate appropriate responses to diverse, culturally relevant queries) remains understudied, partly because frontier open-weight models with multilingual support (e.g., Llama-3, Qwen-2.5) have only become available very recently around mid-2024. 
Consequently, existing studies have either examined off-the-shelf aligned LMs (e.g., Aya, Tulu 2)~\cite{naous-etal-2024-beer,ryan2024unintended} or explored instruction fine-tuning with culturally relevant data~\cite{li2024culturellm}.

\begin{table*}[t]
\centering
\resizebox{\linewidth}{!}{%
\begin{tabular}{lccccccc}
\toprule
\multirow{2}{*}{\textbf{Dataset}} & \textbf{Culture-} & \multirow{2}{*}{\textbf{Topics}} & \textbf{Manually-} & \textbf{Human}\footnotemark & \multirow{2}{*}{\textbf{Languages}} & \multirow{2}{*}{\textbf{Format}}  \\
 & \textbf{Specific} &   & \textbf{Curated} & \textbf{Preferences} &   \\
\midrule
\textit{Cultural Evaluation} \\
\hspace{1em}{Fork~(\citeauthor{palta2023fork})} & \cmark & food-related customs & \cmark & \xmark & en & multiple-choice    \\
\hspace{1em}{CulturalBench~(\citeauthor{chiu2024culturalbench})} & \cmark & daily life, social etiquette, wider society & \xmark & \xmark & en & multiple-choice    \\
\hspace{1em}{GeoMLAMA~(\citeauthor{yin2022geomlama})} & \cmark & cultural commonsense & \cmark & \xmark & en, fa, hi, sw, zh & multiple-choice    \\
\hspace{1em}{BLEnD~(\citeauthor{myung2024blend})} & \cmark & food, sports, family, education, holidays, work-life & \cmark & \xmark & 13 languages & multiple-choice, free text   \\
\hspace{1em}{Include~(\citeauthor{romanou2025include})} & \cmark & general knowledge, social science, professional certifications, etc. & \cmark & \xmark & 44 languages & multiple-choice   \\
\hspace{1em}{CAMeL~(\citeauthor{naous2023having})} & \cmark & beverage, clothing, food, location, religion, sports, etc. & \cmark & \xmark & ar & masked prompts   \\
\hspace{1em}{CANDLE~(\citeauthor{nguyen2023extracting})} & \cmark & geography, religion, occupation, food, clothing, etc. & \xmark & \xmark & en & assertions   \\
\hspace{1em}{LLM-GLOBE~(\citeauthor{karinshak2024llm})} & \cmark & cultural values & \cmark & \xmark & zh, en & multiple-choice, free text   \\
\hspace{1em}{CulturalTeaming~(\citeauthor{chiu2024culturalteaming})} & \cmark & red-teaming cultural questions & \cmark & \xmark & en & multiple-choice    \\
\hspace{1em}{SHADES~(\citeauthor{mitchell-etal-2025-shades})} & \cmark & culture-specific stereotypes & \cmark & \xmark & 16 languages & stereotyped statements   \\
\hspace{1em}{JMMMU~(\citeauthor{onohara2024jmmmu})} & \cmark & art, heritage, history, business, science, medicine, etc. & \cmark & \xmark & ja & multiple-choice \\
\hdashline[1pt/1pt]
\textit{Cultural Fine-tuning} \\
\hspace{1em}{Aya~(\citeauthor{singh2024aya})} & \hmark & news, stories, recipes, scientific texts, etc. & \cmark & \xmark & 65 languages & free text    \\
\hspace{1em}{CIDAR~(\citeauthor{alyafeai2024cidar})} & \hmark & technology, translation, poetry, grammar, etc.  & \xmark & \xmark & ar & free text    \\
\hspace{1em}Cameleval~(\citeauthor{qian2024cameleval}) & \xmark & information provision, reasoning, creative writing, etc.  & \cmark & \cmark & ar & free text    \\
\hspace{1em}{PRISM~(\citeauthor{kirk2024prism})} & \cmark & cross-cultural controversies & \cmark & \cmark & en & multi-turn conversations    \\
\hspace{1em}Palm~(\citeauthor{alwajih2025palm}) & \cmark & history, celebrations, sports, literature, etc. & \cmark & \xmark & ar & free text    \\
\hspace{1em}{CULTUREINSTRUCT~(\citeauthor{pham-etal-2025-cultureinstruct})} & \cmark & art, cuisine, cultural norms, festivals, history, etc. & \xmark & \xmark & en & free text    \\
\hspace{1em}{CultureBank~(\citeauthor{shi2024culturebank})} & \cmark & social norms, food, communication, festivals, etc. & \xmark & \xmark & en & free text    \\
\hspace{1em}{CulturePark~(\citeauthor{li2024culturepark})} & \cmark & human belief, norm, custom & \xmark & \xmark & en & free text    \\
\hspace{1em}{CultureLLM~(\citeauthor{li2024culturellm})} & \cmark & value survey & \xmark & \xmark & 9 languages & multiple-choice    \\
\midrule
\textbf{CARE (Ours) \includegraphics[width=0.5cm]{fig/Hugging.jpg}} & \cmark & cultural entities, opinion, norms, commonsense, literacy & \cmark & \cmark & ar, ja, zh & free text    \\
\bottomrule
\end{tabular}%
}
\caption{Comparison of datasets for studying LMs' cultural awareness. CARE is a multilingual, \textit{human-annotated preference} dataset specifically grounded in culture. 
(\hmark) indicates resources that include some cultural considerations (in the form of culture-related questions or by recruiting native annotators), while cultural coverage is not their only or primary focus (see \S~\ref{subsec:existing_data}).
Representative examples from each dataset are provided in Appendix~\ref{appendix:data-comparison-example}. }
\label{tab:related-work-comparison}
\vspace{-1.0em}
\end{table*}

In this paper, we systematically analyze the extent to which and under what conditions preference optimization (e.g., DPO, KTO, SimPO) can help LMs align with regional user expectations along five key dimensions: cultural entities, cultural commonsense, social norms, opinions, and literacy (see examples in Figure~\ref{fig:figure-intro}). To enable our study, we create \includegraphics[width=0.5cm]{fig/Hugging.jpg} CARE, \textit{a multilingual culture-specific dataset with human preference judgments} on 31.7k LLM/Human-written responses to 3,490 questions about Chinese, Arab, and Japanese cultures. Questions are drawn from a variety of instruction tuning datasets \cite{singh2024aya, alyafeai2024cidar} and evaluation resources \cite{chiu2024culturalbench, palta2023fork} that we manually identified as culturally-relevant. We further collected additional questions to improve the coverage on regionally popular but less digitally documented commonsense knowledge and social norms, with help from native speakers who have overseas experience.

Leveraging CARE, we show that a \textit{modest, high‑quality set of native preferences} can enhance LMs' cultural awareness across model families and sizes, achieving competitive or stronger performance than generic preference data despite using 3–17$\times$ less data, and generalizing to out‑of‑domain cultural tasks~\cite{myung2024blend, romanou2025include}. 
We then conduct controlled studies on various LMs and uncover several important insights for cultural alignment: \textbf{(1)} Base model strength matters: aligning base models with stronger initial cultural awareness such as Qwen2.5‑7B and Gemma2‑9B lifts scores across all cultural categories, whereas models with weaker initial performance hardly improve after alignment, consistent with the post-training analyses of \citet{ivison2023camels}. \textbf{(2)} Data efficiency scales: even 25\% of CARE can lead to a 74\% improvement, with performance continuing to rise as the dataset scales to full size. \textbf{(3)} Cross‑culture synergy exists: mixing samples about local and foreign cultures in preference learning improves overall awareness of local cultures (e.g., Chinese 4.691 $\rightarrow$ 4.944, Arabic 2.980 $\rightarrow$ 3.538). 

\footnotetext{By ``Human Preferences'', we mean pairwise or listwise rankings of multiple candidate answers to the same question, which are the signal needed for preference tuning.}

Beyond these insights, we observe that performance on different cultural dimensions varies across model families. For example, the Chinese-centric Qwen2.5‑72B~\cite{yang2024qwen2, wen2024chinese} excels on Chinese entities and social norms, however, it lags behind GPT-4o on Chinese cultural commonsense. In general, LLMs consistently perform better when queried about culture-specific phenomena in the native language (i.e., Arabic, Chinese, Japanese) rather than English, with the gap narrowing for cultural commonsense. A likely explanation is that everyday cultural commonsense is often unspoken in native contexts while more explicitly expressed and asked about in foreign languages~\cite{yin2022geomlama}. Our analysis with search engines further supports this, highlighting the importance of broad cultural and linguistic coverage in training data.

\section{Related Work}
\label{sec:related-work}


\paragraph{Cultural Evaluation. }
The widespread use of LMs has sparked research interest in their relevance to diverse cultures \cite{adilazuarda2024towards, shen2024understanding, pawar2024survey, liu2024culturally}. Several studies investigate LMs' alignment to different cultures by examining their responses to social surveys that reflect human values and attitudes \cite{wvs, cao2023assessing}. It has been consistently shown that LMs favor answers associated with Western culture \cite{alkhamissi2024investigating, abdulhai2023moral}, even when prompted in different languages \cite{masoud2023cultural, wang2023not, rystrom2025multilingual} or after preference optimization \cite{ryan2024unintended}. Another line of work develops culture-specific evaluation resources such as knowledge bases of cultural facts \cite{keleg2023dlama, yin2022geomlama, zhou2024does}, entity-centric cultural benchmarks \cite{naous-etal-2024-beer}, and user self-reported cultural experiences \cite{shi2024culturebank}. Other works have constructed QA datasets for different cultural aspects, such as culinary customs \cite{palta2023fork}, norms \cite{rao2024normad,zhan2024renovi}, social etiquette \cite{chiu2024culturalteaming, chiu2024culturalbench, qiu2025multimodal}, and more \cite{arora2024calmqa, mousi2024aradice}. However, they are primarily designed for evaluation, often in a multiple-choice QA format, and are not well-suited for aligning LMs through preference optimization, which ideally relies on free-text QA and human preference data. (see Table~\ref{tab:related-work-comparison} and \S \ref{subsec:existing_data}).


\paragraph{Cultural Fine-tuning. }

Several resources~\cite{muennighoff2022crosslingual, singh2024aya, alyafeai2024cidar, OpenOrca, notus2023, ahmadian2024multilingual} have been developed to improve LMs' multilingual performance. While they include culturally relevant samples, they primarily offer general instructions and preferences for safety and helpfulness. A few other works have investigated cultural adaptation of LMs via fine-tuning strategies \cite{li2024culturellm, li2024culturepark, shi2024culturebank, kirk2024prism, choenni-etal-2024-echoes, yuan2024cultural, pham-etal-2025-cultureinstruct, alwajih2025palm, xu-etal-2025-self, yao2025caredio}. 
Unlike past studies (see Table~\ref{tab:related-work-comparison}), CARE provides human preferences on culture-specific topics in three languages (not translated from English), enabling a direct study of how multilingual preference optimization improves models' cultural awareness and by how much.

\paragraph{Multilingual Preference Optimization.}

Existing work aims to improve the alignment of LMs with human preferences across different languages, either by synthesising preference data~\cite{chinese-orca-dpo-pairs,japanese-orca-dpo-pairs,arabic-dpo-pairs} or developing translation-based methods~\cite{she2024mapo, yang2024language}. Only a few studies have released native multilingual human preference datasets that are neither translated from English nor rated by AI models. Notable examples include OpenAssistant~\cite{kopf2023openassistant} and HelpSteer3-Preference~\cite{wang2025helpsteer3preferenceopenhumanannotatedpreference}, both of which focus on general-purpose interactions. We focus on culturally-relevant preferences for an in-depth analysis of multilingual multicultural alignment.

\begin{figure}[b!]
    \centering
    \includegraphics[width=\linewidth]{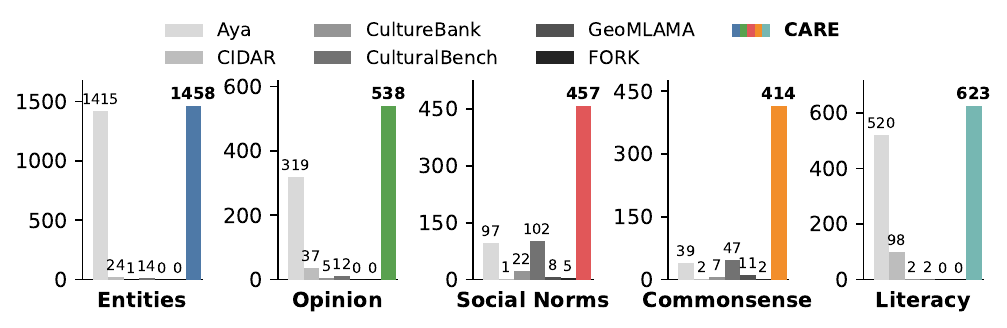}
    \caption{Overall coverage per cultural category. CARE provides 16.6$\times$ more social norm and commonsense questions compared to 2 instruction tuning datasets (Aya and CIDAR) and 4 cultural knowledge bases (CultureBank, CultureBench, GeoMLAMA, FORK) combined.}
    \label{fig:care-statistics}
    \vspace{-1em}
\end{figure}

\section{Constructing CARE \includegraphics[width=0.8cm]{fig/Hugging.jpg}}
\label{sec:care}

We introduce CARE, a multilingual human preference dataset that consists of 3,490 culture-specific questions and 31.7k human/LLM responses \textit{with} human preference ratings.

\subsection{Limitations of Existing Data Resources}
\label{subsec:existing_data}

In this section, we notice limitations of existing resources and take action to fix them. 
We start with the Aya dataset~\cite{singh2024aya}, the largest multilingual instruction-tuning resource that contains human-written questions and answers. 
Though its samples are collected from native speakers, only part of its content focuses on culture, as many examples consist of general questions. After manual inspection and filtering out generic questions (e.g. ``\textit{How many hearts does an octopus have?}''), we yield 1,324 (out of 4,909) and 600 (out of 3,264) question and answer pairs that are culturally relevant in Chinese and Japanese but only 457 (out of 14,210) samples in Arabic.  To expand the Arabic set, we apply the same filtering process to around 2,000 samples from CIDAR~\cite{alyafeai2024cidar}, a human-written Arabic instruction dataset, resulting in 162 relevant samples.


We also examine four existing cultural knowledge bases, namely, CultureBank \cite{shi2024culturebank}, CulturalBench \cite{chiu2024culturalbench}, GeoMLAMA \cite{yin2022geomlama}, and FORK \cite{palta2023fork}.  Since these datasets are exclusively in English, we manually translate them into the corresponding native language.  
We observe that these knowledge bases focus on broad regional coverage, but supply limited samples per culture: namely, 60, 61, and 50 samples for Chinese, Arab, and Japanese cultures. Since these resources rely on multiple-choice or text-infilling formats, we reformatted all answers into free-text responses, adding explanations and removing any instances of stereotyping or overgeneralization. 

We manually classify all questions into one of five categories: \textbf{(1)} \textbf{Cultural Entities}, where the question asks about culture-specific entities \cite{naous2025origin}, \textbf{(2)} \textbf{Cultural Opinion}, where the question asks about a subjective interpretation for a cultural entity, \textbf{(3)} \textbf{Social Norms}, where the question is about social interactions between individuals \cite{huang2023culturally}, \textbf{(4)} \textbf{Cultural Commonsense}, where the question is about daily phenomena that locals may take for granted \cite{shen2024understanding}, and \textbf{(5)} \textbf{Literacy}, where the question is about the language, proverbs, or slang \cite{wuraola2024understanding}. After filtering all resources, we found an imbalanced coverage across categories, with a notable lack of data on social norms and commonsense, as shown in Figure~\ref{fig:care-statistics}. Also, to align LMs via preference optimization, we require human judgments of both appropriate and inappropriate responses, which are not offered by existing resources.

\subsection{Multilingual Cultural Preference Data}
\label{sub:care_data}

To address the aforementioned shortcomings, we collect new data in three languages as follows:

\paragraph{Social Norm and Commonsense.} 
To expand the samples on social norms and cultural commonsense, we ask native Chinese, Arabic, and Japanese speakers who are international college students undergoing culture transfer experiences themselves to curate such samples. 
To help brainstorm, the speakers are instructed to leverage international as well as regional social media platforms or forums (e.g., Twitter, Reddit, Zhihu, RedNote), and search for posts where users describe their culture shift experiences using search keywords such as ``\textit{most surprised abroad}'', ``\textit{culture shock}'', ``\textit{first time to}'', etc. 
Taking inspiration from such discussions, the speakers create 190, 196, and 260 samples in Chinese, Arabic, and Japanese.
This human-curated data is more authentic than synthetic data, preventing some inaccuracy and overgeneralization (e.g., \textit{``In China, it is common to drink soup after the main dish.''}) we have observed in the existing datasets.

\paragraph{Culture-specific Human Preference Judgments.}

Given these culture-specific questions, gathering human preferences on the cultural relevance of responses is crucial to performing cultural alignment through preference optimization. To obtain these cultural preferences, we present the native annotators with the zero-shot responses of 9 different LMs to each question within CARE. We specifically use the instruct version of recent multilingual LMs of \texttt{Llama3.3-70B} \cite{dubey2024llama}, \texttt{Qwen2.5-72B} \cite{yang2024qwen2}, \texttt{Gemma2-27B} \cite{team2024gemma}, \texttt{Mistral-Large}, and \texttt{GPT-4o}. We also use their smaller-sized \texttt{Llama3.1-8B}, \texttt{Qwen2.5-7B}, \texttt{Gemma2-9B}, and \texttt{Mistral-7B}.

We instruct annotators to rank the generated responses from the most to least culturally appropriate and assign a rating on a 1–10 scale (1: \textit{poor} $\rightarrow$ 10: \textit{excellent}). These ratings reflect how well responses align with the cultural expectations of native speakers and are used to construct preference pairs for cultural preference learning (\S\ref{subsec:cultural-performance}).



\begin{table}[t!]
\centering
\resizebox{0.6\linewidth}{!}{%
\begin{tabular}{ccccc}
\toprule
\textbf{Culture} & {$\alpha$} & \textbf{$r$} & \textbf{${\rho}$} & \textbf{$\tau$} \\
\midrule
Arab & 0.86 & 0.90 & 0.88 & 0.71 \\
Chinese & 0.84 & 0.89 & 0.89 & 0.75 \\
Japanese & 0.92 & 0.93 & 0.92 & 0.77 \\
\bottomrule
\end{tabular}%
}
\caption{Inter-annotator agreement based on Krippendorff's $\alpha$, Pearson's $r$, Spearman's $\rho$, and Kendall's $\tau$.}
\label{tab:inter-agreement}
\vspace{-1.0em}
\end{table}

\paragraph{Human Annotators and Agreement.} 
To curate the samples in CARE, we recruited 2 Chinese, 2 Arab, and 5 Japanese native speakers who are familiar with the respective cultures. These annotators manually filtered the aforementioned resources and searched online platforms for additional samples, capturing broader community knowledge and ensuring that CARE reflects cultural experiences beyond our annotator pool. The annotators then rated the responses to the questions in CARE.


To ensure the quality of our preference ratings, we additionally recruited 4 Chinese, 1 Arab, and 4 Japanese annotators, who were not involved in the data curation process, to perform double annotation. Each additional annotator was given 75 randomly sampled questions (15 per cultural category), for which they were asked to provide their ratings of model responses. We then compare the ratings of the additional annotators to our initial set of annotators.  As shown in Table \ref{tab:inter-agreement}, we find substantial inter-annotator agreements across various metrics, indicating consistent response preferences for questions within our considered cultural categories (see  Appendix~\ref{appendix:annotation-agreement} for per-category agreements).

\begin{table*}[t]  
  \centering  
  \begin{adjustbox}{width=\linewidth}  
    \tiny 
    \setlength{\tabcolsep}{0.4pt} 
    \begin{tabularx}{\linewidth}{@{}l c *{6}{>{\centering\arraybackslash}X} *{6}{>{\centering\arraybackslash}X} *{6}{>{\centering\arraybackslash}X} @{}}    
      \toprule   
      & & \multicolumn{6}{c}{\textbf{Chinese}} & \multicolumn{6}{c}{\textbf{Arabic}} & \multicolumn{6}{c}{\textbf{Japanese}} \\   
      \cmidrule(lr){3-8} \cmidrule(l){9-14} \cmidrule(l){15-20}    
      \textbf{Model} & \textbf{} & \textit{Entities} & \textit{Opinion} & \textit{Norms} & \makebox[0pt][c]{\textit{C. sense}} & \textit{Literacy} & \textit{Average} & \textit{Entities} & \textit{Opinion} & \textit{Norms} & \makebox[0pt][c]{\textit{C. sense}} & \textit{Literacy} & \textit{Average} & \textit{Entities} & \textit{Opinion} & \textit{Norms} & \makebox[0pt][c]{\textit{C. sense}} & \textit{Literacy} & \textit{Average} \\
      \midrule
      \multirow{3}{*}{\includegraphics[width=0.2cm]{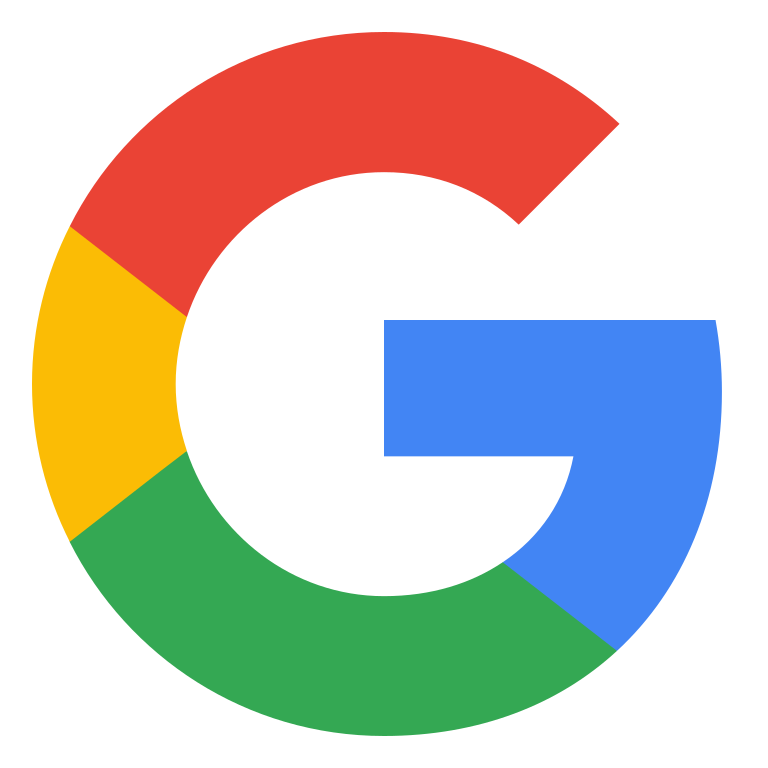}~Gemma2-9B}
      & Vanilla 
        & 4.89 & 8.46 & 7.55 & 6.67 & 4.72 & 6.49 
        & 4.73 & 6.56 & 6.33 & 5.60 & 3.28 & 5.33 & \textbf{4.46} & 6.50 & 5.67 & 6.50 & 2.33 & 5.09 \\
      & Aligned\includegraphics[width=0.25cm]{fig/Hugging.jpg}
        & \textbf{5.50} & \textbf{8.53} & \textbf{7.90} & \textbf{7.26} & \textbf{5.17} & \textbf{6.89} 
        & \textbf{5.30} & \textbf{6.82} & \textbf{6.93} & \textbf{6.55} & \textbf{3.57} & \textbf{5.84} & 4.40 & \textbf{6.60} & \textbf{5.80} & \textbf{6.67} & \textbf{2.93} & \textbf{5.28} \\
      & 
        \multicolumn{1}{l}{} 
        & \multicolumn{6}{l}{\includegraphics[width=0.24\textwidth]{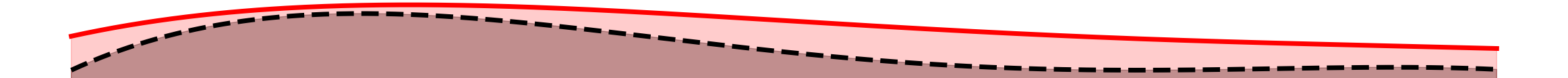}} 
        & \multicolumn{6}{l}{\includegraphics[width=0.24\textwidth]{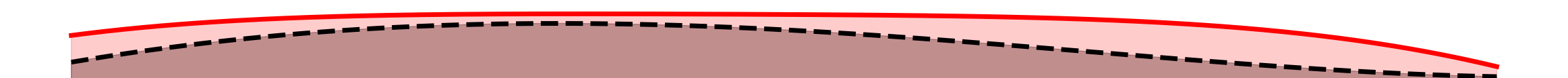}} & \multicolumn{6}{l}{\includegraphics[width=0.24\textwidth]{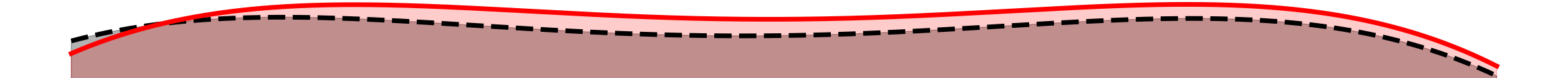}} \\
      \multirow{3}{*}{\includegraphics[width=0.2cm]{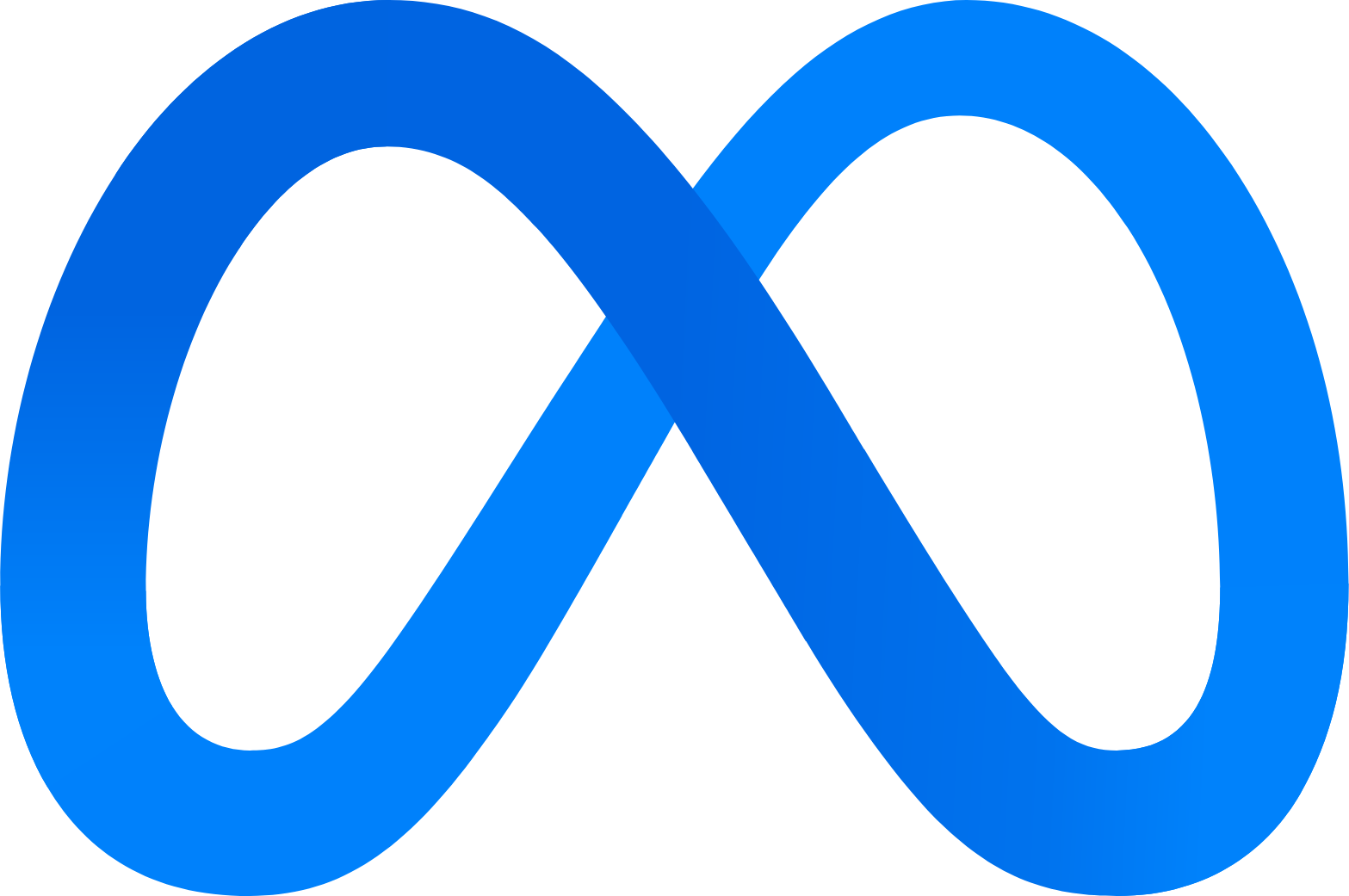}~Llama3.1-8B}
      & Vanilla
        & 3.14 & 4.16 & 4.62 & 3.93 & 3.03 & 3.78
        & 4.08 & 3.62 & 2.87 & 3.07 & 2.07 & 3.30 & 2.20 & 3.93 & 2.47 & 2.57 & \textbf{1.97} & 2.63 \\
      & Aligned\includegraphics[width=0.25cm]{fig/Hugging.jpg}
        & \textbf{3.86} & \textbf{5.90} & \textbf{5.36} & \textbf{5.56} & \textbf{3.69} & \textbf{4.88}
        & \textbf{4.33} & \textbf{4.43} & \textbf{3.70} & \textbf{4.50} & \textbf{2.36} & \textbf{3.86} & \textbf{3.20} & \textbf{4.67} & \textbf{2.80} & \textbf{3.20} & 1.67 & \textbf{3.11} \\
      &
        \multicolumn{1}{l}{} 
        & \multicolumn{6}{l}{\includegraphics[width=0.24\textwidth]{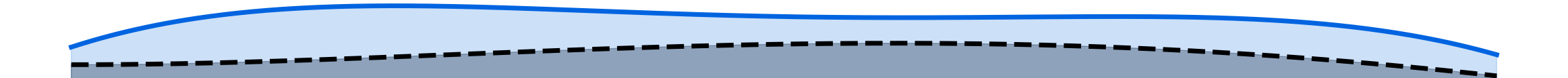}} 
        & \multicolumn{6}{l}{\includegraphics[width=0.24\textwidth]{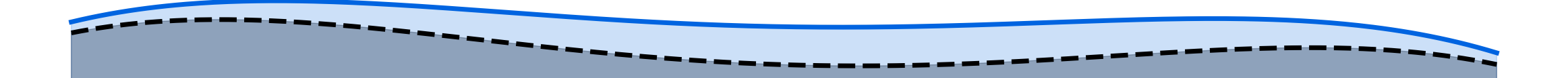}} & \multicolumn{6}{l}{\includegraphics[width=0.24\textwidth]{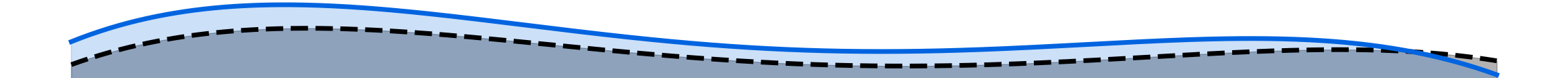}} \\
      \multirow{3}{*}{\includegraphics[width=0.2cm]{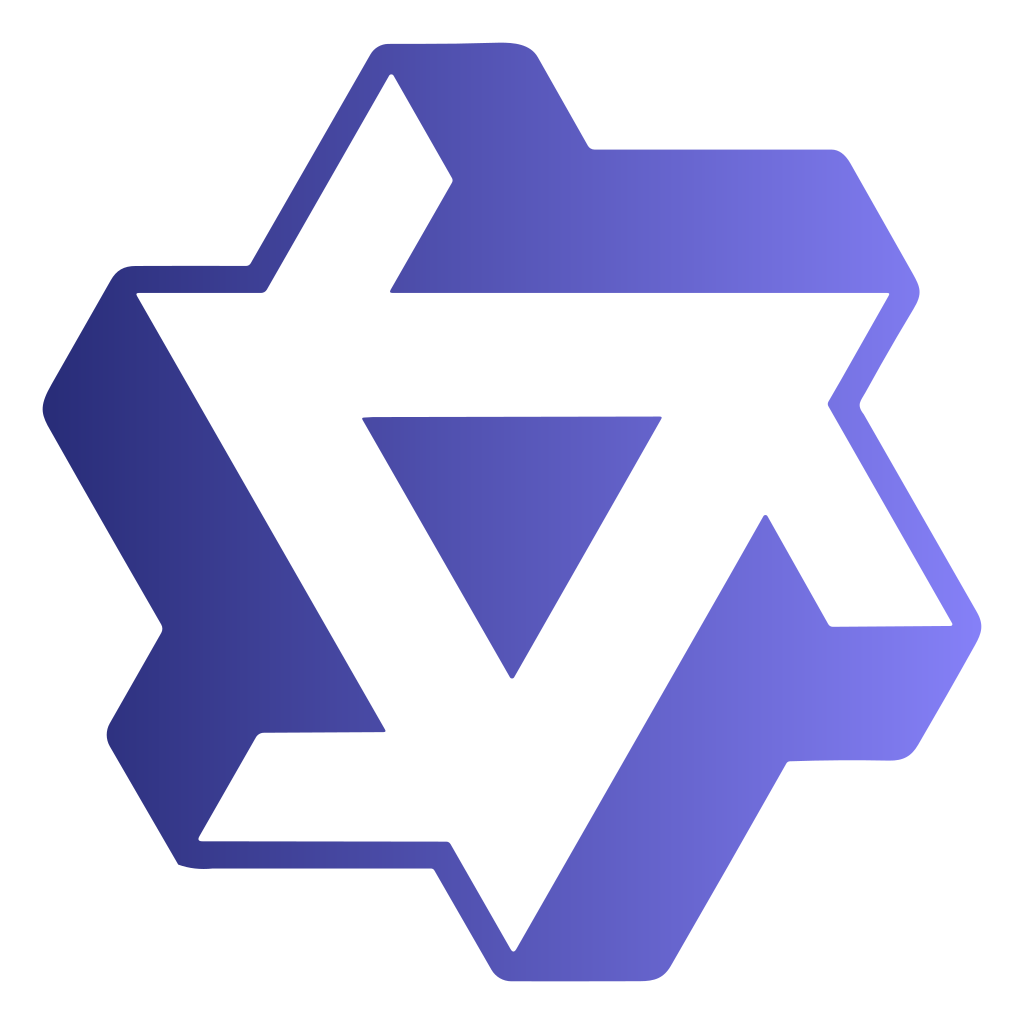}~Qwen2.5-7B}
      & Vanilla
        & 6.89 & 7.86 & 7.48 & 6.80 & 7.37 & 7.28
        & \textbf{4.65} & 5.84 & 5.44 & 4.88 & 2.84 & 4.61 & 2.13 & 5.67 & 4.13 & 4.60 & \textbf{2.37} & 3.78 \\
      & Aligned\includegraphics[width=0.25cm]{fig/Hugging.jpg}
        & \textbf{7.20} & \textbf{8.76} & \textbf{7.66} & \textbf{6.90} & \textbf{7.53} & \textbf{7.61}
        & 4.55 & \textbf{6.40} & \textbf{5.55} & \textbf{5.33} & \textbf{3.35} & \textbf{5.06} & \textbf{2.77} & \textbf{5.73} & \textbf{4.30} & \textbf{5.17} & 1.90 & \textbf{3.97}  \\
      &
        \multicolumn{1}{l}{}
        & \multicolumn{6}{l}{\includegraphics[width=0.24\textwidth]{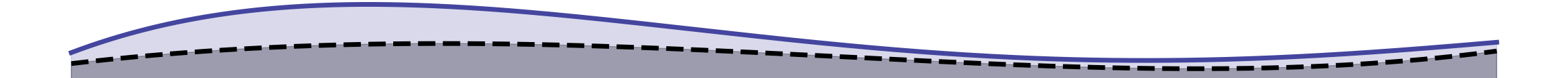}}
        & \multicolumn{6}{l}{\includegraphics[width=0.24\textwidth]{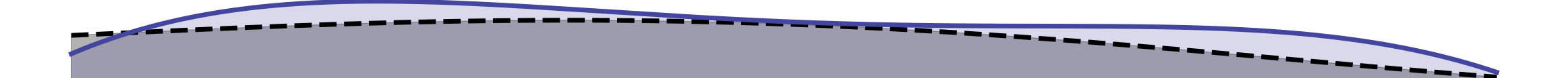}} & \multicolumn{6}{l}{\includegraphics[width=0.24\textwidth]{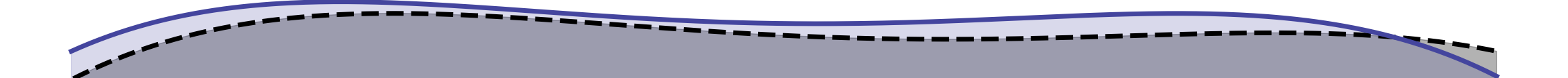}} \\
      \multirow{3}{*}{\includegraphics[width=0.2cm]{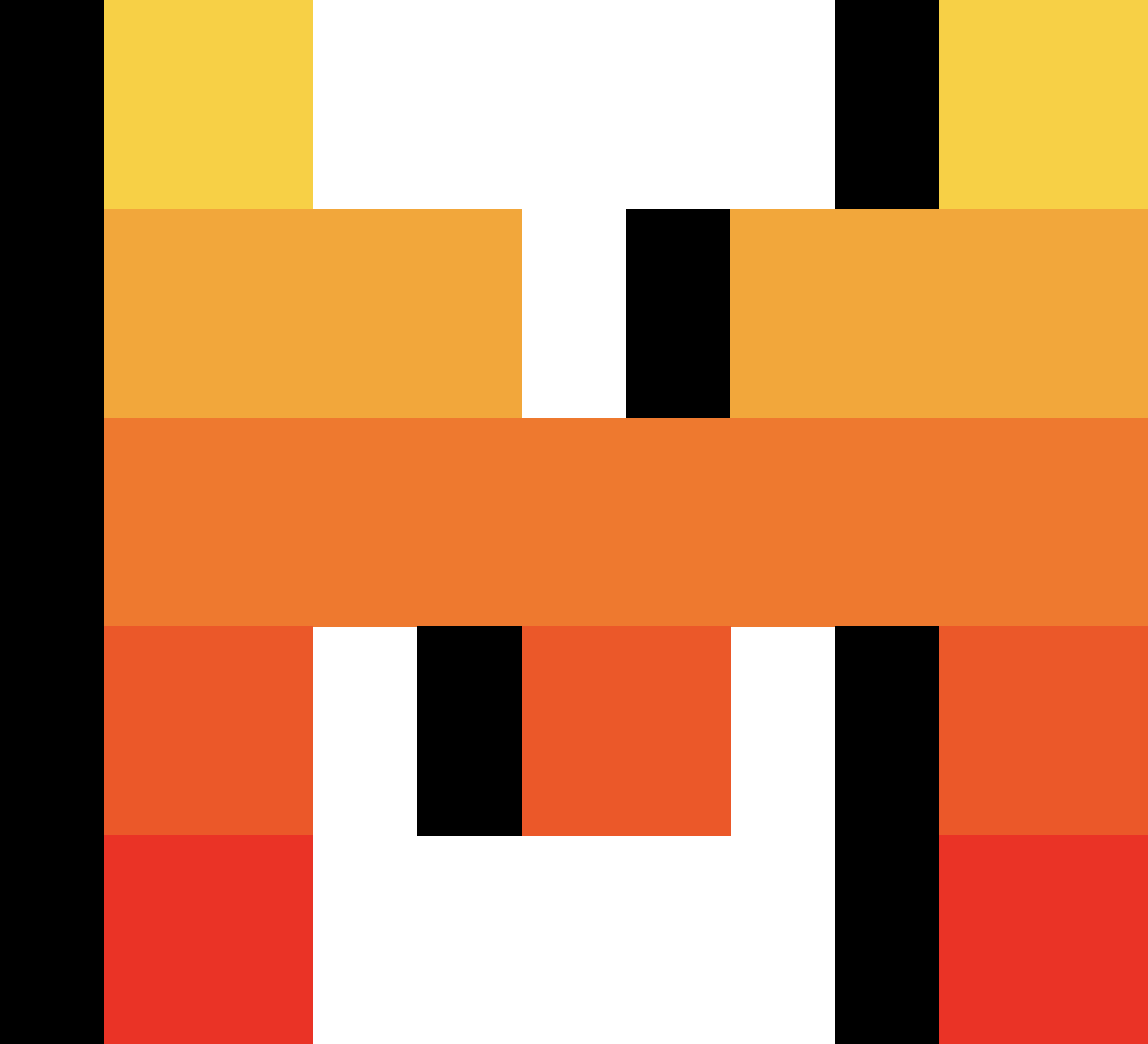}~Mistral-7B}
      & Vanilla
        & \textbf{3.03} & 3.83 & 3.93 & 4.38 & \textbf{2.43} & 3.53
        & 2.56 & 2.46 & 2.03 & \textbf{2.13} & 1.34 & 2.11 & 1.70 & \textbf{3.96} & 2.40 & \textbf{2.48} & 1.20 & 2.34 \\
      & Aligned\includegraphics[width=0.25cm]{fig/Hugging.jpg}
        & 2.43 & \textbf{3.90} & \textbf{4.53} & \textbf{5.00} & 2.20 & \textbf{3.61}
        & \textbf{2.60} & \textbf{3.36} & \textbf{2.46} & 2.10 & \textbf{1.40} & \textbf{2.38} & \textbf{1.83} & 3.90 & \textbf{2.43} & 2.33 & \textbf{1.23} & \textbf{2.35} \\
      &
        \multicolumn{1}{l}{}
        & \multicolumn{6}{l}{\includegraphics[width=0.24\textwidth]{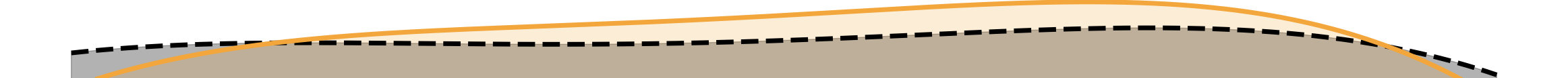}}
        & \multicolumn{6}{l}{\includegraphics[width=0.24\textwidth]{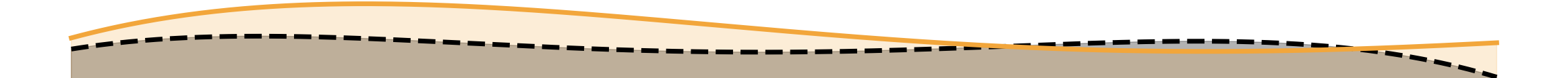}} & \multicolumn{6}{l}{\includegraphics[width=0.24\textwidth]{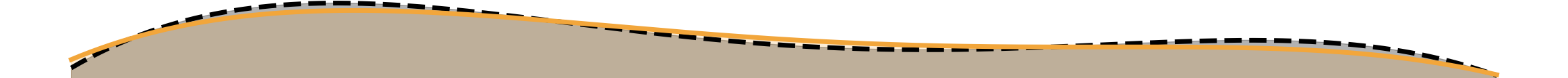}} \\
      \bottomrule    
    \end{tabularx}    
  \end{adjustbox}    
  
  \caption{
    Average scores (1: \textit{poor} $\rightarrow$ 10: \textit{excellent}) in responding to questions related to Chinese culture in Chinese, Arab culture in Arabic, and Japanese culture in Japanese. 
    Performances are presented for vanilla LMs and LMs after cultural alignment using DPO on CARE\includegraphics[width=0.5cm]{fig/Hugging.jpg}. 
    For each LM, the row labeled ``\textit{Vanilla}'' corresponds to the original model (gray plot), and ``\textit{Aligned}'' is after cultural preference learning (colored plot).
  }
  \label{tab:main-results-evaluation}  
  \vspace{-1em}
\end{table*}

Besides the 5-class cultural categories, we also label each sample with the \textbf{Associated Culture} for experiments in \S \ref{subsec:source-culture-analysis}: \textit{Native} (questions about the local culture; i.e., Chinese, Arab, or Japanese), \textit{Foreign} (questions about other cultures; e.g., US, German, etc.), or \textit{General} (questions that are not specific to a particular culture). For fine-grained regional cultural knowledge evaluation (Appendix~\ref{appendix:geographic_scope}), we also annotate samples for their \textbf{Geographic Scope} (\textit{sub-nationwide}, \textit{nationwide}, \textit{continent-wide}, or \textit{worldwide}).  See Appendix~\ref{appendix:annotation-guideline} for the annotation guideline.

\paragraph{Data Split.} 
For the experiments in \S \ref{sec:cultural-alignment} and \S \ref{sec:prompting_evaluation}, we construct culture-specific test sets of 150 questions each for Chinese, Arab, and Japanese cultures, sampling 30 questions per cultural category. Remaining CARE questions (1,513 Chinese, 644 Arabic, 757 Japanese), along with rated responses, form the training set. 
This training data also includes questions about other cultures (e.g. U.S., German, etc.) beyond Chinese, Arab, and Japanese.

\section{Aligning LMs for Cultural Awareness}
\label{sec:cultural-alignment}

We investigate whether, and to what extent, high-quality culture-specific data can enhance LMs. We perform human preference learning for cultural alignment of medium-sized LMs (7$\sim$9B parameters) using CARE that fit in 8 Nvidia A40 GPUs.





\subsection{Experiment Setup}
\label{subsec:cultural-performance}


\paragraph{Human Preference Optimization.}
The goal of cultural preference optimization is to align model outputs with culturally appropriate responses, informed by human judgments. 
Each training instance $\mathcal{D}$ consists of a prompt $x$ and a pair of responses $(y_w, y_\ell)$, where $y_w$ is the preferred and $y_\ell$ is the dispreferred. 
In {CARE}, we construct these pairs by taking the highest- and lowest-scored responses from the human annotations.
We evaluate the following representative algorithms:

\textbf{DPO}~\cite{rafailov2024direct} directly optimizes the log-odds difference between the preferred and dispreferred responses, measured against a reference model:
\vspace{-.8em}
\begin{center}
\resizebox{\linewidth}{!}{%
  $\begin{aligned}
    \mathcal{L}_{\mathrm{DPO}}
    &= - \log \sigma\!\Big(
    \beta \big[\log \pi_\theta(y_w\mid x)-\log \pi_{\mathrm{ref}}(y_w\mid x)\big] \\
    &\quad - \beta \big[\log \pi_\theta(y_\ell\mid x)-\log \pi_{\mathrm{ref}}(y_\ell\mid x)\big]
    \Big),
  \end{aligned}$%
}
\end{center}

\noindent where $\sigma$ is the sigmoid and $\beta$ is a scaling factor.

\textbf{KTO}~\cite{ethayarajh2024kto} introduces a reference baseline:

\vspace{-.2em}
\begin{center}
\resizebox{0.9\linewidth}{!}{%
  $\begin{aligned}
   z_{\mathrm{ref}}
   &= \mathbb{E}_{(x, y)\sim \mathcal{D}}\!\left[
      \beta\, D_{\mathrm{KL}}\!\big(
        \pi_\theta(y\mid x)\,\|\,\pi_{\mathrm{ref}}(y\mid x)
      \big)
     \right]
  \end{aligned}$%
}
\end{center}
and weighting desirable and undesirable examples with $\lambda_w,\lambda_\ell>0$:
\vspace{-.8em}
\begin{center}
\resizebox{\linewidth}{!}{%
  $\begin{aligned}
   \mathcal{L}_{\mathrm{KTO}}
   &= - \lambda_w\, \sigma\!\Big(
   \beta \big[\log \pi_\theta(y_w\mid x)-\log \pi_{\mathrm{ref}}(y_w\mid x)\big] - z_{\mathrm{ref}}
   \Big) \\
   &\quad + \lambda_\ell\, \sigma\!\Big(
   z_{\mathrm{ref}} - \beta \big[\log \pi_\theta(y_\ell\mid x)-\log \pi_{\mathrm{ref}}(y_\ell\mid x)\big]
   \Big).
  \end{aligned}$%
}
\end{center}

\textbf{SimPO}~\cite{meng2024simpo} removes the dependence on a reference model by directly comparing the length-normalized log-likelihoods of the two responses, with a margin $\gamma$:
\vspace{-1em}
\begin{center}
\resizebox{\linewidth}{!}{%
  $\begin{aligned}
   \mathcal{L}_{\mathrm{SimPO}}
   &= - \log \sigma\!\Big(
   \tfrac{\beta}{|y_w|}\,\log \pi_\theta(y_w\mid x)
   - \tfrac{\beta}{|y_\ell|}\,\log \pi_\theta(y_\ell\mid x)
   - \gamma
   \Big).
  \end{aligned}$%
}
\end{center}

Across these methods, we find that all achieve comparable performance on CARE (Table~\ref{tab:baselines-results}). 
Subsequently, we present most of the experiments in this paper with DPO unless otherwise specified.

\begin{table*}[t!]
\centering
\begin{adjustbox}{width=\linewidth}
\begin{tabular}{@{}lcccccccccccc@{}}
\toprule
 & \multicolumn{3}{c}{\includegraphics[width=0.4cm] {fig/google-logo.png} \textbf{Gemma2-9B}} & \multicolumn{3}{c}{\includegraphics[width=0.4cm] {fig/qwen-logo.png} \textbf{Qwen2.5-7B}} & \multicolumn{3}{c}{\includegraphics[width=0.4cm] {fig/meta-logo.png} \textbf{Llama3.1-8B}} & \multicolumn{3}{c}{\includegraphics[width=0.4cm] {fig/mistral-logo.png} \textbf{Mistral-7B}}   \\ \cmidrule(l){2-13} 
\textbf{Approach (w/ data)} & \textit{Arabic} & \textit{Chinese} & \textit{Japanese} & \textit{Arabic} & \textit{Chinese} & \textit{Japanese} & \textit{Arabic} & \textit{Chinese} & \textit{Japanese} & \textit{Arabic} & \textit{Chinese} & \textit{Japanese} \\ \midrule
\textit{0-shot Prompting} & \multicolumn{1}{l}{} & \multicolumn{1}{l}{} & \multicolumn{1}{l}{} & \multicolumn{1}{l}{} & \multicolumn{1}{l}{} & \multicolumn{1}{l}{} & \multicolumn{1}{l}{} & \multicolumn{1}{l}{} \\
Vanilla & 5.331 & 6.490 & 5.093 & 4.618 & 7.286 & 3.780 & 3.304 & 3.784 & 2.627 & 2.114 & 3.534 & 2.339   \\
SFT (w/ Alpaca) & 5.443 & 6.416 & 3.447 & 4.689 & 5.093 & {3.387} & 3.141 & 3.709 & 2.433 & {1.287} & {2.100} & {1.673}   \\
SFT (w/ CARE\includegraphics[width=0.5cm]{fig/Hugging.jpg}) & {5.463} & {6.440} & {3.493} & {4.700} & {5.396} & 3.219 & {3.440} & {3.813} & {2.673} & 1.360 & {2.627} & 1.653   \\
DPO (w/ UltraFeedback) & 5.765 & 6.380 & -- & 4.845 & 7.547 & -- & 3.880 & 4.160 & -- & 2.220 & 3.307 & --  \\
DPO (w/ OpenOrca) & 5.564 & 6.260 & 5.060 & \colorbox{green4!10}{\underline{4.878}} & 7.433 & 3.653 & 3.456 & 3.260 & 2.547 & 2.067 & 3.480 & 1.747 \\
DPO (w/ HelpSteer3) & -- & 6.133 & 4.973 & -- & 7.000 & 3.800 & -- & 3.280 & 2.673 & -- & \colorbox{green4!30}{\textbf{3.687}} & 2.215 \\
DPO (w/ CARE\includegraphics[width=0.5cm]{fig/Hugging.jpg}) & 5.848 & \colorbox{green4!30}{\textbf{6.899}} & \colorbox{green4!10}{\underline{5.280}} & \colorbox{green4!30}{\textbf{5.062}} & 7.613 & \colorbox{green4!10}{\underline{3.980}} & 3.867 & \colorbox{green4!10}{\underline{4.886}} & \colorbox{green4!10}{\underline{3.107}} & \colorbox{green4!10}{\underline{2.387}} & \colorbox{green4!10}{\underline{3.613}} & \colorbox{green4!10}{\underline{2.349}}   \\  
KTO (w/ CARE\includegraphics[width=0.5cm]{fig/Hugging.jpg}) & \colorbox{green4!30}{\textbf{6.387}} & \colorbox{green4!10}{\underline{6.713}} & \colorbox{green4!30}{\textbf{5.473}} & 4.822 & \colorbox{green4!10}{\underline{7.617}} & \colorbox{green4!30}{\textbf{4.147}} & 3.911 & 4.691 & 3.013 & \colorbox{green4!30}{\textbf{2.687}} & 3.513 & \colorbox{green4!30}{\textbf{2.473}}  \\  
SimPO (w/ CARE\includegraphics[width=0.5cm]{fig/Hugging.jpg}) & \colorbox{green4!10}{\underline{5.932}} & 6.647 & 5.033 & 4.765 & 7.427 & 3.847 & \colorbox{green4!10}{\underline{3.917}} & \colorbox{green4!30}{\textbf{4.946}} & 2.947 & 2.253 & 3.480 & 2.093  \\  
MAPO (w/ CARE\includegraphics[width=0.5cm]{fig/Hugging.jpg}) & 5.758 & 6.340 & 5.153 & 4.820 & \colorbox{green4!30}{\textbf{7.640}} & 3.613 & \colorbox{green4!30}{\textbf{4.107}} & 4.753 & \colorbox{green4!30}{\textbf{3.287}} & 2.327 & {3.580} & 2.133  \\  
  \hdashline[1pt/1pt]
\textit{CoT Prompting} & \multicolumn{1}{l}{} & \multicolumn{1}{l}{} & \multicolumn{1}{l}{} & \multicolumn{1}{l}{} & \multicolumn{1}{l}{} & \multicolumn{1}{l}{} & \multicolumn{1}{l}{} & \multicolumn{1}{l}{} \\
Vanilla & {5.946} & 6.081 & 4.613 & 4.703 & 7.667 & 3.873 & 3.107 & 3.887 & \colorbox{green4!30}{\textbf{2.927}} & {2.333} & \colorbox{green4!30}{\textbf{4.373}} & \colorbox{green4!30}{\textbf{2.273}}   \\
DPO (w/ CARE\includegraphics[width=0.5cm]{fig/Hugging.jpg}) & \colorbox{green4!30}{\textbf{6.096}} & \colorbox{green4!30}{\textbf{6.407}} & \colorbox{green4!30}{\textbf{5.093}} & \colorbox{green4!30}{\textbf{4.946}} & \colorbox{green4!30}{\textbf{7.703}} & \colorbox{green4!30}{\textbf{4.220}} & \colorbox{green4!30}{\textbf{3.678}} & \colorbox{green4!30}{\textbf{5.087}} & 2.840 & \colorbox{green4!30}{\textbf{2.427}} & 4.233  & 2.173  \\  \hdashline[1pt/1pt]
\textit{Role-Play Prompting} & \multicolumn{1}{l}{} & \multicolumn{1}{l}{} & \multicolumn{1}{l}{} & \multicolumn{1}{l}{} & \multicolumn{1}{l}{} & \multicolumn{1}{l}{} & \multicolumn{1}{l}{} & \multicolumn{1}{l}{} \\
Vanilla & 4.073 & 6.396 & 5.207 & 4.899 & \colorbox{green4!30}{\textbf{7.939}} & 4.100 & 3.500 & 4.087 & 2.547 & \colorbox{green4!30}{\textbf{2.513}} & 3.530 & \colorbox{green4!30}{\textbf{2.320}}   \\
DPO (w/ CARE\includegraphics[width=0.5cm]{fig/Hugging.jpg}) & \colorbox{green4!30}{\textbf{5.938}} & \colorbox{green4!30}{\textbf{6.561}} & \colorbox{green4!30}{\textbf{5.527}} & \colorbox{green4!30}{\textbf{5.129}} & {7.878} & \colorbox{green4!30}{\textbf{4.313}} & \colorbox{green4!30}{\textbf{3.899}} & \colorbox{green4!30}{\textbf{5.093}}  & \colorbox{green4!30}{\textbf{2.953}} & 2.362 & \colorbox{green4!30}{\textbf{3.720}} & 2.167  \\ \bottomrule
\end{tabular}
\end{adjustbox}
\vspace{-.2em}
\caption{
Average scores (1: \textit{poor} $\rightarrow$ 10: \textit{excellent}) on Chinese, Arab, and Japanese cultures for a variety of prompting approaches, supervised fine-tuning, and preference learning using culture-specific (CARE) vs. general instruction-tuning (multilingual Alpaca) and preference (translated and filtered OpenOrca/UltraFeedback) data. 
SFT is performed on the instruction data only, while preference learning is conducted on the preference pairs. 
}
\label{tab:baselines-results}
\vspace{-1em}
\end{table*}

\paragraph{Baselines.} 

We compare cultural preference tuning on CARE with four groups of baselines:
(1) General-domain preference datasets: we adopt the multilingual translations~\cite{arabic-dpo-pairs,chinese-orca-dpo-pairs,japanese-orca-dpo-pairs} of two synthetic English preference datasets, \textbf{OpenOrca}~\cite{intel-orca-dpo-pairs} (2.1k Ar, 2.0k Zh, 12.9k Ja pairs) and \textbf{UltraFeedback}~\cite{cui2023ultrafeedback} (2.0k Ar, 1.7k Zh pairs). 
We also compare with \textbf{HelpSteer3}~\cite{wang2025helpsteer3preferenceopenhumanannotatedpreference}, a human-annotated multilingual preference dataset (1.1k Zh, 0.2k Ja pairs). 
All focus on truthfulness, honesty, and helpfulness. 
(2) Instruction‐tuned models: we conduct SFT on our CARE and the multilingual \textbf{Alpaca} corpus~\cite{chinese-llama-alpaca, Chen_MultilingualSIFT_Multilingual_Supervised_2023, japanese-alpaca}.
(3) Multilingual preference tuning: \textbf{MAPO}~\cite{she2024mapo}.
(4) Prompting methods: \textbf{CoT}~\cite{wei2022chain} and \textbf{role-play}~\cite{kong2023better}.
Implementation details are in Appendix~\ref{appendix:implementation-hyperparameters}.

\paragraph{LM-as-a-judge Evaluation and its Reliability.} 

For evaluation, we adopt the LM-as-a-judge strategy \cite{zheng2023judging}, prompting \texttt{GPT-4o} to generate a rationale and assign a 1-10 score on how aligned the response is to the human reference. We validate this setup by comparing \texttt{GPT-4o}'s ratings with native annotators', where we achieve a high Pearson correlation of 0.93. More details and full evaluation prompts can be found in Appendix~\ref{appendix:llm-judge-prompts}.

\subsection{Main Results} 
\label{subsec:main-res}

Table~\ref{tab:main-results-evaluation} presents the results of medium-sized LMs before and after cultural preference learning with CARE, while Table~\ref{tab:baselines-results} compares results with various baselines. We see the following key findings:

Using CARE, culturally-aligned LMs achieve higher average scores (up to 29\% improvement) compared to the vanilla checkpoints across all three cultures. 
This resonates with the findings of \citet{zhou2024lima} and shows that a relatively small amount of carefully curated data by humans can improve LMs' alignment. 
We also see a noticeable gap among LMs developed by different regions. The \texttt{Qwen2.5-7B} developed by China-based Alibaba performs the best in Chinese and can be further improved from 7.28 to 7.61 (out of 10) by conducting preference optimization with CARE.  
Interestingly, \textbf{\textit{the aligned version of \texttt{Gemma2-9B} outperforms \texttt{Qwen2.5-7B} on Chinese social norms and commonsense}} (more on this in \S \ref{subsec:presence-language}).

\textbf{\textit{Cultural preference learning strengthens stronger LMs but fails to address weaknesses in weaker LMs.}} 
\texttt{Qwen2.5-7B} and \texttt{Gemma2-9B}, the top-performing LMs on Chinese and Arabic, respectively, show consistent improvement across cultural categories after preference learning.
However, preference learning shows limited gains where the model's initial performance is poor. 
\texttt{Mistral-7B} is not improved on Chinese entities and literacy, where its starting scores are only 3.03 and 2.43. 
Similarly, \texttt{Qwen2.5-7B}, does not benefit from preference learning on Arabic entities. 
Models also show limited progress on Japanese data, reflecting their shallow exposure to Japanese culture. 
Similarly, when the model possesses foundational cultural knowledge, utilizing the preference signals from the model itself via methods like MAPO rather than directly from human preferences can also lead to improvements. 
All these suggest that base models need basic cultural knowledge for preference learning to be effective.

\textbf{\textit{Culture-specific human preference data provides complementary benefits.}}
Table~\ref{tab:baselines-results} shows that preference tuning with CARE yields competitive or stronger results than training on larger general datasets (3–17$\times$ more pairs). 
Cultural preference learning with CARE also improves performance in both CoT and role-play prompting setups. 
While culture-specific SFT outperforms generic SFT, preference tuning yields greater gains, highlighting the value of \textit{\textbf{human cultural preferences over ground-truth demonstrations alone}}.

\begin{figure}[t]
    \centering
    \includegraphics[width=0.9\linewidth]{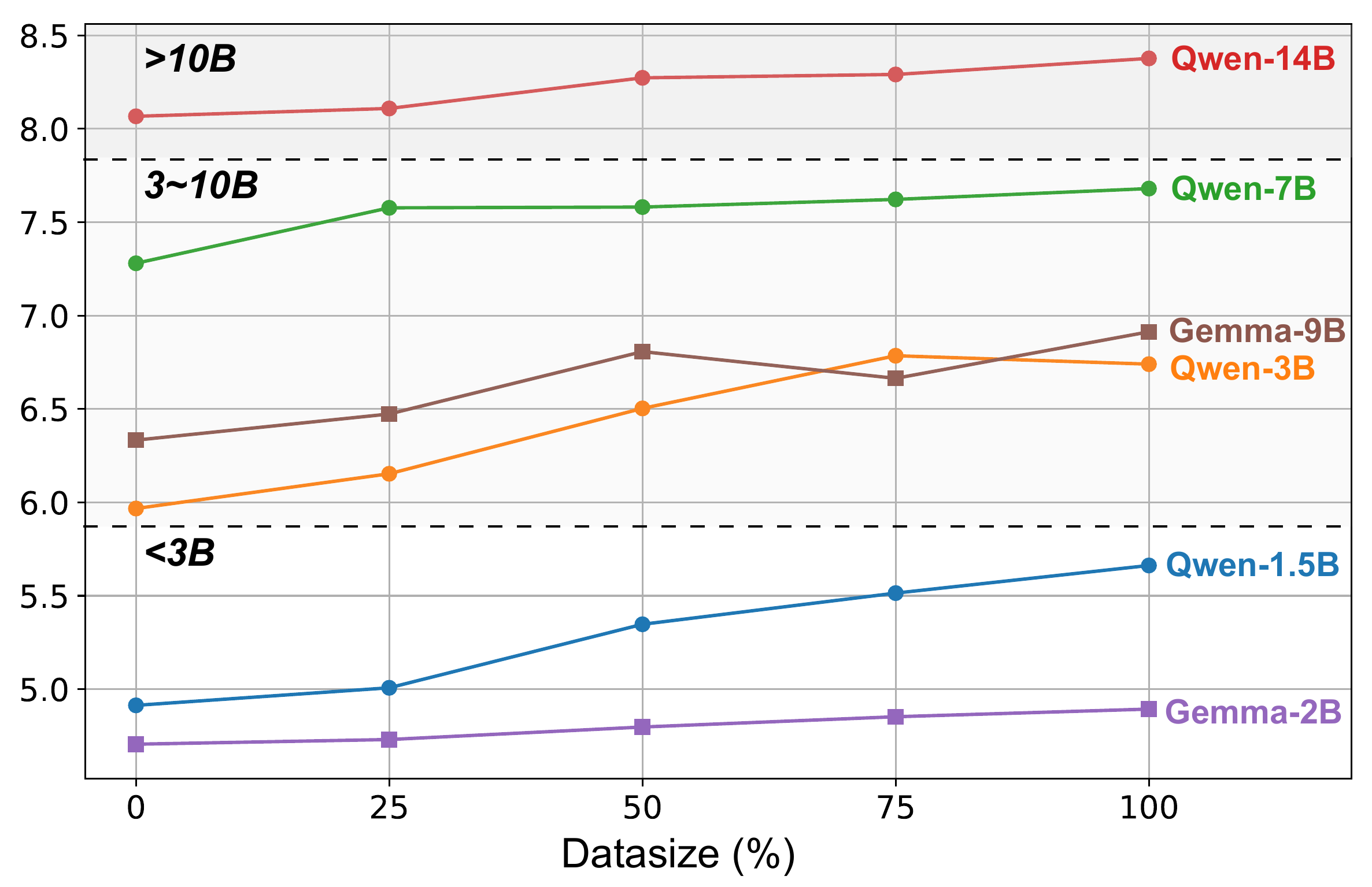}
    \vspace{-.3em}
    \caption{Impact of model size and preference data volume on cultural awareness performance. The average scores (1: \textit{poor} $\rightarrow$ 10: \textit{excellent})  of aligned models are plotted against different \% of preference pairs in CARE. Improvements are achieved across different model sizes and data sizes, in comparison to the vanilla model (0\%).}
    \label{fig:scaling}
    \vspace{-1em}
\end{figure}

\subsection{Scaling of Data and Model Sizes}
\label{subsec:scaling-exp}

We examine how scaling both model size and data volume influences the performance before and after cultural preference learning. Specifically, we consider the \texttt{Qwen2.5} \texttt{\{1.5B, 3B, 7B, 14B\}} and the \texttt{Gemma2} \texttt{\{2B, 9B\}} series, which offer multiple model sizes. For data sizes, we align models using DPO on different proportions \texttt{\{0\%, 25\%, 50\%, 75\%, 100\%\}} of Chinese cultural preference pairs within CARE.  
As shown in Figure~\ref{fig:scaling}, consistent improvements are observed across different model and data sizes. Even a relatively small amount (e.g., 25\%) of data in CARE  can lead to improvements, particularly for smaller-sized LMs.
Scaling up the data leads to progressively better performance, highlighting \textbf{\textit{the benefits of employing more cultural preference data in future work}}.





\subsection{Cultural Generalization} 
\label{sebsec:exp-generalize}

\begin{figure}[t]
    \centering
    \includegraphics[width=\linewidth]{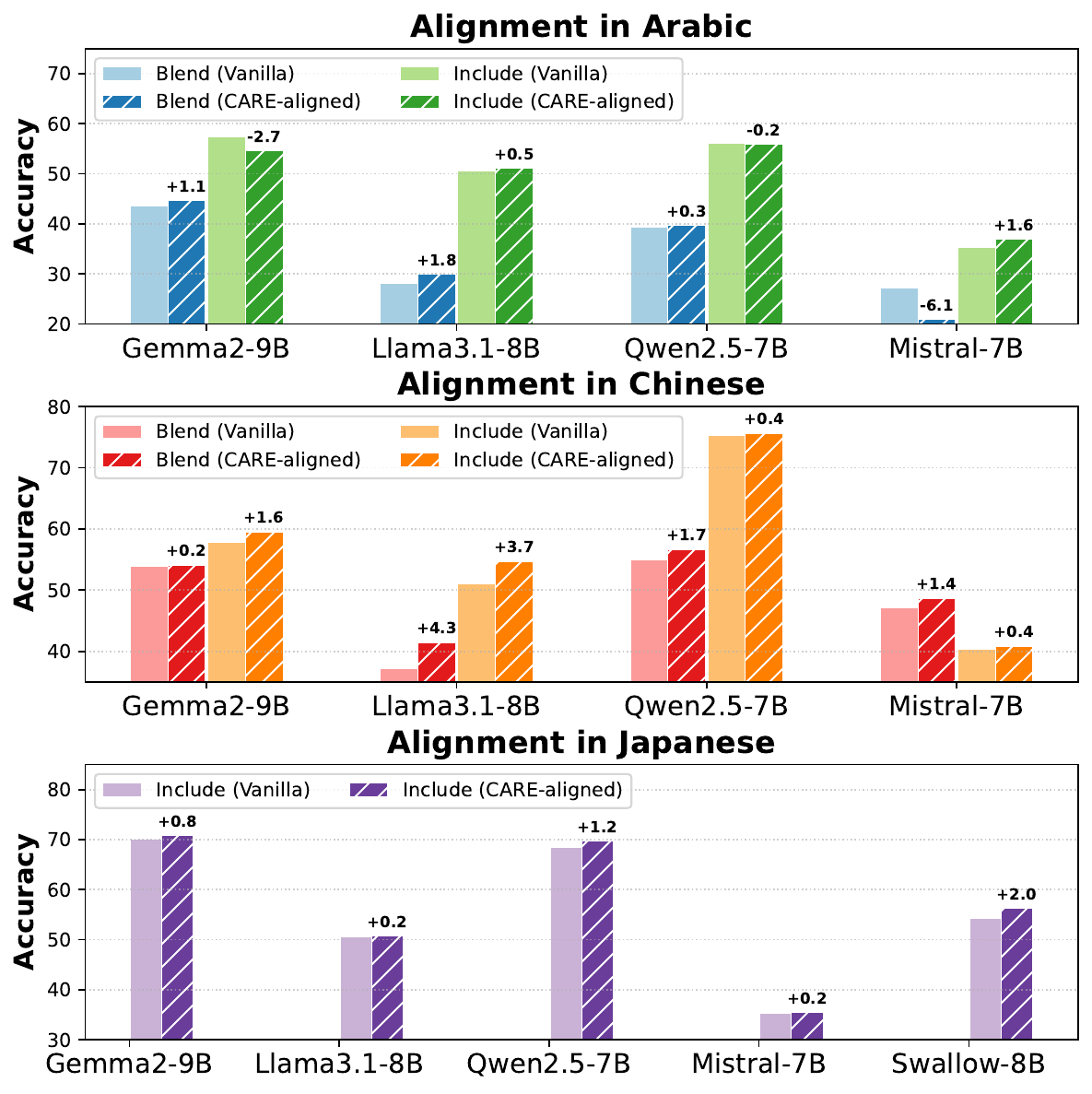}
    \vspace{-2em}
    \caption{
    Accuracy on the Chinese, Arabic, and Japanese subsets of the short answer questions in Blend~\cite{myung2024blend} and multi-choice questions in Include~\cite{romanou2025include}.
    CARE preference tuning improves out-of-domain cultural tasks for most cultures and models, except on the Arabic subset of Include. Blend does not provide a Japanese subset. 
    }
    \label{fig:ood-culture-tasks}
    \vspace{-1em}
\end{figure}

We evaluate CARE-aligned models on other cultural tasks using culture-related data that is not included in CARE. 
Specifically, we use the short answer questions within the Blend~\cite{myung2024blend} benchmark and multi-choice questions within the Include~\cite{romanou2025include} benchmark, both written in the native languages. 
From Figure~\ref{fig:ood-culture-tasks}, we observe that \textbf{\textit{preference learning with culture-specific human preference can also benefit out-of-domain cultural tasks}}, indicating the generalization capability of CARE. 
Appendix~\ref{apendix-ood-blend} extends this analysis to additional fine-tuned models, and Appendix~\ref{appendix:general-capabilities} demonstrates that overall NLP capabilities remain unaffected.

\subsection{Multi-cultural Alignment} 
\label{subsec:source-culture-analysis}

\begin{figure}[t]
    \centering
    \includegraphics[width=\linewidth]{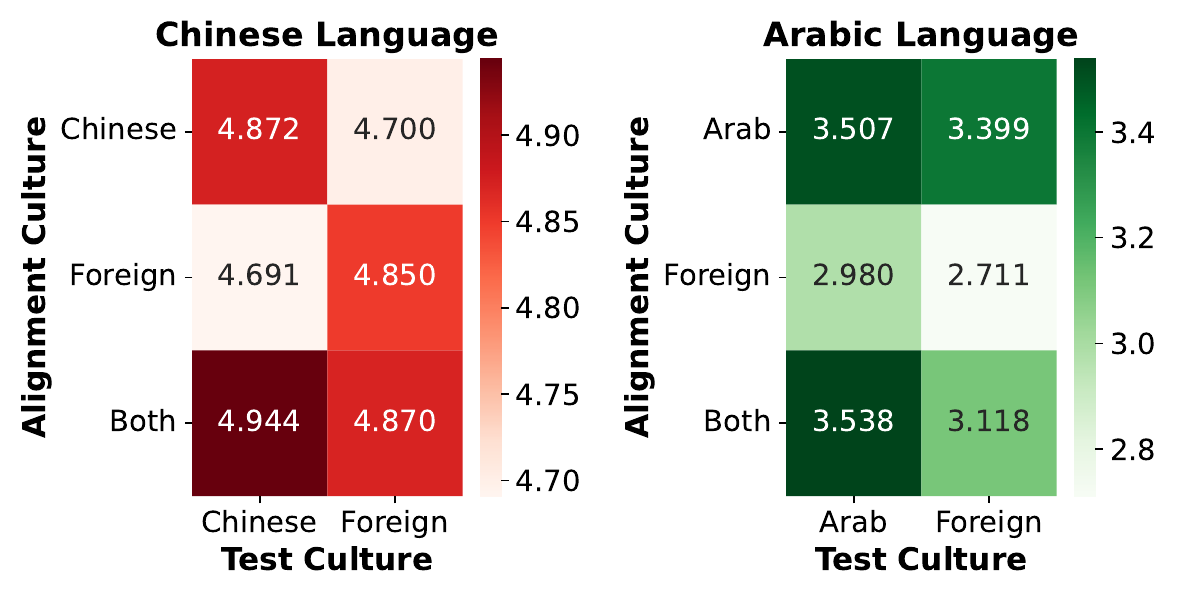}
    \vspace{-2em}
    \caption{
    Average scores of \texttt{Llama3.1-8B-Instruct} on local and foreign cultures when aligned using data from native, foreign, or mixed cultures. 
    The highest performance on local culture is achieved when mixing local and foreign samples during preference learning.
    }
    \label{fig:source-impact}
    \vspace{-.2em}
\end{figure}

We analyze the impact of incorporating data about different cultures in preference learning. To ensure a fair comparison, we select an equal number of samples for each trial. Specifically, we align \texttt{Llama3.1-8B-Instruct} with 675 Chinese and 547 Arabic pairs sampled from three different contexts: native (Chinese or Arab) culture, foreign cultures (non-Chinese or non-Arab), and mixed pairs (half native and half foreign). 
Results are shown in Figure~\ref{fig:source-impact}. 
Tuning on foreign-culture pairs alone lowers native performance; adding native pairs raises them (Chinese 4.69 $\rightarrow$ 4.87, Arab 2.98 $\rightarrow$ 3.50), and combining native and foreign pairs further improves performance, indicating that \textbf{\textit{geographically diverse preference data can further strengthen cultural awareness}}.

\begin{table}[t]
\centering
\begin{adjustbox}{width=\linewidth}
\begin{tabular}{lcccccc}
\toprule
\textbf{Model} & \textit{Entities} & \textit{Opinion} & \textit{Norms} & \textit{C. sense} & \textit{Literacy} & Avg. \\ \midrule
\multicolumn{7}{c}{\textit{Llama family}} \\ \midrule
Llama3.1-8B & 2.20 & 3.93 & 2.47 & 2.57 & 1.97 & 2.63 \\
\phantom{~~}$\hookrightarrow$ CARE Aligned\includegraphics[width=0.4cm]{fig/Hugging.jpg} & 3.20 & 4.67 & 2.80 & 3.20 & 1.67 & 3.11 \\
\phantom{~~}$\hookrightarrow$ Swallow-8B & 5.63 & 6.40 & 5.47 & 6.33 & 4.63 & 5.69 \\
\phantom{~~~~}$\hookrightarrow$ CARE Aligned\includegraphics[width=0.4cm]{fig/Hugging.jpg} & \textbf{5.77} & \textbf{6.57} & \textbf{5.60} & \textbf{6.70} & \textbf{5.40} & \textbf{6.01} \\
Llama3.3-70B & 6.17 & 6.07 & 4.77 & 6.13 & 5.97 & 5.82 \\
\phantom{~~}$\hookrightarrow$ Swallow-70B & \textbf{7.20} & \textbf{7.37} & \textbf{7.13} & \textbf{7.53} & \textbf{7.27} & \textbf{7.30} \\[1pt]
\midrule
\multicolumn{7}{c}{\textit{Gemma family}} \\ \midrule
Gemma2-9B & 4.46 & 6.50 & 5.67 & 6.50 & 2.33 & 5.09 \\
\phantom{~~}$\hookrightarrow$ CARE Aligned\includegraphics[width=0.4cm]{fig/Hugging.jpg} & 4.40 & 6.60 & 5.80 & 6.67 & 2.93 & 5.28 \\
\phantom{~~}$\hookrightarrow$ Swallow-9B & 6.07 & \textbf{7.27} & 6.90 & \textbf{7.20} & 5.10 & 6.51 \\
\phantom{~~~~}$\hookrightarrow$ CARE Aligned\includegraphics[width=0.4cm]{fig/Hugging.jpg} & \textbf{6.10} & {7.13} & \textbf{7.37} & {6.97} & \textbf{5.50} & \textbf{6.61} \\
Gemma2-27B & 5.37 & 7.07 & \textbf{6.50} & 6.47 & 3.83 & 5.85 \\
\phantom{~~}$\hookrightarrow$ Swallow-27B & \textbf{6.37} & \textbf{7.43} & {6.43} & \textbf{7.43} & \textbf{6.00} & \textbf{6.73} \\[1pt]
\midrule
\multicolumn{7}{c}{\textit{Mistral family}} \\ \midrule
Mistral-7B & 1.70 & 3.96 & 2.40 & 2.48 & 1.20 & 2.34 \\
\phantom{~~}$\hookrightarrow$ CARE Aligned\includegraphics[width=0.4cm]{fig/Hugging.jpg} & 1.83 & 3.90 & 2.43 & 2.33 & 1.23 & 2.35 \\
\phantom{~~}$\hookrightarrow$ Swallow-7B & 2.70 & \textbf{4.43} & 4.07 & 3.87 & 2.00 & 3.41 \\
\phantom{~~~~}$\hookrightarrow$ CARE Aligned\includegraphics[width=0.4cm]{fig/Hugging.jpg} & \textbf{3.13} & {4.00} & \textbf{4.50} & \textbf{4.90} & \textbf{2.53} & \textbf{3.81} \\
Mistral-8×7B & 2.10 & 4.37 & 3.23 & 3.30 & 2.30 & 3.06 \\
\phantom{~~}$\hookrightarrow$ Swallow-8×7B & \textbf{4.33} & \textbf{5.40} & \textbf{5.30} & \textbf{5.87} & \textbf{2.90} & \textbf{4.76} \\
\bottomrule
\end{tabular}
\end{adjustbox}
\caption{Average scores on CARE's Japanese test set for (i) each backbone, (ii) each backbone further tuned on CARE, (iii) its Japanese-pretrained \texttt{Swallow} variant, and (iv) that variant further tuned on CARE.}
\vspace{-1.em}
\label{tab:jp_swallow}
\end{table}

\subsection{Language-specific LLMs}



We also explore the combined benefits of both language-specific training and cultural preference tuning. In particular, we conduct DPO tuning on CARE's Japanese training split with the Swallow model variants (\texttt{Llama3.1-Swallow-8B}, \texttt{Gemma2-Swallow-9B}, and \texttt{Mistral-Swallow-7B}), which are continuously pre-trained and instruction-tuned on the Japanese corpus~\cite{Fujii:COLM2024}. Table \ref{tab:jp_swallow} shows that language-specific continual pre-training improves each base model by 1–3 points (on a 10-point scale), while subsequent preference tuning on CARE contributing an additional 0.2–0.4 points. Combined, these steps enable smaller models to outperform much larger LLMs. For example, aligned \texttt{Llama-Swallow-8B} outperforms vanilla \texttt{Llama-70B} (6.01 vs. 5.82), and aligned \texttt{Gemma-Swallow-9B} surpasses \texttt{Gemma-27B}  (6.61 vs. 5.85). The results echo our findings in \S \ref{subsec:main-res}: cultural alignment is consistently helpful, even when base models have enough language proficiency.

\begin{table*}[t]
\centering
\begin{adjustbox}{width=\linewidth}
\tiny
\setlength{\tabcolsep}{0.5pt}
\begin{tabularx}{\linewidth}{@{}l *{6}{>{\centering\arraybackslash}X} *{6}{>{\centering\arraybackslash}X} *{6}{>{\centering\arraybackslash}X} @{}}
\toprule
 & \multicolumn{6}{c}{\textbf{Chinese}} & \multicolumn{6}{c}{\textbf{Arabic}} & \multicolumn{6}{c}{\textbf{Japanese}} \\ \cmidrule(l){2-7} \cmidrule(l){8-13} \cmidrule(l){14-19} 
\textbf{Model} & \textit{Entities} & \textit{Opinion} & \textit{Norms} & \textit{C. sense} & \textit{Literacy} & \textit{Average} & \textit{Entities} & \textit{Opinion} & \textit{Norms} & \textit{C. sense} & \textit{Literacy} & \textit{Average} & \textit{Entities} & \textit{Opinion} & \textit{Norms} & \textit{C. sense} & \textit{Literacy} & \textit{Average} \\ \midrule
 & 50\% & 53\% & {96\%} & {100\%} & 56\% & \multicolumn{1}{c}{71\%} & 70\% & 90\% & {100\%} & {90\%} & 20\% & \multicolumn{1}{c}{74\%} & 17\% & {87\%} & {87\%} & {93\%} & 23\% & 63\% \\
\multirow{-2}{*}{{Human Awareness (\%)}} & \multicolumn{6}{l}{\includegraphics[width=0.24\textwidth]{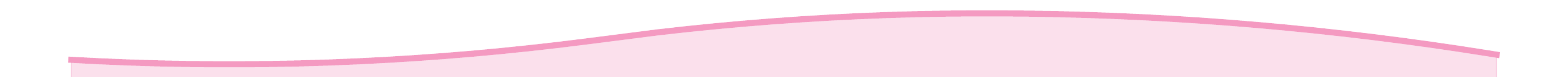}} & \multicolumn{6}{l}{\includegraphics[width=0.24\textwidth]{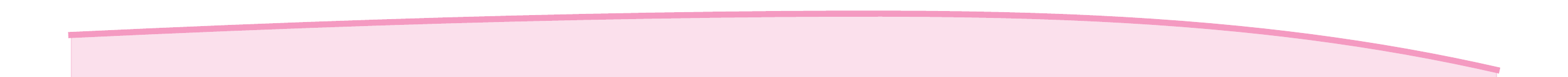}} & \multicolumn{6}{l}{\includegraphics[width=0.24\textwidth]{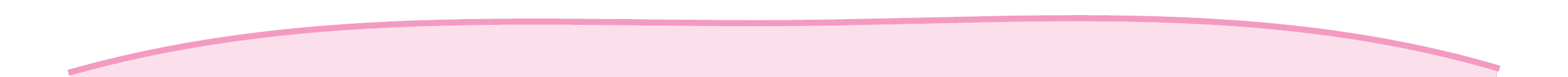}}  \\ 
\hdashline[1pt/1pt] 
\\ [-5pt] 
 & 5.63 & 4.53 & 2.69 & 2.90 & 6.53 & 4.46 & 5.87 & 3.11 & 2.31 & 2.96 & 7.19 & 4.30 & 3.43 & 2.77 & 2.63 & 3.00 & 2.97 & 2.96 \\
\multirow{-2}{*}{{Search Engine}} & \multicolumn{6}{l}{\includegraphics[width=0.24\textwidth]{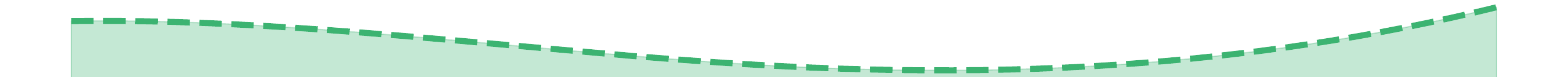}} & \multicolumn{6}{l}{\includegraphics[width=0.24\textwidth]{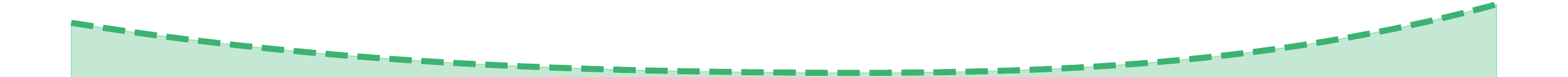}} & \multicolumn{6}{l}{\includegraphics[width=0.24\textwidth]{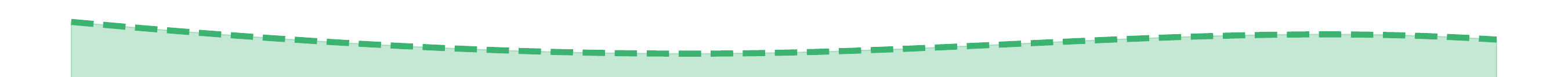}}  \\ 
\hdashline[1pt/1pt] 
\\ [-5pt] 
& 5.64 & {8.33} & 8.13 & 6.96 & 5.73 & 6.97 & 6.24 & {7.76} & {7.76} & 6.53 & 3.50 & 6.44 & 5.37 & 7.07 & 6.50 & 6.47 & 3.83 & 5.85 \\
\multirow{-2}{*}{\includegraphics[width=0.25cm] {fig/google-logo.png} Gemma2-27B}& \multicolumn{6}{l}{\includegraphics[width=0.24\textwidth]{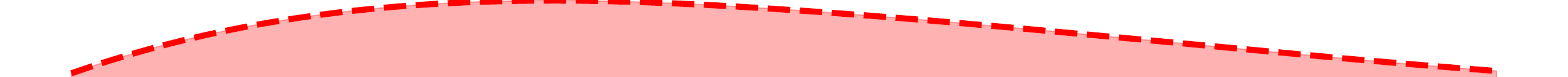}} & \multicolumn{6}{l}{\includegraphics[width=0.24\textwidth]{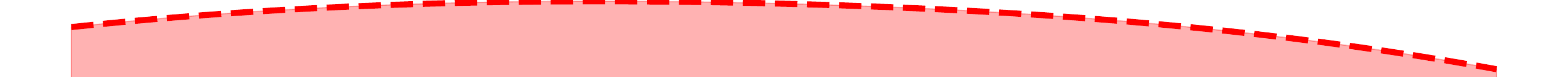}} & \multicolumn{6}{l}{\includegraphics[width=0.24\textwidth]{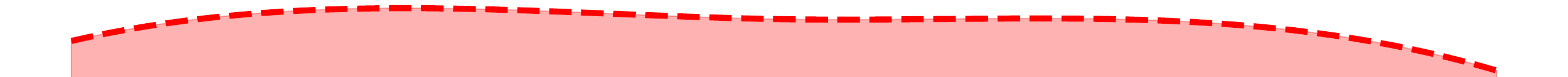}} \\
 & 5.82 & 7.33 & {7.27} & 7.06 & 5.93 & 6.69 & 5.79 & 6.93 & {7.13} & 5.86 & 3.86 & 5.93 & 6.17 & 6.07 & 4.77 & 6.13 & 5.97 & 5.82 \\
\multirow{-2}{*}{\includegraphics[width=0.25cm] {fig/meta-logo.png} Llama3.3-70B} & \multicolumn{6}{l}{\includegraphics[width=0.24\textwidth]{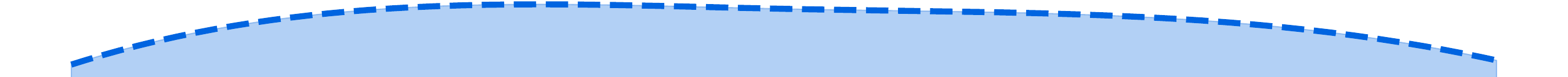}} & \multicolumn{6}{l}{\includegraphics[width=0.24\textwidth]{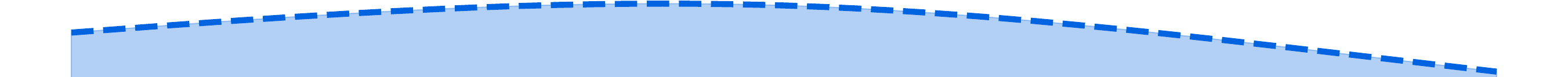}} & \multicolumn{6}{l}{\includegraphics[width=0.24\textwidth]{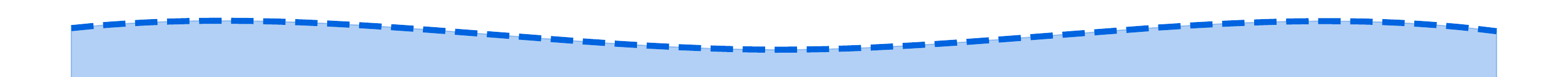}}  \\
 & \underline{8.42} & {8.96} & \textbf{9.06} & 8.09 & \underline{8.53} & {8.61} & {7.17} & 7.56 & {7.90} & {7.20} & {5.25} & {7.04} & 5.37 & 8.10 & 6.27 & 7.57 & 5.47 & 6.55 \\
\multirow{-2}{*}{\includegraphics[width=0.25cm]{fig/qwen-logo.png} Qwen2.5-72B} & \multicolumn{6}{l}{\includegraphics[width=0.24\textwidth]{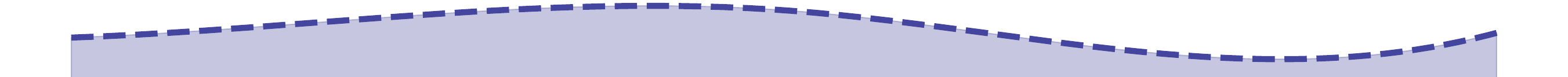}} & \multicolumn{6}{l}{\includegraphics[width=0.24\textwidth]{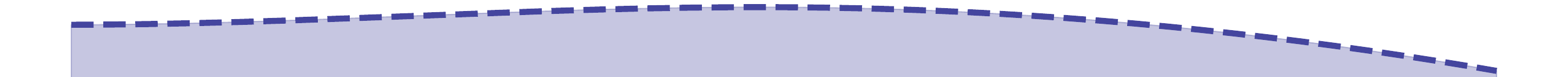}} & \multicolumn{6}{l}{\includegraphics[width=0.24\textwidth]{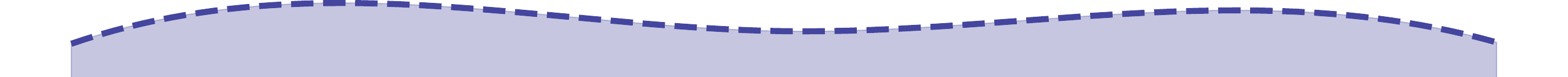}} \\
 & 7.57 & {8.96} & 8.24 & {8.58} & 6.66 & 8.01 & 6.34 & {7.63} & 7.70 & 7.13 & 4.80 & 6.72 & 4.83 & 7.63 & 7.10 & 7.17 & 5.23 & 6.39 \\
\multirow{-2}{*}{\includegraphics[width=0.25cm] {fig/mistral-logo.png} Mistral-Large} & \multicolumn{6}{l}{\includegraphics[width=0.24\textwidth]{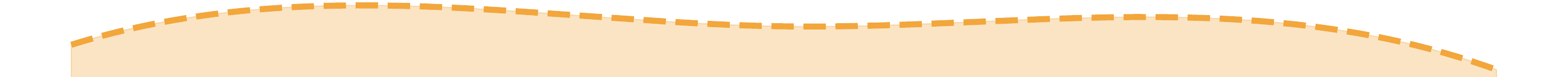}} & \multicolumn{6}{l}{\includegraphics[width=0.24\textwidth]{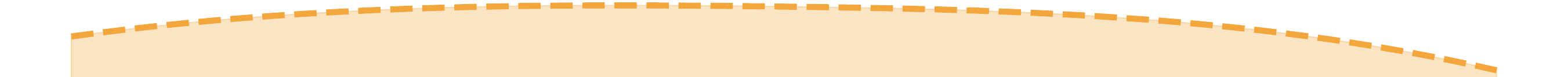}} & \multicolumn{6}{l}{\includegraphics[width=0.24\textwidth]{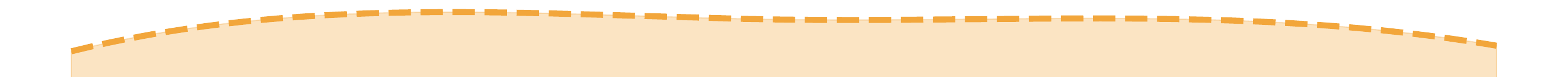}} \\
 & \textbf{8.96} & \underline{9.30} & \underline{9.03} & \textbf{9.23} & \textbf{9.30} & \textbf{9.16} & \textbf{7.90} & \underline{8.16} & \underline{8.70} & \textbf{8.00} & \textbf{7.39} & \underline{8.04} & \underline{7.67} & \textbf{8.87} & \textbf{8.90} & \textbf{9.07} & \underline{7.77} & \textbf{8.45} \\ 
\multirow{-2}{*}{\includegraphics[width=0.25cm] {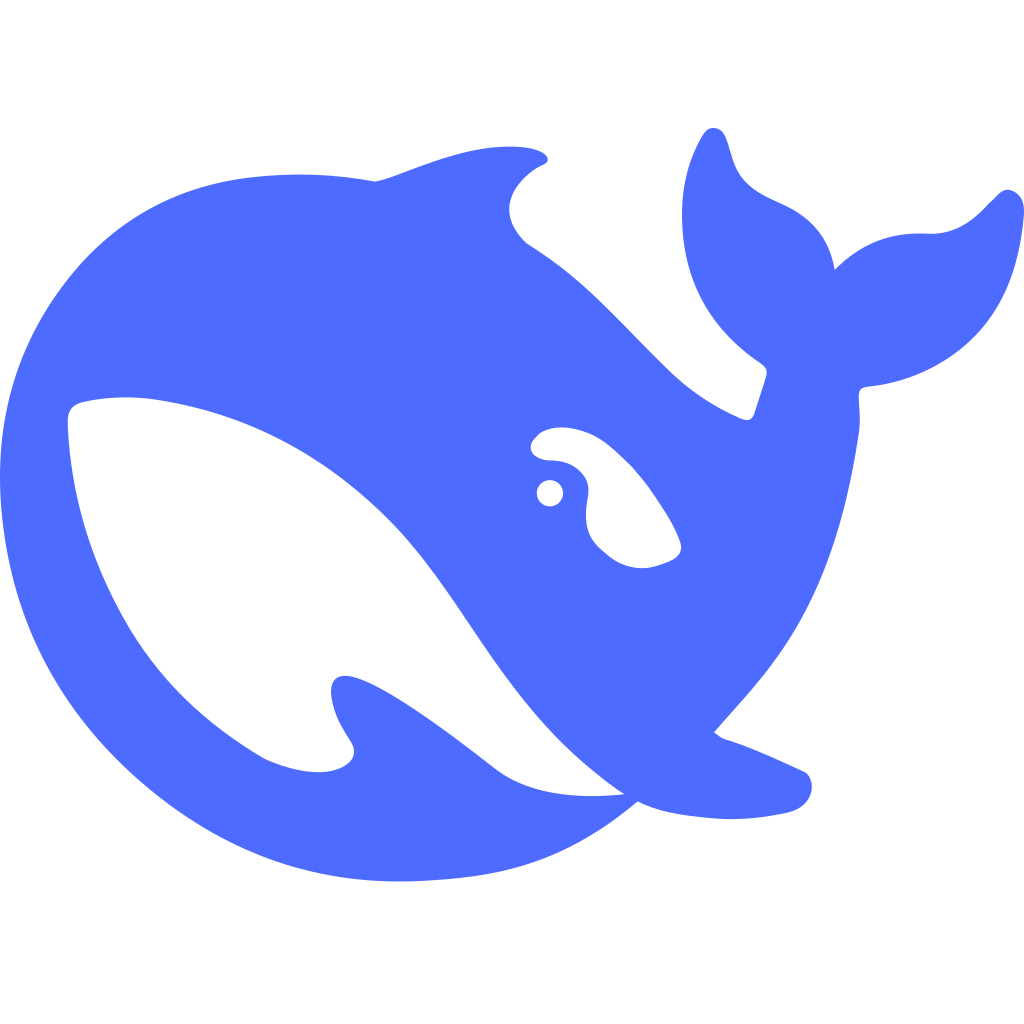} Deepseek-v3} & \multicolumn{6}{l}{\includegraphics[width=0.24\textwidth]{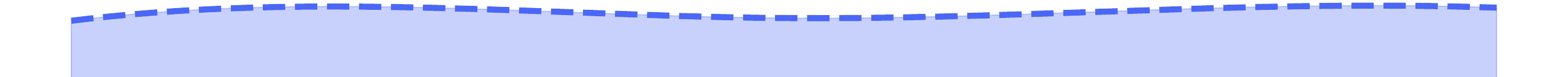}} & \multicolumn{6}{l}{\includegraphics[width=0.24\textwidth]{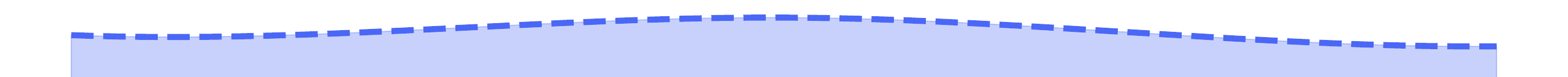}} & \multicolumn{6}{l}{\includegraphics[width=0.24\textwidth]{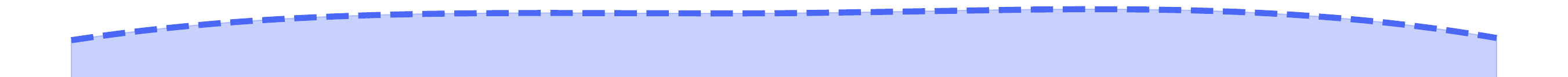}}  \\ 
& {8.39} & \textbf{9.43} & {8.72} & \underline{9.00} & {8.06} & \underline{8.73} & \underline{7.51} & \textbf{8.66} & \textbf{8.80} & \underline{7.83} & \underline{7.37} & \textbf{8.05} & \textbf{8.37} & \underline{8.37} & \underline{7.73} & \underline{8.60} & \textbf{8.30} & \underline{8.27} \\ 
\multirow{-2}{*}{\includegraphics[width=0.25cm] {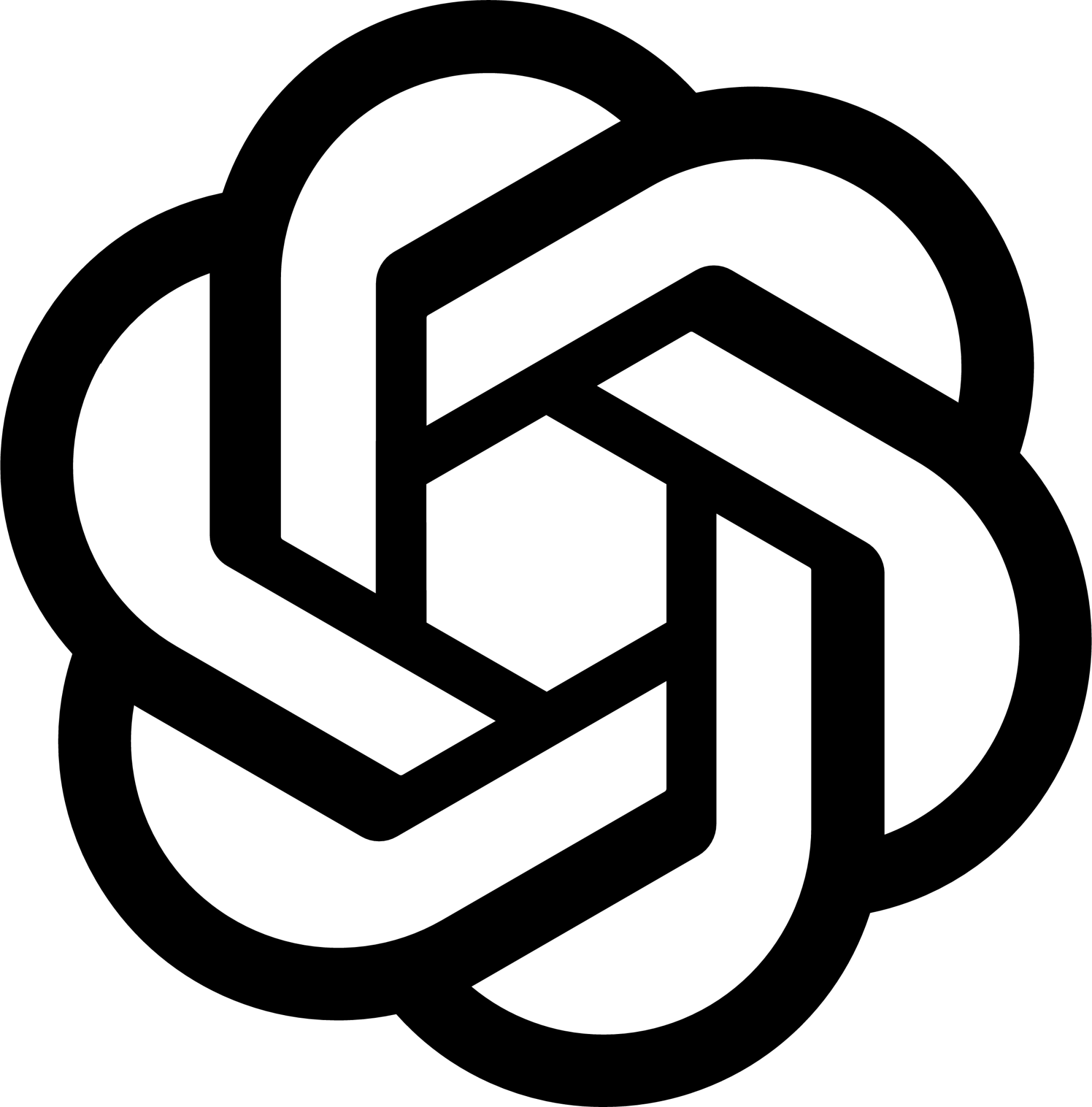} GPT-4o} & \multicolumn{6}{l}{\includegraphics[width=0.24\textwidth]{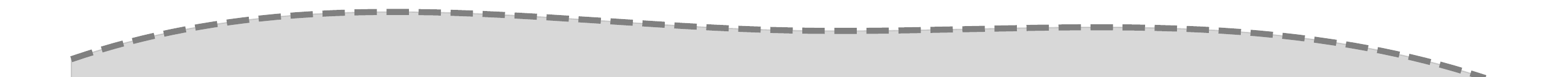}} & \multicolumn{6}{l}{\includegraphics[width=0.24\textwidth]{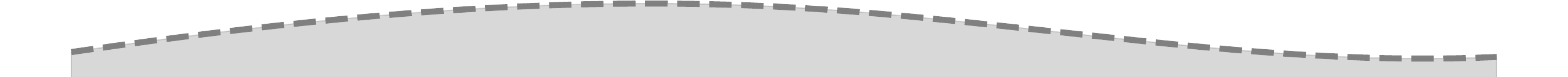}} & \multicolumn{6}{l}{\includegraphics[width=0.24\textwidth]{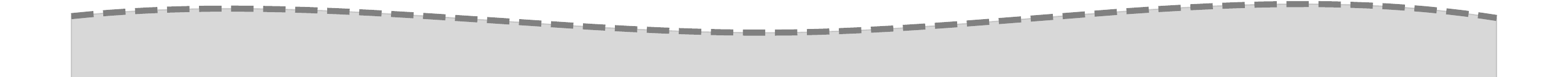}}  \\
\bottomrule
\end{tabularx} 
\end{adjustbox}
\caption{
Average scores on CARE of larger LMs and web content, evaluated by the judge LM across all samples (1: \textit{poor} $\rightarrow$ 10: \textit{excellent}). 
``{Human Awareness (\%)}'' indicates the percentage of questions for which locals know the correct answer. 
Chinese-developed \texttt{Deepseek-v3} and \texttt{Qwen2.5} excel on Chinese entities, norms, and literacy, while \texttt{GPT-4o} and \texttt{Mistral-Large} are comparably good at opinions and commonsense.
For social norms and cultural commonsense, where local humans show high confidence, Google search often returns fewer relevant answers.
}
\vspace{-1em}
\label{tab:larger-lm-human}
\end{table*}

\section{Prompting LMs for Cultural Awareness}
\label{sec:prompting_evaluation}




We evaluate larger LMs and compare to human familiarity and web-based sources (\S\ref{subsec:largerlm-compare-human}), as well as querying in different languages (\S\ref{subsec:presence-language}).


\subsection{Main Results}
\label{subsec:largerlm-compare-human}

We follow the evaluation setup in \S\ref{subsec:cultural-performance} and assess the larger ($>$ 25B parameters) versions of the evaluated LMs with zero-shot prompting. Table~\ref{tab:larger-lm-human} shows the results. Larger models are better for all three cultures, with closed-source ones like \texttt{GPT-4o} and \texttt{Deepseek-v3} leading the overall performance. The more Chinese-centric model \texttt{Deepseek-v3} and \texttt{Qwen2.5-72B} excels in the knowledge about cultural entities (8.96 and 8.42), social norms (9.03 and 9.06), and literacy (9.03 and 8.53) for Chinese; meanwhile, \texttt{GPT-4o} and \texttt{Mistral-Large} are comparably good if not better at offering opinions (9.43 and 8.96) and answering commonsense questions (9.00 and 8.58) about Chinese culture. This is interesting and likely because \textit{\textbf{cultural commonsense knowledge is often unstated in the native language, while more balanced and thoughtful opinions may be expressed in foreign languages}}. We further test this hypothesis in \S \ref{subsec:presence-language}. 


\begin{figure*}[t]
    \centering

    \begin{subfigure}[t]{0.48\textwidth}
        \centering
        \includegraphics[width=\linewidth]{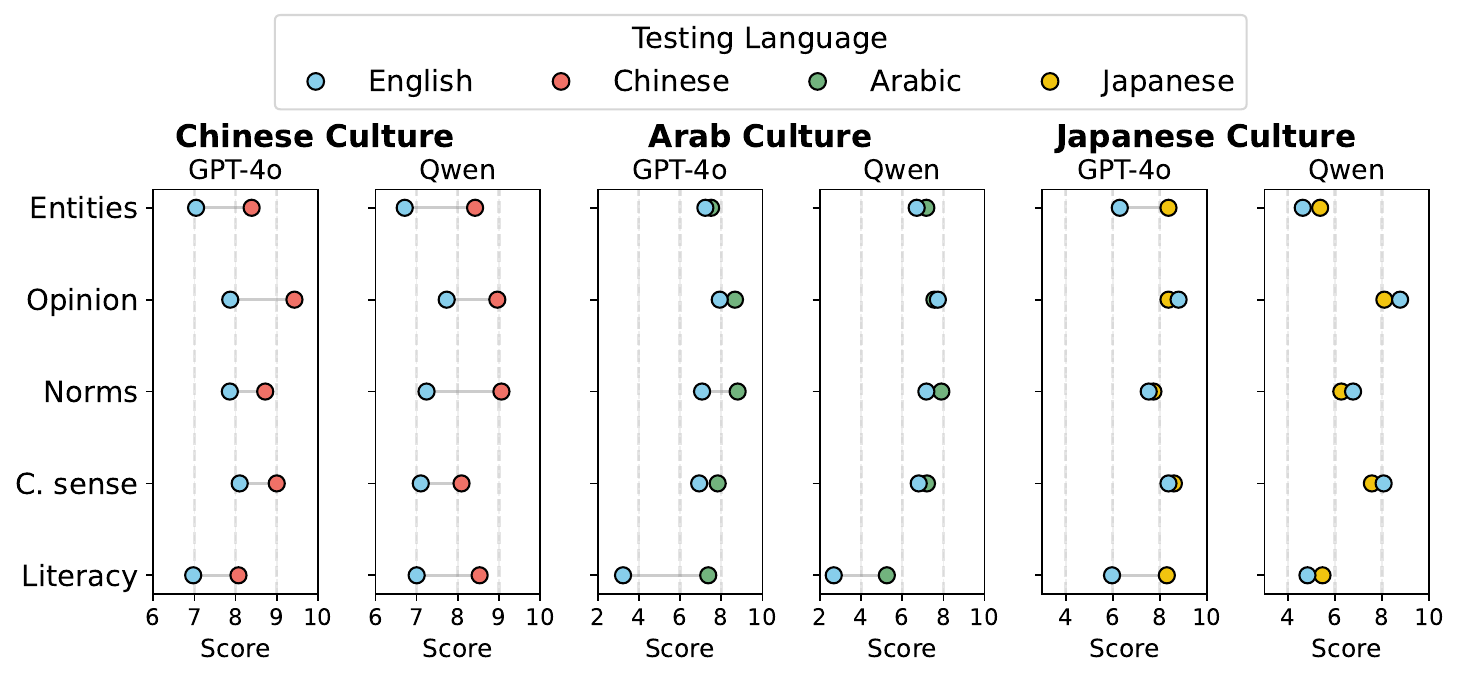}
        \caption{
        Average scores achieved by \texttt{Qwen2.5-72B-Instruct} and \texttt{GPT-4o} when prompted in native languages versus English. 
        }
        \label{fig:testing-lang}
    \end{subfigure}
    \hfill
    \begin{subfigure}[t]{0.48\textwidth}
        \centering
        \includegraphics[width=\linewidth]{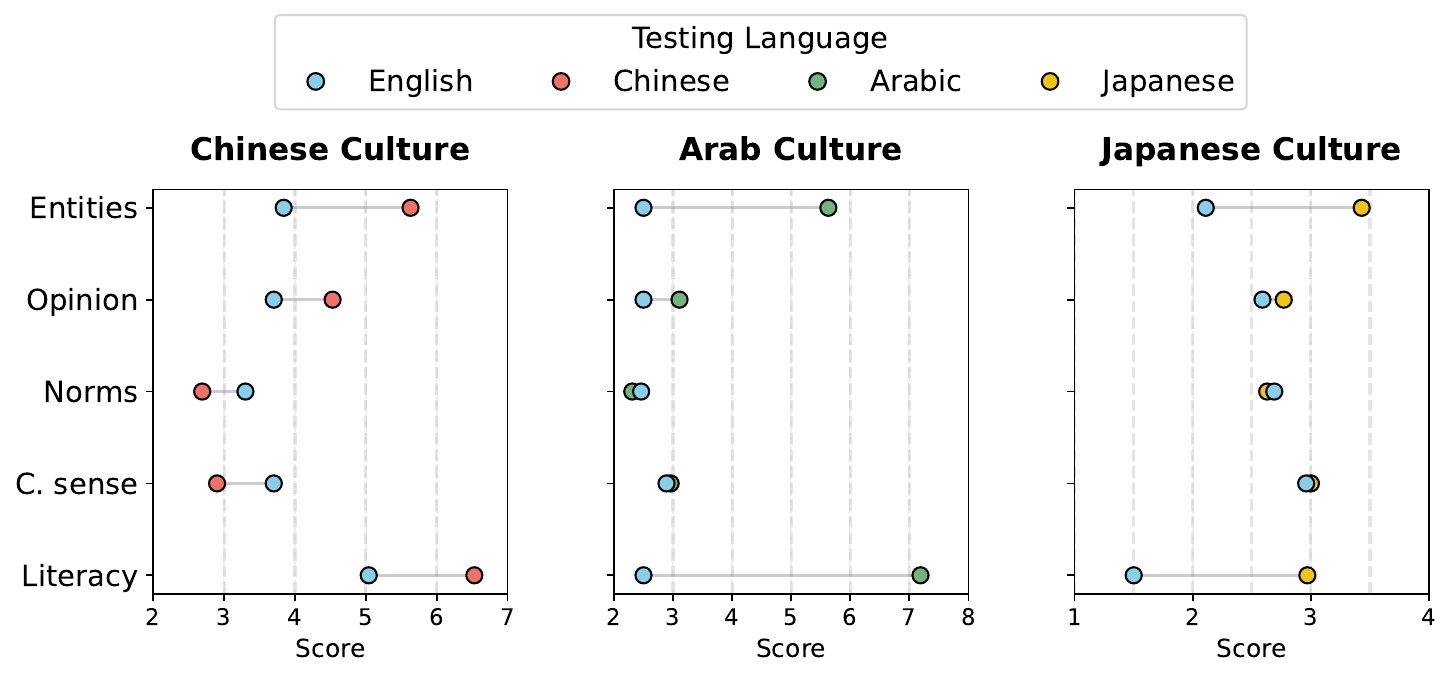}
        \caption{
        Average scores of retrieved webpage main content when searching CARE questions in native languages versus English.
        }
        \label{fig:search-analysis}
    \end{subfigure}
    \caption{
    Performance comparison of cultural awareness in different languages. 
    Comparison between (a) LM performance and (b) retrieved webpage content, showing that native-language questions capture cultural nuances for literacy, opinion, and entities. However, the gaps narrow or sometimes reverse for cultural norms and commonsense.
    }
    \label{fig:culture-joined}
    \vspace{-1em}
\end{figure*}

\paragraph{Human Awareness.} To assess human familiarity with the culture-specific questions, we ask native-speaking annotators whether they knew the correct answers without looking them up when presented with questions from CARE. Table~\ref{tab:larger-lm-human} shows the percentages of questions humans could answer immediately with confidence, which are very high for social norms and commonsense knowledge. 


\paragraph{Search Engine.} To further examine the coverage of cultural knowledge in real-world text, for each CARE question we used the Google Programmable Search Engine\footnote{\url{https://programmablesearchengine.google.com}} to retrieve the top-10 results (URLs and associated snippets), fetched each HTML page, extracted the paragraph containing the snippet, scored all paragraphs with the same LM-as-a-judge framework, and recorded the maximum as the question's score. Results are shown in Table~\ref{tab:larger-lm-human}. Since many retrieved paragraphs only partially address the questions, their scores are generally below LMs' scores.

\subsection{Cross-lingual Analysis}
\label{subsec:presence-language}

We translate all questions and answers in CARE that are written in Chinese, Arabic, and Japanese into English to enable cross-lingual evaluation. Figure~\ref{fig:testing-lang} shows that LMs generally perform better on culture-specific questions when prompted in the native languages than in English, but more notably for Chinese than Arabic and Japanese. This advantage is also especially obvious for the literacy category, while the gap narrows in categories such as cultural commonsense. 

We also examine how language choice impacts search engine results. As shown in Figure~\ref{fig:search-analysis}, cultural information on entities, opinions, and literacy tends to be of higher quality when retrieved in native languages. Interestingly, this trend reverses for questions related to cultural commonsense and social norms, such as \textit{``In China, does the ticket time indicate when the feature movie starts, or are there trailers played before the main feature?''}, where English searches yield higher-quality content. This aligns with the potential documentation gap mentioned in \S\ref{subsec:largerlm-compare-human}, where native speakers often assume everyday cultural knowledge is implicit, but non-native speakers are more likely to seek out information or document it.

\section{Conclusion}

We introduce CARE, a multilingual, human-written resource comprising 3,490 culture-specific questions about Chinese, Arab, and Japanese cultures, along with human-rated responses. 
Through extensive experiments, we investigate that a small amount of high-quality cultural preferences can improve LMs via preference optimization, across various model families and scales. 
Our analyses reveal gaps in cultural awareness among LMs, native speakers, and search engines, especially regarding implicit cultural commonsense. 
By releasing CARE, we hope to foster the development of more culturally adaptive LMs.

\section*{Limitations} 

During data collection, we intentionally excluded highly sensitive and controversial topics (e.g., ``\textit{Are security concerns about the niqab exaggerated or justified?}''), as these tend to elicit divergent personal views. Our work addresses generally shared topics within each cultural group (e.g., ``\textit{Do Chinese people like stinky tofu?}'' in the Opinion category of CARE), allowing us to capture collective preferences that reflect broader cultural consensus \cite{bakker2022fine} with a neutral point of view (e.g., ``\textit{Some Chinese like stinky tofu. It is a traditional food made by $\ldots$ It is most popular in $\ldots$ regions. At the same time, many Chinese people don't like it $\ldots$}'' ). This approach facilitates more objective and replicable evaluation, as evidenced by the high inter-annotator agreement reported in \S\ref{sub:care_data}, thus supports more reliable algorithmic comparisons, which are the primary focus of this paper. The scope of our study is complementary to other work ~\cite{chiu2024culturalteaming,kirk2024prism,huang2025valueswilddiscoveringanalyzing} that explored diverse values and subjective viewpoints \cite{zhong-etal-2021-wikibias-detecting}.


\begin{CJK*}{UTF8}{gbsn}
We also note that LMs may occasionally respond to culture-specific questions in a stereotypical or biased manner, stemming from a lack of cultural understanding. For instance, when asked ``\textit{In China, what does the term ``laowai'' mean when referring to foreigners?}'', LMs often incorrectly interpret the term as disrespectful. In reality, ``\textit{laowai} (老外)'' is typically a neutral descriptor in Chinese, used to denote foreigners without negative connotations. While CARE includes examples that clarify such misunderstandings through human-written responses, future studies in cultural red-teaming can further investigate and address these failure cases.
\end{CJK*}
\\
\\

\section*{Ethics Statement} 

In this study, we employed in-house annotators to collect cultural samples and provide preference judgments in CARE. 
The annotators for the Chinese and Arabic samples were university-level students fluent in the respective languages. Japanese annotators were Japanese workers with university degrees and cross-cultural experience abroad. Each Chinese and Arabic annotator was paid at \$18 per hour, exceeding the U.S. federal minimum wage. 
We ensured that no personal data was collected from the annotators and emphasized that participation was voluntary. 
We also informed the annotators that the collected data would be used to enhance the cultural awareness of LMs. 
The culture-related questions and reference responses in CARE are either from the existing works or human-written sentences provided by our annotators.

\section*{Acknowledgments}

The authors would like to thank Chao Jiang, Jad Matthew Bardawil, Nour Allah El Senary, Ruohao Guo, Xiaofeng Wu, Yuming Pan, and Zirui Shao for their assistance with data annotation; and Yao Dou, Jonathan Zheng, Xiaofeng Wu as well as three anonymous reviewers for their helpful feedback for their helpful feedback. 
We would also like to thank Microsoft's Azure Accelerate Foundation Models Research Program and NVIDIA's Academic Grant Program for providing computational resources to support this work.
This work was funded in part through a Sony Faculty Innovation Award, and by the NSF under grant number IIS-2144493, IIS-2052498 and SMA-2418946.
Any opinions, findings, and conclusions or recommendations expressed in this material are those of the author(s) and do not necessarily reflect the views of the National Science Foundation.

\bibliography{references}

\begin{thebibliography}{89}
\providecommand{\natexlab}[1]{#1}

\bibitem[{2A2I(2024)}]{arabic-dpo-pairs}
2A2I. 2024.
\newblock \href {https://huggingface.co/datasets/2A2I/argilla-dpo-mix-7k-arabic} {{2A2I/argilla-dpo-mix-7k-arabic}}.

\bibitem[{Abdulhai et~al.(2023)Abdulhai, Serapio-Garcia, Crepy, Valter, Canny, and Jaques}]{abdulhai2023moral}
Marwa Abdulhai, Gregory Serapio-Garcia, Cl{\'e}ment Crepy, Daria Valter, John Canny, and Natasha Jaques. 2023.
\newblock Moral foundations of large language models.
\newblock \emph{arXiv preprint arXiv:2310.15337}.

\bibitem[{Adilazuarda et~al.(2024)Adilazuarda, Mukherjee, Lavania, Singh, Aji, O'Neill, Modi, and Choudhury}]{adilazuarda2024towards}
Muhammad~Farid Adilazuarda, Sagnik Mukherjee, Pradhyumna Lavania, Siddhant Singh, Alham~Fikri Aji, Jacki O'Neill, Ashutosh Modi, and Monojit Choudhury. 2024.
\newblock Towards measuring and modeling" culture" in llms: A survey.
\newblock \emph{arXiv preprint arXiv:2403.15412}.

\bibitem[{Ahmadian et~al.(2024)Ahmadian, Ermis, Goldfarb-Tarrant, Kreutzer, Fadaee, Hooker et~al.}]{ahmadian2024multilingual}
Arash Ahmadian, Beyza Ermis, Seraphina Goldfarb-Tarrant, Julia Kreutzer, Marzieh Fadaee, Sara Hooker, et~al. 2024.
\newblock The multilingual alignment prism: Aligning global and local preferences to reduce harm.
\newblock \emph{arXiv preprint arXiv:2406.18682}.

\bibitem[{AlKhamissi et~al.(2024)AlKhamissi, ElNokrashy, AlKhamissi, and Diab}]{alkhamissi2024investigating}
Badr AlKhamissi, Muhammad ElNokrashy, Mai AlKhamissi, and Mona Diab. 2024.
\newblock Investigating cultural alignment of large language models.
\newblock \emph{arXiv preprint arXiv:2402.13231}.

\bibitem[{Alwajih et~al.(2025)Alwajih, Mekki, Magdy, Elmadany, Nacar, Nagoudi, Abdel-Salam, Atwany, Nafea, Yahya et~al.}]{alwajih2025palm}
Fakhraddin Alwajih, Abdellah~El Mekki, Samar~Mohamed Magdy, Abdelrahim~A Elmadany, Omer Nacar, El~Moatez~Billah Nagoudi, Reem Abdel-Salam, Hanin Atwany, Youssef Nafea, Abdulfattah~Mohammed Yahya, et~al. 2025.
\newblock Palm: A culturally inclusive and linguistically diverse dataset for arabic llms.
\newblock \emph{arXiv preprint arXiv:2503.00151}.

\bibitem[{Alyafeai et~al.(2024)Alyafeai, Almubarak, Ashraf, Alnuhait, Alshahrani, Abdulrahman, Ahmed, Gawah, Saleh, Ghaleb et~al.}]{alyafeai2024cidar}
Zaid Alyafeai, Khalid Almubarak, Ahmed Ashraf, Deema Alnuhait, Saied Alshahrani, Gubran~AQ Abdulrahman, Gamil Ahmed, Qais Gawah, Zead Saleh, Mustafa Ghaleb, et~al. 2024.
\newblock {CIDAR}: Culturally relevant instruction dataset for arabic.
\newblock \emph{arXiv preprint arXiv:2402.03177}.

\bibitem[{Arora et~al.(2024)Arora, Karpinska, Chen, Bhattacharjee, Iyyer, and Choi}]{arora2024calmqa}
Shane Arora, Marzena Karpinska, Hung-Ting Chen, Ipsita Bhattacharjee, Mohit Iyyer, and Eunsol Choi. 2024.
\newblock {CaLMQA}: Exploring culturally specific long-form question answering across 23 languages.
\newblock \emph{arXiv preprint arXiv:2406.17761}.

\bibitem[{Bakker et~al.(2022)Bakker, Chadwick, Sheahan, Tessler, Campbell-Gillingham, Balaguer, McAleese, Glaese, Aslanides, Botvinick et~al.}]{bakker2022fine}
Michiel~A Bakker, Martin~J Chadwick, Hannah Sheahan, Michael~Henry Tessler, Lucy Campbell-Gillingham, Jan Balaguer, Nat McAleese, Amelia Glaese, John Aslanides, Matthew Botvinick, et~al. 2022.
\newblock Fine-tuning language models to find agreement among humans with diverse preferences.
\newblock In \emph{Advances in Neural Information Processing Systems}.

\bibitem[{Bartolome et~al.(2023)Bartolome, Martin, and Vila}]{notus2023}
Alvaro Bartolome, Gabriel Martin, and Daniel Vila. 2023.
\newblock Notus.
\newblock \url{https://github.com/argilla-io/notus}.

\bibitem[{Cao et~al.(2023)Cao, Zhou, Lee, Cabello, Chen, and Hershcovich}]{cao2023assessing}
Yong Cao, Li~Zhou, Seolhwa Lee, Laura Cabello, Min Chen, and Daniel Hershcovich. 2023.
\newblock Assessing cross-cultural alignment between {ChatGPT} and human societies: An empirical study.
\newblock \emph{arXiv preprint arXiv:2303.17466}.

\bibitem[{Chen et~al.(2023)Chen, Yan, Liang, Jiang, Wu, Yu, Chen, Chen, Zhang, Jianquan, Xiang, and Wang}]{Chen_MultilingualSIFT_Multilingual_Supervised_2023}
Zhihong Chen, Shuo Yan, Juhao Liang, Feng Jiang, Xiangbo Wu, Fei Yu, Guiming~Hardy Chen, Junying Chen, Hongbo Zhang, Li~Jianquan, Wan Xiang, and Benyou Wang. 2023.
\newblock \href {https://github.com/FreedomIntelligence/MultilingualSIFT.git} {{MultilingualSIFT: Multilingual Supervised Instruction Fine-tuning}}.

\bibitem[{Chiu et~al.(2024{\natexlab{a}})Chiu, Jiang, Antoniak, Park, Li, Bhatia, Ravi, Tsvetkov, Shwartz, and Choi}]{chiu2024culturalteaming}
Yu~Ying Chiu, Liwei Jiang, Maria Antoniak, Chan~Young Park, Shuyue~Stella Li, Mehar Bhatia, Sahithya Ravi, Yulia Tsvetkov, Vered Shwartz, and Yejin Choi. 2024{\natexlab{a}}.
\newblock {CulturalTeaming}: {AI}-assisted interactive red-teaming for challenging {LLMs}'(lack of) multicultural knowledge.
\newblock \emph{arXiv preprint arXiv:2404.06664}.

\bibitem[{Chiu et~al.(2024{\natexlab{b}})Chiu, Jiang, Lin, Park, Li, Ravi, Bhatia, Antoniak, Tsvetkov, Shwartz et~al.}]{chiu2024culturalbench}
Yu~Ying Chiu, Liwei Jiang, Bill~Yuchen Lin, Chan~Young Park, Shuyue~Stella Li, Sahithya Ravi, Mehar Bhatia, Maria Antoniak, Yulia Tsvetkov, Vered Shwartz, et~al. 2024{\natexlab{b}}.
\newblock {CulturalBench}: a robust, diverse and challenging benchmark on measuring the (lack of) cultural knowledge of {LLMs}.
\newblock \emph{arXiv preprint arXiv:2410.02677}.

\bibitem[{Choenni et~al.(2024)Choenni, Lauscher, and Shutova}]{choenni-etal-2024-echoes}
Rochelle Choenni, Anne Lauscher, and Ekaterina Shutova. 2024.
\newblock \href {https://aclanthology.org/2024.acl-long.803/} {The echoes of multilinguality: Tracing cultural value shifts during language model fine-tuning}.
\newblock In \emph{Proceedings of the 62nd Annual Meeting of the Association for Computational Linguistics (Volume 1: Long Papers)}, Bangkok, Thailand. Association for Computational Linguistics.

\bibitem[{Costa-Juss{\`a} et~al.(2022)Costa-Juss{\`a}, Cross, {\c{C}}elebi, Elbayad, Heafield, Heffernan, Kalbassi, Lam, Licht, Maillard et~al.}]{costa2022no}
Marta~R Costa-Juss{\`a}, James Cross, Onur {\c{C}}elebi, Maha Elbayad, Kenneth Heafield, Kevin Heffernan, Elahe Kalbassi, Janice Lam, Daniel Licht, Jean Maillard, et~al. 2022.
\newblock No language left behind: Scaling human-centered machine translation.
\newblock \emph{arXiv preprint arXiv:2207.04672}.

\bibitem[{Cui et~al.(2023{\natexlab{a}})Cui, Yuan, Ding, Yao, Zhu, Ni, Xie, Liu, and Sun}]{cui2023ultrafeedback}
Ganqu Cui, Lifan Yuan, Ning Ding, Guanming Yao, Wei Zhu, Yuan Ni, Guotong Xie, Zhiyuan Liu, and Maosong Sun. 2023{\natexlab{a}}.
\newblock \href {https://arxiv.org/abs/2310.01377} {Ultrafeedback: Boosting language models with high-quality feedback}.
\newblock \emph{Preprint}, arXiv:2310.01377.

\bibitem[{Cui et~al.(2023{\natexlab{b}})Cui, Yang, and Yao}]{chinese-llama-alpaca}
Yiming Cui, Ziqing Yang, and Xin Yao. 2023{\natexlab{b}}.
\newblock \href {https://arxiv.org/abs/2304.08177} {Efficient and effective text encoding for chinese llama and alpaca}.
\newblock \emph{arXiv preprint arXiv:2304.08177}.

\bibitem[{Dubey et~al.(2024)Dubey, Jauhri, Pandey, Kadian, Al-Dahle, Letman, Mathur, Schelten, Yang, Fan et~al.}]{dubey2024llama}
Abhimanyu Dubey, Abhinav Jauhri, Abhinav Pandey, Abhishek Kadian, Ahmad Al-Dahle, Aiesha Letman, Akhil Mathur, Alan Schelten, Amy Yang, Angela Fan, et~al. 2024.
\newblock The llama 3 herd of models.
\newblock \emph{arXiv preprint arXiv:2407.21783}.

\bibitem[{Ethayarajh et~al.(2024)Ethayarajh, Xu, Muennighoff, Jurafsky, and Kiela}]{ethayarajh2024kto}
Kawin Ethayarajh, Winnie Xu, Niklas Muennighoff, Dan Jurafsky, and Douwe Kiela. 2024.
\newblock Kto: Model alignment as prospect theoretic optimization.
\newblock \emph{arXiv preprint arXiv:2402.01306}.

\bibitem[{Fujii et~al.(2024)Fujii, Nakamura, Loem, Iida, Ohi, Hattori, Shota, Mizuki, Yokota, and Okazaki}]{Fujii:COLM2024}
Kazuki Fujii, Taishi Nakamura, Mengsay Loem, Hiroki Iida, Masanari Ohi, Kakeru Hattori, Hirai Shota, Sakae Mizuki, Rio Yokota, and Naoaki Okazaki. 2024.
\newblock Continual pre-training for cross-lingual llm adaptation: Enhancing japanese language capabilities.
\newblock In \emph{Proceedings of the First Conference on Language Modeling}, COLM, page (to appear), University of Pennsylvania, USA.

\bibitem[{fujiki(2023)}]{japanese-alpaca}
fujiki. 2023.
\newblock fujiki/japanese\_alpaca\_data.
\newblock \url{https://huggingface.co/datasets/fujiki/japanese_alpaca_data}.

\bibitem[{Haerpfer et~al.(2021)Haerpfer, Inglehart, Moreno, Welzel, Kizilova, Diez-Medrano, Lagos, Norris, Ponarin, and Puranen}]{wvs}
Christian Haerpfer, Ronald Inglehart, Alejandro Moreno, Christian Welzel, Kseniya Kizilova, Jaime Diez-Medrano, Marta Lagos, Pippa Norris, E~Ponarin, and B~Puranen. 2021.
\newblock World values survey: Round seven.
\newblock \emph{JD Systems Institute \& WVSA Secretariat. Data File Version, 2(0)}.

\bibitem[{Hendrycks et~al.(2020)Hendrycks, Burns, Basart, Zou, Mazeika, Song, and Steinhardt}]{hendrycks2020measuring}
Dan Hendrycks, Collin Burns, Steven Basart, Andy Zou, Mantas Mazeika, Dawn Song, and Jacob Steinhardt. 2020.
\newblock Measuring massive multitask language understanding.
\newblock \emph{arXiv preprint arXiv:2009.03300}.

\bibitem[{Huang and Yang(2023)}]{huang2023culturally}
Jing Huang and Diyi Yang. 2023.
\newblock Culturally aware natural language inference.
\newblock In \emph{Findings of the Association for Computational Linguistics: EMNLP 2023}, pages 7591--7609.

\bibitem[{Huang et~al.(2025)Huang, Durmus, McCain, Handa, Tamkin, Hong, Stern, Somani, Zhang, and Ganguli}]{huang2025valueswilddiscoveringanalyzing}
Saffron Huang, Esin Durmus, Miles McCain, Kunal Handa, Alex Tamkin, Jerry Hong, Michael Stern, Arushi Somani, Xiuruo Zhang, and Deep Ganguli. 2025.
\newblock \href {https://arxiv.org/abs/2504.15236} {Values in the {W}ild: Discovering and analyzing values in real-world language model interactions}.
\newblock \emph{Preprint}, arXiv:2504.15236.

\bibitem[{Hwang(2024)}]{japanese-orca-dpo-pairs}
Yongtae Hwang. 2024.
\newblock \href {https://huggingface.co/datasets/yongtae-jp/orca_dpo_pairs_ja} {{yongtae-jp/orca\_dpo\_pairs\_ja}}.

\bibitem[{Intel(2024)}]{intel-orca-dpo-pairs}
Intel. 2024.
\newblock \href {https://huggingface.co/datasets/Intel/orca_dpo_pairs} {{Intel/orca\_dpo\_pairs}}.

\bibitem[{Ivison et~al.(2023)Ivison, Wang, Pyatkin, Lambert, Peters, Dasigi, Jang, Wadden, Smith, Beltagy et~al.}]{ivison2023camels}
Hamish Ivison, Yizhong Wang, Valentina Pyatkin, Nathan Lambert, Matthew Peters, Pradeep Dasigi, Joel Jang, David Wadden, Noah~A Smith, Iz~Beltagy, et~al. 2023.
\newblock Camels in a changing climate: Enhancing lm adaptation with tulu 2.
\newblock \emph{arXiv preprint arXiv:2311.10702}.

\bibitem[{Karinshak et~al.(2024)Karinshak, Hu, Kong, Rao, Wang, Wang, and Zeng}]{karinshak2024llm}
Elise Karinshak, Amanda Hu, Kewen Kong, Vishwanatha Rao, Jingren Wang, Jindong Wang, and Yi~Zeng. 2024.
\newblock Llm-globe: A benchmark evaluating the cultural values embedded in llm output.
\newblock \emph{arXiv preprint arXiv:2411.06032}.

\bibitem[{Keleg and Magdy(2023)}]{keleg2023dlama}
Amr Keleg and Walid Magdy. 2023.
\newblock {DLAMA}: A framework for curating culturally diverse facts for probing the knowledge of pretrained language models.
\newblock \emph{arXiv preprint arXiv:2306.05076}.

\bibitem[{Kirk et~al.(2024)Kirk, Whitefield, R{\"o}ttger, Bean, Margatina, Ciro, Mosquera, Bartolo, Williams, He et~al.}]{kirk2024prism}
Hannah~Rose Kirk, Alexander Whitefield, Paul R{\"o}ttger, Andrew Bean, Katerina Margatina, Juan Ciro, Rafael Mosquera, Max Bartolo, Adina Williams, He~He, et~al. 2024.
\newblock The prism alignment project: What participatory, representative and individualised human feedback reveals about the subjective and multicultural alignment of large language models.
\newblock \emph{arXiv preprint arXiv:2404.16019}.

\bibitem[{Kong et~al.(2023)Kong, Zhao, Chen, Li, Qin, Sun, Zhou, Wang, and Dong}]{kong2023better}
Aobo Kong, Shiwan Zhao, Hao Chen, Qicheng Li, Yong Qin, Ruiqi Sun, Xin Zhou, Enzhi Wang, and Xiaohang Dong. 2023.
\newblock Better zero-shot reasoning with role-play prompting.
\newblock \emph{arXiv preprint arXiv:2308.07702}.

\bibitem[{K{\"o}pf et~al.(2023)K{\"o}pf, Kilcher, Von~R{\"u}tte, Anagnostidis, Tam, Stevens, Barhoum, Nguyen, Stanley, Nagyfi et~al.}]{kopf2023openassistant}
Andreas K{\"o}pf, Yannic Kilcher, Dimitri Von~R{\"u}tte, Sotiris Anagnostidis, Zhi~Rui Tam, Keith Stevens, Abdullah Barhoum, Duc Nguyen, Oliver Stanley, Rich{\'a}rd Nagyfi, et~al. 2023.
\newblock Openassistant conversations-democratizing large language model alignment.
\newblock \emph{Advances in Neural Information Processing Systems}, 36:47669--47681.

\bibitem[{Koto et~al.(2024)Koto, Li, Shatnawi, Doughman, Sadallah, Alraeesi, Almubarak, Alyafeai, Sengupta, Shehata et~al.}]{koto2024arabicmmlu}
Fajri Koto, Haonan Li, Sara Shatnawi, Jad Doughman, Abdelrahman~Boda Sadallah, Aisha Alraeesi, Khalid Almubarak, Zaid Alyafeai, Neha Sengupta, Shady Shehata, et~al. 2024.
\newblock {ArabicMMLU}: Assessing massive multitask language understanding in arabic.
\newblock \emph{arXiv preprint arXiv:2402.12840}.

\bibitem[{Kwon et~al.(2023)Kwon, Li, Zhuang, Sheng, Zheng, Yu, Gonzalez, Zhang, and Stoica}]{kwon2023efficient}
Woosuk Kwon, Zhuohan Li, Siyuan Zhuang, Ying Sheng, Lianmin Zheng, Cody~Hao Yu, Joseph Gonzalez, Hao Zhang, and Ion Stoica. 2023.
\newblock Efficient memory management for large language model serving with pagedattention.
\newblock In \emph{Proceedings of the 29th Symposium on Operating Systems Principles}, pages 611--626.

\bibitem[{Li et~al.(2024{\natexlab{a}})Li, Chen, Wang, Sitaram, and Xie}]{li2024culturellm}
Cheng Li, Mengzhou Chen, Jindong Wang, Sunayana Sitaram, and Xing Xie. 2024{\natexlab{a}}.
\newblock {CultureLLM}: Incorporating cultural differences into large language models.
\newblock \emph{arXiv preprint arXiv:2402.10946}.

\bibitem[{Li et~al.(2024{\natexlab{b}})Li, Teney, Yang, Wen, Xie, and Wang}]{li2024culturepark}
Cheng Li, Damien Teney, Linyi Yang, Qingsong Wen, Xing Xie, and Jindong Wang. 2024{\natexlab{b}}.
\newblock Culturepark: Boosting cross-cultural understanding in large language models.
\newblock \emph{arXiv preprint arXiv:2405.15145}.

\bibitem[{Li et~al.(2023)Li, Zhang, Koto, Yang, Zhao, Gong, Duan, and Baldwin}]{li2023cmmlu}
Haonan Li, Yixuan Zhang, Fajri Koto, Yifei Yang, Hai Zhao, Yeyun Gong, Nan Duan, and Timothy Baldwin. 2023.
\newblock {CMMLU}: Measuring massive multitask language understanding in chinese.
\newblock \emph{arXiv preprint arXiv:2306.09212}.

\bibitem[{Lian et~al.(2023)Lian, Goodson, Pentland, Cook, Vong, and "Teknium"}]{OpenOrca}
Wing Lian, Bleys Goodson, Eugene Pentland, Austin Cook, Chanvichet Vong, and "Teknium". 2023.
\newblock {OpenOrca}: An open dataset of gpt augmented flan reasoning traces.
\newblock \url{https://https://huggingface.co/Open-Orca/OpenOrca}.

\bibitem[{Lin et~al.(2021)Lin, Hilton, and Evans}]{lin2021truthfulqa}
Stephanie Lin, Jacob Hilton, and Owain Evans. 2021.
\newblock Truthfulqa: Measuring how models mimic human falsehoods.
\newblock \emph{arXiv preprint arXiv:2109.07958}.

\bibitem[{Liu et~al.(2024)Liu, Gurevych, and Korhonen}]{liu2024culturally}
Chen~Cecilia Liu, Iryna Gurevych, and Anna Korhonen. 2024.
\newblock Culturally aware and adapted {NLP}: A taxonomy and a survey of the state of the art.
\newblock \emph{arXiv preprint arXiv:2406.03930}.

\bibitem[{Masoud et~al.(2023)Masoud, Liu, Ferianc, Treleaven, and Rodrigues}]{masoud2023cultural}
Reem~I Masoud, Ziquan Liu, Martin Ferianc, Philip Treleaven, and Miguel Rodrigues. 2023.
\newblock Cultural alignment in large language models: An explanatory analysis based on hofstede's cultural dimensions.
\newblock \emph{arXiv preprint arXiv:2309.12342}.

\bibitem[{Meng et~al.(2024)Meng, Xia, and Chen}]{meng2024simpo}
Yu~Meng, Mengzhou Xia, and Danqi Chen. 2024.
\newblock Simpo: Simple preference optimization with a reference-free reward.
\newblock \emph{arXiv preprint arXiv:2405.14734}.

\bibitem[{Mitchell et~al.(2025)Mitchell, Attanasio, Baldini, Clinciu, Clive, Delobelle, Dey, Hamilton, Dill, Doughman, Dutt, Ghosh, Forde, Holtermann, Kaffee, Laud, Lauscher, Lopez-Davila, Masoud, Nangia, Ovalle, Pistilli, Radev, Savoldi, Raheja, Qin, Ploeger, Subramonian, Dhole, Sun, Djanibekov, Mansurov, Yin, Cueva, Mukherjee, Huang, Shen, Gala, Al-Ali, Djanibekov, Mukhituly, Nie, Sharma, Stanczak, Szczechla, Timponi~Torrent, Tunuguntla, Viridiano, Van Der~Wal, Yakefu, N{\'e}v{\'e}ol, Zhang, Zink, and Talat}]{mitchell-etal-2025-shades}
Margaret Mitchell, Giuseppe Attanasio, Ioana Baldini, Miruna Clinciu, Jordan Clive, Pieter Delobelle, Manan Dey, Sil Hamilton, Timm Dill, Jad Doughman, Ritam Dutt, Avijit Ghosh, Jessica~Zosa Forde, Carolin Holtermann, Lucie-Aim{\'e}e Kaffee, Tanmay Laud, Anne Lauscher, Roberto~L Lopez-Davila, Maraim Masoud, Nikita Nangia, Anaelia Ovalle, Giada Pistilli, Dragomir Radev, Beatrice Savoldi, Vipul Raheja, Jeremy Qin, Esther Ploeger, Arjun Subramonian, Kaustubh Dhole, Kaiser Sun, Amirbek Djanibekov, Jonibek Mansurov, Kayo Yin, Emilio~Villa Cueva, Sagnik Mukherjee, Jerry Huang, Xudong Shen, Jay Gala, Hamdan Al-Ali, Tair Djanibekov, Nurdaulet Mukhituly, Shangrui Nie, Shanya Sharma, Karolina Stanczak, Eliza Szczechla, Tiago Timponi~Torrent, Deepak Tunuguntla, Marcelo Viridiano, Oskar Van Der~Wal, Adina Yakefu, Aur{\'e}lie N{\'e}v{\'e}ol, Mike Zhang, Sydney Zink, and Zeerak Talat. 2025.
\newblock {SHADES}: Towards a multilingual assessment of stereotypes in large language models.
\newblock In \emph{Proceedings of the 2025 Conference of the Nations of the Americas Chapter of the Association for Computational Linguistics: Human Language Technologies (Volume 1: Long Papers)}. Association for Computational Linguistics.

\bibitem[{Mousi et~al.(2024)Mousi, Durrani, Ahmad, Hasan, Hasanain, Kabbani, Dalvi, Chowdhury, and Alam}]{mousi2024aradice}
Basel Mousi, Nadir Durrani, Fatema Ahmad, Md~Arid Hasan, Maram Hasanain, Tameem Kabbani, Fahim Dalvi, Shammur~Absar Chowdhury, and Firoj Alam. 2024.
\newblock Aradice: Benchmarks for dialectal and cultural capabilities in llms.
\newblock \emph{arXiv preprint arXiv:2409.11404}.

\bibitem[{Muennighoff et~al.(2022)Muennighoff, Wang, Sutawika, Roberts, Biderman, Scao, Bari, Shen, Yong, Schoelkopf et~al.}]{muennighoff2022crosslingual}
Niklas Muennighoff, Thomas Wang, Lintang Sutawika, Adam Roberts, Stella Biderman, Teven~Le Scao, M~Saiful Bari, Sheng Shen, Zheng-Xin Yong, Hailey Schoelkopf, et~al. 2022.
\newblock Crosslingual generalization through multitask finetuning.
\newblock \emph{arXiv preprint arXiv:2211.01786}.

\bibitem[{Myung et~al.(2024)Myung, Lee, Zhou, Jin, Putri, Antypas, Borkakoty, Kim, Perez-Almendros, Ayele et~al.}]{myung2024blend}
Junho Myung, Nayeon Lee, Yi~Zhou, Jiho Jin, Rifki Putri, Dimosthenis Antypas, Hsuvas Borkakoty, Eunsu Kim, Carla Perez-Almendros, Abinew~Ali Ayele, et~al. 2024.
\newblock {BLEnD}: A benchmark for llms on everyday knowledge in diverse cultures and languages.
\newblock \emph{Advances in Neural Information Processing Systems}, 37:78104--78146.

\bibitem[{Naous et~al.(2023)Naous, Ryan, Ritter, and Xu}]{naous2023having}
Tarek Naous, Michael~J Ryan, Alan Ritter, and Wei Xu. 2023.
\newblock Having beer after prayer? measuring cultural bias in large language models.
\newblock \emph{arXiv preprint arXiv:2305.14456}.

\bibitem[{Naous et~al.(2024)Naous, Ryan, Ritter, and Xu}]{naous-etal-2024-beer}
Tarek Naous, Michael~J Ryan, Alan Ritter, and Wei Xu. 2024.
\newblock \href {https://doi.org/10.18653/v1/2024.acl-long.862} {Having beer after prayer? measuring cultural bias in large language models}.
\newblock In \emph{Proceedings of the 62nd Annual Meeting of the Association for Computational Linguistics (Volume 1: Long Papers)}, pages 16366--16393, Bangkok, Thailand. Association for Computational Linguistics.

\bibitem[{Naous and Xu(2025)}]{naous2025origin}
Tarek Naous and Wei Xu. 2025.
\newblock On the origin of cultural biases in language models: From pre-training data to linguistic phenomena.
\newblock \emph{arXiv preprint arXiv:2501.04662}.

\bibitem[{Nguyen et~al.(2023)Nguyen, Razniewski, Varde, and Weikum}]{nguyen2023extracting}
Tuan-Phong Nguyen, Simon Razniewski, Aparna Varde, and Gerhard Weikum. 2023.
\newblock Extracting cultural commonsense knowledge at scale.
\newblock In \emph{Proceedings of the ACM Web Conference 2023}, pages 1907--1917.

\bibitem[{Onohara et~al.(2024)Onohara, Miyai, Imajuku, Egashira, Baek, Yue, Neubig, and Aizawa}]{onohara2024jmmmu}
Shota Onohara, Atsuyuki Miyai, Yuki Imajuku, Kazuki Egashira, Jeonghun Baek, Xiang Yue, Graham Neubig, and Kiyoharu Aizawa. 2024.
\newblock {JMMMU}: A {J}apanese massive multi-discipline multimodal understanding benchmark for culture-aware evaluation.
\newblock \emph{arXiv preprint arXiv:2410.17250}.

\bibitem[{Ouyang et~al.(2022)Ouyang, Wu, Jiang, Almeida, Wainwright, Mishkin, Zhang, Agarwal, Slama, Ray et~al.}]{ouyang2022training}
Long Ouyang, Jeffrey Wu, Xu~Jiang, Diogo Almeida, Carroll Wainwright, Pamela Mishkin, Chong Zhang, Sandhini Agarwal, Katarina Slama, Alex Ray, et~al. 2022.
\newblock Training language models to follow instructions with human feedback.
\newblock \emph{Advances in neural information processing systems}, 35:27730--27744.

\bibitem[{Palta and Rudinger(2023)}]{palta2023fork}
Shramay Palta and Rachel Rudinger. 2023.
\newblock {FORK}: A bite-sized test set for probing culinary cultural biases in commonsense reasoning models.
\newblock In \emph{Findings of the Association for Computational Linguistics: ACL 2023}, pages 9952--9962.

\bibitem[{Pan(2024)}]{chinese-orca-dpo-pairs}
Wenbo Pan. 2024.
\newblock \href {https://huggingface.co/datasets/wenbopan/Chinese-dpo-pairs} {{wenbopan/Chinese-dpo-pairs}}.

\bibitem[{Pawar et~al.(2024)Pawar, Park, Jin, Arora, Myung, Yadav, Haznitrama, Song, Oh, and Augenstein}]{pawar2024survey}
Siddhesh Pawar, Junyeong Park, Jiho Jin, Arnav Arora, Junho Myung, Srishti Yadav, Faiz~Ghifari Haznitrama, Inhwa Song, Alice Oh, and Isabelle Augenstein. 2024.
\newblock Survey of cultural awareness in language models: Text and beyond.
\newblock \emph{arXiv preprint arXiv:2411.00860}.

\bibitem[{Pham et~al.(2025)Pham, Li, Qu, and Haffari}]{pham-etal-2025-cultureinstruct}
Viet~Thanh Pham, Zhuang Li, Lizhen Qu, and Gholamreza Haffari. 2025.
\newblock {C}ulture{I}nstruct: Curating multi-cultural instructions at scale.
\newblock In \emph{Proceedings of the 2025 Conference of the Nations of the Americas Chapter of the Association for Computational Linguistics: Human Language Technologies (Volume 1: Long Papers)}. Association for Computational Linguistics.

\bibitem[{Qian et~al.(2024)Qian, Altam, Alqurishi, and Souissi}]{qian2024cameleval}
Zhaozhi Qian, Faroq Altam, Muhammad Alqurishi, and Riad Souissi. 2024.
\newblock Cameleval: Advancing culturally aligned arabic language models and benchmarks.
\newblock \emph{arXiv preprint arXiv:2409.12623}.

\bibitem[{Qiu et~al.(2025)Qiu, Huang, Zheng, Sun, and Peng}]{qiu2025multimodal}
Haoyi Qiu, Kung-Hsiang Huang, Ruichen Zheng, Jiao Sun, and Nanyun Peng. 2025.
\newblock Multimodal cultural safety: Evaluation frameworks and alignment strategies.
\newblock \emph{arXiv preprint arXiv:2505.14972}.

\bibitem[{Rafailov et~al.(2024)Rafailov, Sharma, Mitchell, Manning, Ermon, and Finn}]{rafailov2024direct}
Rafael Rafailov, Archit Sharma, Eric Mitchell, Christopher~D Manning, Stefano Ermon, and Chelsea Finn. 2024.
\newblock Direct preference optimization: Your language model is secretly a reward model.
\newblock \emph{Advances in Neural Information Processing Systems}, 36.

\bibitem[{Rao et~al.(2024)Rao, Yerukola, Shah, Reinecke, and Sap}]{rao2024normad}
Abhinav Rao, Akhila Yerukola, Vishwa Shah, Katharina Reinecke, and Maarten Sap. 2024.
\newblock Normad: A benchmark for measuring the cultural adaptability of large language models.
\newblock \emph{arXiv preprint arXiv:2404.12464}.

\bibitem[{Romanou et~al.(2025)Romanou, Foroutan, Sotnikova, Nelaturu, Singh, Maheshwary, Altomare, Chen, Haggag, A, Amayuelas, Amirudin, Boiko, Chang, Chim, Cohen, Dalmia, Diress, Duwal, Dzenhaliou, Florez, Farestam, Imperial, Islam, Isotalo, Jabbarishiviari, Karlsson, Khalilov, Klamm, Koto, Krzemi{\'n}ski, de~Melo, Montariol, Nan, Niklaus, Novikova, Ceron, Paul, Ploeger, Purbey, Rajwal, Ravi, Rydell, Santhosh, Sharma, Skenduli, Moakhar, soltani moakhar, Tarun, Wasi, Weerasinghe, Yilmaz, Zhang, Schlag, Fadaee, Hooker, and Bosselut}]{romanou2025include}
Angelika Romanou, Negar Foroutan, Anna Sotnikova, Sree~Harsha Nelaturu, Shivalika Singh, Rishabh Maheshwary, Micol Altomare, Zeming Chen, Mohamed~A. Haggag, Snegha A, Alfonso Amayuelas, Azril~Hafizi Amirudin, Danylo Boiko, Michael Chang, Jenny Chim, Gal Cohen, Aditya~Kumar Dalmia, Abraham Diress, Sharad Duwal, Daniil Dzenhaliou, Daniel Fernando~Erazo Florez, Fabian Farestam, Joseph~Marvin Imperial, Shayekh~Bin Islam, Perttu Isotalo, Maral Jabbarishiviari, B{\"o}rje~F. Karlsson, Eldar Khalilov, Christopher Klamm, Fajri Koto, Dominik Krzemi{\'n}ski, Gabriel~Adriano de~Melo, Syrielle Montariol, Yiyang Nan, Joel Niklaus, Jekaterina Novikova, Johan Samir~Obando Ceron, Debjit Paul, Esther Ploeger, Jebish Purbey, Swati Rajwal, Selvan~Sunitha Ravi, Sara Rydell, Roshan Santhosh, Drishti Sharma, Marjana~Prifti Skenduli, Arshia~Soltani Moakhar, Bardia soltani moakhar, Ayush~Kumar Tarun, Azmine~Toushik Wasi, Thenuka~Ovin Weerasinghe, Serhan Yilmaz, Mike Zhang, Imanol Schlag, Marzieh Fadaee, Sara Hooker, and Antoine
  Bosselut. 2025.
\newblock \href {https://openreview.net/forum?id=k3gCieTXeY} {{INCLUDE}: Evaluating multilingual language understanding with regional knowledge}.
\newblock In \emph{The Thirteenth International Conference on Learning Representations}.

\bibitem[{Rudinger et~al.(2018)Rudinger, Naradowsky, Leonard, and Van~Durme}]{rudinger2018gender}
Rachel Rudinger, Jason Naradowsky, Brian Leonard, and Benjamin Van~Durme. 2018.
\newblock Gender bias in coreference resolution.
\newblock \emph{arXiv preprint arXiv:1804.09301}.

\bibitem[{Ryan et~al.(2024)Ryan, Held, and Yang}]{ryan2024unintended}
Michael~J Ryan, William Held, and Diyi Yang. 2024.
\newblock Unintended impacts of llm alignment on global representation.
\newblock \emph{arXiv preprint arXiv:2402.15018}.

\bibitem[{Rystr{\o}m et~al.(2025)Rystr{\o}m, Kirk, and Hale}]{rystrom2025multilingual}
Jonathan Rystr{\o}m, Hannah~Rose Kirk, and Scott Hale. 2025.
\newblock Multilingual!= multicultural: Evaluating gaps between multilingual capabilities and cultural alignment in llms.
\newblock \emph{arXiv preprint arXiv:2502.16534}.

\bibitem[{She et~al.(2024)She, Zou, Huang, Zhu, Liu, Geng, and Chen}]{she2024mapo}
Shuaijie She, Wei Zou, Shujian Huang, Wenhao Zhu, Xiang Liu, Xiang Geng, and Jiajun Chen. 2024.
\newblock {MAPO}: Advancing multilingual reasoning through multilingual alignment-as-preference optimization.
\newblock \emph{arXiv preprint arXiv:2401.06838}.

\bibitem[{Shen et~al.(2024)Shen, Logeswaran, Lee, Lee, Poria, and Mihalcea}]{shen2024understanding}
Siqi Shen, Lajanugen Logeswaran, Moontae Lee, Honglak Lee, Soujanya Poria, and Rada Mihalcea. 2024.
\newblock Understanding the capabilities and limitations of large language models for cultural commonsense.
\newblock \emph{arXiv preprint arXiv:2405.04655}.

\bibitem[{Shi et~al.(2024)Shi, Li, Zhang, Ziems, Horesh, de~Paula, Yang et~al.}]{shi2024culturebank}
Weiyan Shi, Ryan Li, Yutong Zhang, Caleb Ziems, Raya Horesh, Rog{\'e}rio~Abreu de~Paula, Diyi Yang, et~al. 2024.
\newblock Culturebank: An online community-driven knowledge base towards culturally aware language technologies.
\newblock \emph{arXiv preprint arXiv:2404.15238}.

\bibitem[{Singh et~al.(2024)Singh, Vargus, Dsouza, Karlsson, Mahendiran, Ko, Shandilya, Patel, Mataciunas, OMahony et~al.}]{singh2024aya}
Shivalika Singh, Freddie Vargus, Daniel Dsouza, B{\"o}rje~F Karlsson, Abinaya Mahendiran, Wei-Yin Ko, Herumb Shandilya, Jay Patel, Deividas Mataciunas, Laura OMahony, et~al. 2024.
\newblock Aya dataset: An open-access collection for multilingual instruction tuning.
\newblock \emph{arXiv preprint arXiv:2402.06619}.

\bibitem[{Team et~al.(2024)Team, Riviere, Pathak, Sessa, Hardin, Bhupatiraju, Hussenot, Mesnard, Shahriari, Ram{\'e} et~al.}]{team2024gemma}
Gemma Team, Morgane Riviere, Shreya Pathak, Pier~Giuseppe Sessa, Cassidy Hardin, Surya Bhupatiraju, L{\'e}onard Hussenot, Thomas Mesnard, Bobak Shahriari, Alexandre Ram{\'e}, et~al. 2024.
\newblock Gemma 2: Improving open language models at a practical size.
\newblock \emph{arXiv preprint arXiv:2408.00118}.

\bibitem[{Wang et~al.(2023)Wang, Jiao, Huang, Dai, Huang, Tu, and Lyu}]{wang2023not}
Wenxuan Wang, Wenxiang Jiao, Jingyuan Huang, Ruyi Dai, Jen-tse Huang, Zhaopeng Tu, and Michael~R Lyu. 2023.
\newblock Not all countries celebrate thanksgiving: On the cultural dominance in large language models.
\newblock \emph{arXiv preprint arXiv:2310.12481}.

\bibitem[{Wang et~al.(2025{\natexlab{a}})Wang, Chen, Li, Cho, Deng, Zhang, Chen, Wang, Grama, and Hong}]{wang2025more}
Yifan Wang, Runjin Chen, Bolian Li, David Cho, Yihe Deng, Ruqi Zhang, Tianlong Chen, Zhangyang Wang, Ananth Grama, and Junyuan Hong. 2025{\natexlab{a}}.
\newblock More is less: The pitfalls of multi-model synthetic preference data in dpo safety alignment.
\newblock \emph{arXiv preprint arXiv:2504.02193}.

\bibitem[{Wang et~al.(2025{\natexlab{b}})Wang, Zeng, Delalleau, Shin, Soares, Bukharin, Evans, Dong, and Kuchaiev}]{wang2025helpsteer3preferenceopenhumanannotatedpreference}
Zhilin Wang, Jiaqi Zeng, Olivier Delalleau, Hoo-Chang Shin, Felipe Soares, Alexander Bukharin, Ellie Evans, Yi~Dong, and Oleksii Kuchaiev. 2025{\natexlab{b}}.
\newblock \href {https://arxiv.org/abs/2505.11475} {Helpsteer3-preference: Open human-annotated preference data across diverse tasks and languages}.
\newblock \emph{Preprint}, arXiv:2505.11475.

\bibitem[{Wei et~al.(2022)Wei, Wang, Schuurmans, Bosma, Xia, Chi, Le, Zhou et~al.}]{wei2022chain}
Jason Wei, Xuezhi Wang, Dale Schuurmans, Maarten Bosma, Fei Xia, Ed~Chi, Quoc~V Le, Denny Zhou, et~al. 2022.
\newblock Chain-of-thought prompting elicits reasoning in large language models.
\newblock \emph{Advances in neural information processing systems}, 35:24824--24837.

\bibitem[{Wen-Yi et~al.(2024)Wen-Yi, Jo, Lin, and Mimno}]{wen2024chinese}
Andrea~W Wen-Yi, Unso Eun~Seo Jo, Lu~Jia Lin, and David Mimno. 2024.
\newblock How chinese are chinese language models? the puzzling lack of language policy in china's llms.
\newblock \emph{arXiv preprint arXiv:2407.09652}.

\bibitem[{Wuraola et~al.(2024)Wuraola, Dethlefs, and Marciniak}]{wuraola2024understanding}
Ifeoluwa Wuraola, Nina Dethlefs, and Daniel Marciniak. 2024.
\newblock Understanding slang with llms: Modelling cross-cultural nuances through paraphrasing.
\newblock In \emph{Proceedings of the 2024 Conference on Empirical Methods in Natural Language Processing}, pages 15525--15531.

\bibitem[{Xu et~al.(2025)Xu, Leng, Yu, and Xiong}]{xu-etal-2025-self}
Shaoyang Xu, Yongqi Leng, Linhao Yu, and Deyi Xiong. 2025.
\newblock \href {https://aclanthology.org/2025.naacl-long.350/} {Self-pluralising culture alignment for large language models}.
\newblock In \emph{Proceedings of the 2025 Conference of the Nations of the Americas Chapter of the Association for Computational Linguistics: Human Language Technologies (Volume 1: Long Papers)}. Association for Computational Linguistics.

\bibitem[{Xuan et~al.(2025)Xuan, Yang, Qi, Zeng, Xiao, Xing, Wang, Li, Li, Yu et~al.}]{xuan2025mmlu}
Weihao Xuan, Rui Yang, Heli Qi, Qingcheng Zeng, Yunze Xiao, Yun Xing, Junjue Wang, Huitao Li, Xin Li, Kunyu Yu, et~al. 2025.
\newblock Mmlu-prox: A multilingual benchmark for advanced large language model evaluation.
\newblock \emph{arXiv preprint arXiv:2503.10497}.

\bibitem[{Yang et~al.(2024{\natexlab{a}})Yang, Yang, Zhang, Hui, Zheng, Yu, Li, Liu, Huang, Wei et~al.}]{yang2024qwen2}
An~Yang, Baosong Yang, Beichen Zhang, Binyuan Hui, Bo~Zheng, Bowen Yu, Chengyuan Li, Dayiheng Liu, Fei Huang, Haoran Wei, et~al. 2024{\natexlab{a}}.
\newblock Qwen2.5 technical report.
\newblock \emph{arXiv e-prints}, pages arXiv--2412.

\bibitem[{Yang et~al.(2024{\natexlab{b}})Yang, Wu, Wang, Zong, and Zhang}]{yang2024language}
Wen Yang, Junhong Wu, Chen Wang, Chengqing Zong, and Jiajun Zhang. 2024{\natexlab{b}}.
\newblock Language imbalance driven rewarding for multilingual self-improving.
\newblock \emph{arXiv preprint arXiv:2410.08964}.

\bibitem[{Yao et~al.(2025)Yao, Yi, Wang, Dou, and Xie}]{yao2025caredio}
Jing Yao, Xiaoyuan Yi, Jindong Wang, Zhicheng Dou, and Xing Xie. 2025.
\newblock Caredio: Cultural alignment of llm via representativeness and distinctiveness guided data optimization.
\newblock \emph{arXiv preprint arXiv:2504.08820}.

\bibitem[{Yin et~al.(2022)Yin, Bansal, Monajatipoor, Li, and Chang}]{yin2022geomlama}
Da~Yin, Hritik Bansal, Masoud Monajatipoor, Liunian~Harold Li, and Kai-Wei Chang. 2022.
\newblock Geomlama: Geo-diverse commonsense probing on multilingual pre-trained language models.
\newblock \emph{arXiv preprint arXiv:2205.12247}.

\bibitem[{Yuan et~al.(2024)Yuan, Di, Zhao, and Naseem}]{yuan2024cultural}
Jiahao Yuan, Zixiang Di, Shangzixin Zhao, and Usman Naseem. 2024.
\newblock Cultural palette: Pluralising culture alignment via multi-agent palette.
\newblock \emph{arXiv preprint arXiv:2412.11167}.

\bibitem[{Zhan et~al.(2024)Zhan, Li, Kang, Feng, Hua, Qu, Ying, Chandra, Rosalin, Jureynolds et~al.}]{zhan2024renovi}
Haolan Zhan, Zhuang Li, Xiaoxi Kang, Tao Feng, Yuncheng Hua, Lizhen Qu, Yi~Ying, Mei~Rianto Chandra, Kelly Rosalin, Jureynolds Jureynolds, et~al. 2024.
\newblock {RENOVI}: A benchmark towards remediating norm violations in socio-cultural conversations.
\newblock \emph{arXiv preprint arXiv:2402.11178}.

\bibitem[{Zheng et~al.(2023)Zheng, Chiang, Sheng, Zhuang, Wu, Zhuang, Lin, Li, Li, Xing et~al.}]{zheng2023judging}
Lianmin Zheng, Wei-Lin Chiang, Ying Sheng, Siyuan Zhuang, Zhanghao Wu, Yonghao Zhuang, Zi~Lin, Zhuohan Li, Dacheng Li, Eric Xing, et~al. 2023.
\newblock Judging llm-as-a-judge with mt-bench and chatbot arena.
\newblock \emph{Advances in Neural Information Processing Systems}, 36:46595--46623.

\bibitem[{Zhong et~al.(2021)Zhong, Yang, Xu, and Yang}]{zhong-etal-2021-wikibias-detecting}
Yang Zhong, Jingfeng Yang, Wei Xu, and Diyi Yang. 2021.
\newblock \href {https://doi.org/10.18653/v1/2021.findings-emnlp.155} {{WIKIBIAS}: Detecting multi-span subjective biases in language}.
\newblock In \emph{Findings of the Association for Computational Linguistics: EMNLP 2021}, pages 1799--1814, Punta Cana, Dominican Republic. Association for Computational Linguistics.

\bibitem[{Zhou et~al.(2024{\natexlab{a}})Zhou, Liu, Xu, Iyer, Sun, Mao, Ma, Efrat, Yu, Yu et~al.}]{zhou2024lima}
Chunting Zhou, Pengfei Liu, Puxin Xu, Srinivasan Iyer, Jiao Sun, Yuning Mao, Xuezhe Ma, Avia Efrat, Ping Yu, Lili Yu, et~al. 2024{\natexlab{a}}.
\newblock Lima: Less is more for alignment.
\newblock \emph{Advances in Neural Information Processing Systems}, 36.

\bibitem[{Zhou et~al.(2024{\natexlab{b}})Zhou, Karidi, Liu, Garneau, Cao, Chen, Li, and Hershcovich}]{zhou2024does}
Li~Zhou, Taelin Karidi, Wanlong Liu, Nicolas Garneau, Yong Cao, Wenyu Chen, Haizhou Li, and Daniel Hershcovich. 2024{\natexlab{b}}.
\newblock Does mapo tofu contain coffee? probing llms for food-related cultural knowledge.
\newblock \emph{arXiv preprint arXiv:2404.06833}.

\end{thebibliography}

\clearpage

\appendix


\begin{table*}[t]
\centering
{\footnotesize 
\begin{tabular}{m{10cm} m{1cm} m{4cm}}
\toprule
\textbf{QA Example} & \textbf{Pref.} & \textbf{Features} \\
\midrule
\rowcolor{gray!15}\textbf{\textit{CultureLLM}} {\scriptsize\cite{li2024culturellm}} &  &  \\ 
\midrule
Q: \begin{CJK}{UTF8}{gbsn}请从1到4选择一个回答：你是否同意我的人生主要目标之一就是让我的父母感到骄傲？1. 强烈同意 2. 同意 3. 不同意 4. 强烈不同意。\end{CJK}  
(Select an answer from 1 to 4: Do you agree that one of the main goals in life is to make parents proud? 1. Strongly Agree 2. Agree 3. Disagree 4. Strongly Disagree.) \newline A: 2. & \multicolumn{1}{c}{---} & 
\begin{itemize}[left=1pt,itemsep=0pt,topsep=0pt]
  \item Synthetic data
  \item SFT data
  \item No preferences
\end{itemize}
\\
\midrule
\rowcolor{gray!15}\textbf{\textit{Prism}} {\scriptsize\cite{kirk2024prism}} &  &  \\ 
\midrule
Q: What can you do about the inequality of wealth?\newline
A: Make it illegal for corporations to own a majority of the wealth. & \multicolumn{1}{c}{17/100} & 
\begin{minipage}[t]{\linewidth}
\begin{itemize}[left=1pt,itemsep=0pt,topsep=0pt]
  \item English only
  \item Generic topics
  \item Human-rated
\end{itemize}
\end{minipage}\vspace{0.3\baselineskip}
\\
\midrule
\rowcolor{gray!15}\textbf{\textit{CARE (Ours)}} \includegraphics[width=0.5cm]{fig/Hugging.jpg} &  &   \\
\midrule
Q: \begin{CJK}{UTF8}{gbsn}在中国，用名字来叫更年长的人是合适的吗？\end{CJK}  
(In China, is it appropriate to address older people by their first names?) \newline
A: \begin{CJK}{UTF8}{gbsn}在中国文化中，直接用名字来叫更年长的人通常被视为不尊重的行为。中国社会非常重视礼节和尊重长辈，通常会使用一些尊称或者称谓来称呼年长者，例如“叔叔”、“阿姨”、“爷爷”、“奶奶”等...\end{CJK}  
(In Chinese culture, it is often considered disrespectful...) & \multicolumn{1}{c}{9/10} & 
\begin{itemize}[left=1pt,itemsep=0pt,topsep=0pt]
  \item Multilingual (Zh/Ar/Ja)
  \item Culture-specific
  \item Human-rated
\end{itemize}
\\
\bottomrule
\end{tabular}
}
\caption{Example data within different resources for improving LMs' cultural awareness. ``Pref.'' refers to whether it offers human preference ratings for responses. The English in brackets is for translation only.}
\label{tab:related-work-comparison-example}
\end{table*}

\begin{table*}[t!]
\begin{adjustbox}{width=\linewidth}
\begin{tabular}{@{}lcccccc!{\vrule}cccccc!{\vrule}cccccc@{}}
\toprule
 & \multicolumn{6}{c}{\textbf{Chinese Culture}} & \multicolumn{6}{c}{\textbf{Arab Culture}} & \multicolumn{6}{c}{\textbf{Japanese Culture}} \\
 \cmidrule(l){2-7} 
 \cmidrule(l){8-13} 
 \cmidrule(l){14-19} 
\textbf{Source} & \multicolumn{1}{l}{\textit{Entities}} & \multicolumn{1}{l}{\textit{Opinion}} & \multicolumn{1}{l}{\textit{Norms}} & \multicolumn{1}{l}{\textit{Commonsense}} & \multicolumn{1}{l}{\textit{Literacy}} & \multicolumn{1}{l}{\textbf{Total}} & \multicolumn{1}{l}{\textit{Entities}} & \multicolumn{1}{l}{\textit{Opinion}} & \multicolumn{1}{l}{\textit{Norms}} & \multicolumn{1}{l}{\textit{Commonsense}} & \multicolumn{1}{l}{\textit{Literacy}} & \multicolumn{1}{l}{\textbf{Total}} & \multicolumn{1}{l}{\textit{Entities}} & \multicolumn{1}{l}{\textit{Opinion}} & \multicolumn{1}{l}{\textit{Norms}} & \multicolumn{1}{l}{\textit{Commonsense}} & \multicolumn{1}{l}{\textit{Literacy}} & \multicolumn{1}{l}{\textbf{Total}} \\ 
\midrule
Instruction Datasets & 224 & 115 & 3 & 2 & 231 & 575 & 242 & 42 & 8 & 7 & 141 & 440 & 293 & 1 & 15 & 5 & 98 & 412 \\
Cultural Benchmarks & 3 & 5 & 33 & 17 & 2 & 60 & 8 & 12 & 20 & 19 & 2 & 61 & 4 & 0 & 33 & 13 & 0 & 50 \\
Native Human Curation & 26 & 16 & 66 & 82 & 0 & 190 & 0 & 50 & 72 & 74 & 0 & 196 & 0 & 99 & 75 & 85 & 2 & 262 \\ \midrule
\multicolumn{1}{r}{\textbf{Total}} & 253 & 136 & 102 & 101 & 233 & \textbf{825} & 250 & 104 & 100 & 100 & 143 & \textbf{697} & 297 & 100 & 123 & 103 & 100 & \textbf{723} \\ \bottomrule
\end{tabular}
\end{adjustbox}
\caption{Statistics per cultural category for questions specific to Arab, Chinese, and Japanese cultures in CARE.}
\label{tab:statistics-arab-chinese}
\end{table*}

\begin{table}[t]
\begin{adjustbox}{width=\linewidth}
\begin{tabular}{@{}lcccccc@{}}
\toprule
 & \multicolumn{6}{c}{\textbf{Other Foreign Cultures}} \\ \cmidrule(l){2-7} 
\textbf{Source} & \multicolumn{1}{l}{\textit{Entities}} & \multicolumn{1}{l}{\textit{Opinion}} & \multicolumn{1}{l}{\textit{Norms}} & \multicolumn{1}{l}{\textit{Commonsense}} & \multicolumn{1}{l}{\textit{Literacy}} & \multicolumn{1}{l}{\textbf{Total}} \\ \midrule
Instruction Datasets & 658 & 198 & 72 & 27 & 147 & 1102 \\
Cultural Benchmarks & 0 & 0 & 51 & 18 & 0 & 69 \\
Web Resources & 0 & 0 & 9 & 65 & 0 & 74 \\ \midrule
\multicolumn{1}{r}{\textbf{Total}} & 504 & 197 & 128 & 109 & 120 & \textbf{1245} \\ \bottomrule
\end{tabular}
\end{adjustbox}
\caption{Statistics per cultural category for questions specific to other foreign cultures in CARE.}
\label{tab:statistics-foreign-cultures}
\end{table}


\begin{table}[t]
\centering
\resizebox{\linewidth}{!}{%
\begin{tabular}{lccccc}
\toprule
\multicolumn{1}{l}{\textit{Language}} & \textit{Sub-nationwide} & \textit{Nationwide} & \textit{Continent-wide} & \textit{Worldwide} & \textbf{Total} \\
\midrule
{Chinese} & 66 & 1420 & 99 & 133 & 1718 \\
{Arabic} & 7 & 675 & 157 & 23 & 862 \\
{Japanese} & 28 & 714 & 121 & 47 & 910 \\
\midrule
\textbf{Total} & 101 & 2809 & 377 & 203 & \textbf{3490} \\
\bottomrule
\end{tabular}%
}
\caption{Statistics per geographic scope for questions in CARE.}
\label{tab:statistics-geographic-scope}
\end{table}

\section{CARE: Details}

\subsection{Annotation Details}
\label{appendix:annotation-guideline}

\begin{figure*}[htbp]
    \centering
    \includegraphics[width=0.7\linewidth]{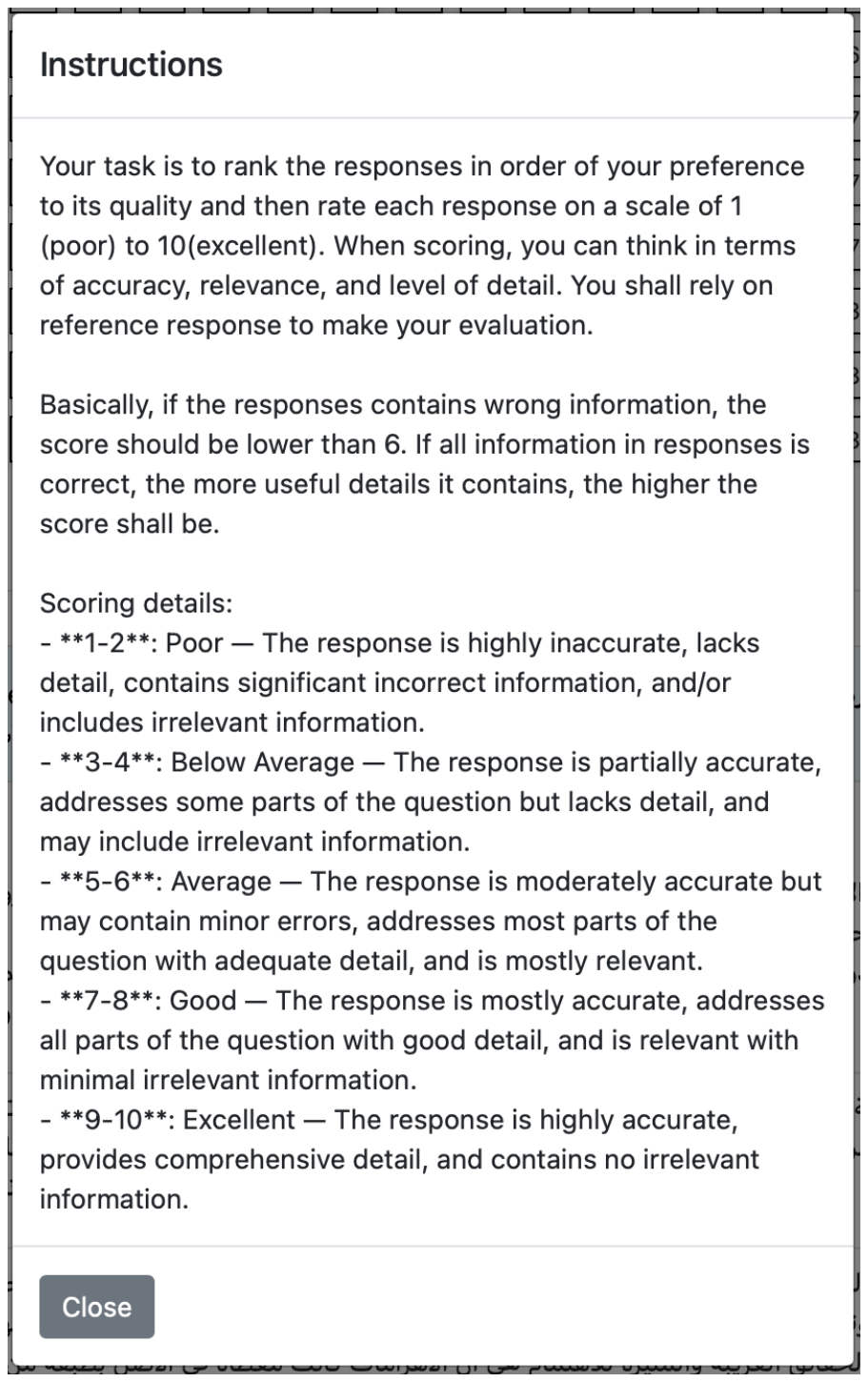}
    \caption{Cultural preference annotation instructions.}
    \label{fig:instruction}
\end{figure*}

\begin{figure*}[t]
    \centering
    \includegraphics[width=\linewidth]{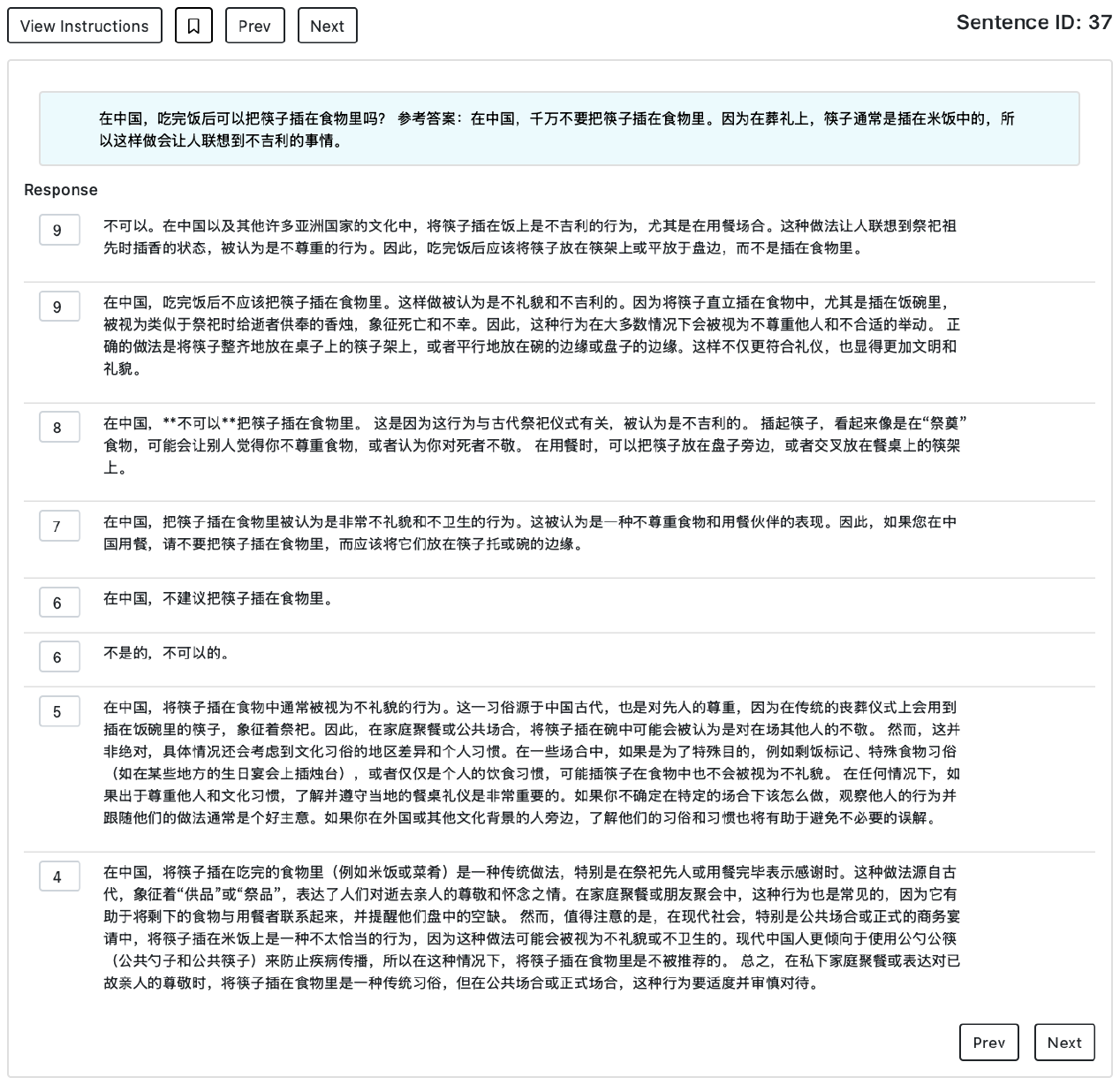}
    \caption{The interface for annotating culture-specific human preference.}
    \label{fig:rank-and-rate-interface}
\end{figure*}

To construct the question-answer pairs in CARE, we recruited native Chinese, Arabic, and Japanese speakers as mentioned in \S~\ref{sub:care_data}. 
Among our annotators, 5 Japanese annotators were workers with experience living abroad (e.g., in the US), while the rest were international college students.
The detailed annotation guideline for dataset construction is provided in Figures~\ref{fig:annotation-guideline-1} to \ref{fig:annotation-guideline-5}, and the instruction for preference annotation is provided in Figures~\ref{fig:instruction}. 
Our rank-and-rate web interface for collecting human cultural preference pairs is shown in Figure~\ref{fig:rank-and-rate-interface}.

\subsection{Annotation Agreements}
\label{appendix:annotation-agreement}

To assess annotators' agreement on this culture preference judgment task, we invite additional native speakers for each culture, and report the Pearson correlation on different cultural categories in Table~\ref{tab:per-category-correlation}. It shows that high correlation is achieved among annotators across five cultural categories.

\begin{table}[t]
\centering
\resizebox{\linewidth}{!}{%
\begin{tabular}{cccccc}
\toprule
\textbf{Culture} & \textit{Entities} & \textit{Opinion} & \textit{Norms} & \textit{Commonsense} & \textit{Literacy} \\
\midrule
Arab & 0.96&	0.90&	0.84&	0.88&	0.97 \\
Chinese & 0.84	&0.90&	0.96&	0.94&	0.99 \\
Japanese & 0.99&	0.88&	0.96&	0.85&	0.99 \\
\bottomrule
\end{tabular}%
}
\caption{Inter-annotator agreement measured by Spearman's $\rho$ on five cultural categories. Agreement for this preference judgment task is consistently high across different cultural categories. 
}
\label{tab:per-category-correlation}
\vspace{-1.0em}
\end{table}

\subsection{Statistics}
\label{appendix:dataset-stats}

Table~\ref{tab:statistics-arab-chinese} shows the detailed statistics for Chinese, Arab, and Japanese cultures in CARE across the cultural categories, stratified by the data source. The questions specific to Chinese, Arab, and Japanese cultures are written in Chinese, Arabic, and Japanese, respectively. The culturally-relevant samples obtained from the instruction-tuning dataset mostly fall within the \textit{cultural entities}, \textit{cultural opinion}, or \textit{literacy} categories, while samples obtained from cultural knowledge bases provide more \textit{social norms} and \textit{cultural commonsense} data. 

Table~\ref{tab:statistics-foreign-cultures} shows the detailed statistics for other foreign cultures in CARE for each cultural category. These samples were obtained when filtering the instruction tuning datasets and cultural knowledge bases (\S\ref{subsec:existing_data}). We wrote these foreign samples in the native language and used them as part of our training set and in our analysis on the impact of the source culture in preference learning (\S\ref{subsec:source-culture-analysis}). We also manually collected more samples for social norms and cultural commonsense from online websites to cross 100 samples in each category. We ensured these foreign samples do not include anything relevant to Chinese, Arab, or Japanese cultures.

Table~\ref{tab:statistics-geographic-scope} shows the detailed statistics for Chinese, Arabic, and Japanese samples in CARE across the geographic scope. 
Most samples fall into \textit{nationwide} category, while \textit{sub-nationwide} samples provide insights about more detailed cultural information, \textit{continent-wide} and \textit{worldwide} samples assess understanding between several cultures.

\subsection{Data Overlap Analysis}
We removed exact-duplicate questions in Section~\ref{sub:care_data}. To further probe potential paraphrase overlap, within each language, we embedded questions using \texttt{Salesforce/SFR-Embedding-Mistral}, computed cosine similarities for every train-test pair, and retained for each test question its most similar neighbor in the training set. We then manually checked each of them. We identified \textbf{6}/\textbf{4}/\textbf{3} paraphrase overlaps for \textbf{zh}/\textbf{ar}/\textbf{ja}, respectively, indicating minimal residual overlap - consistent with real-world applications where users often pose similar or recurring questions.

\begin{figure}[t]
    \centering
    \includegraphics[width=\linewidth]{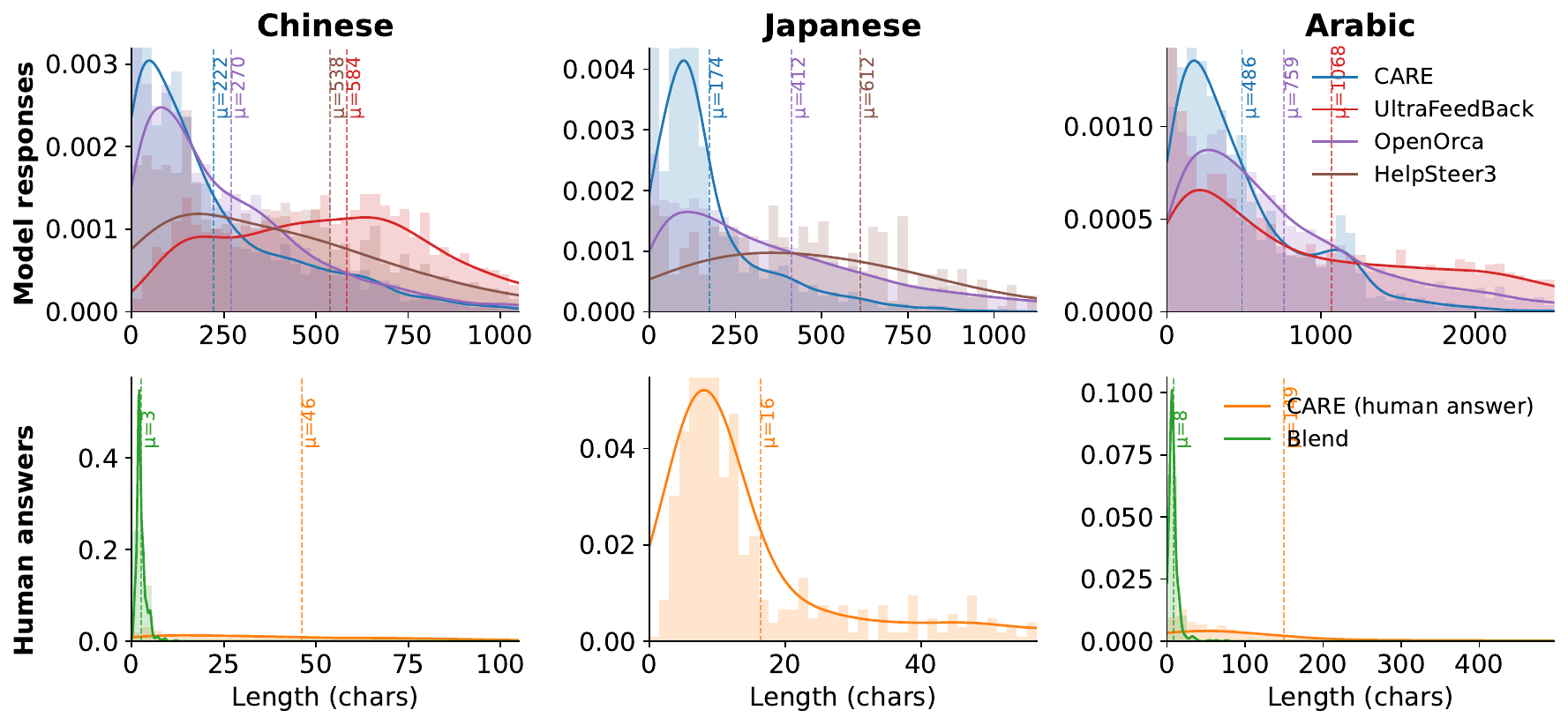}
    \caption{Length distributions across datasets and languages. Top: LLM-generated; Bottom: human-written. Human answers are concise; LLM responses are longer and heavy-tailed; dashed lines show means.}
    \label{fig:length_dis-compare}
    \vspace{-1em}
\end{figure}

\subsection{Comparison with existing datasets}
\label{appendix:data-comparison-example}

Table~\ref{tab:related-work-comparison-example} presents illustrative examples from representative datasets that aim to improve LMs' cultural awareness. 
While CultureLLM~\cite{li2024culturellm} focuses on synthetic data without preference supervision, and Prism~\cite{kirk2024prism} provides human ratings on generic English questions, neither provides proper multilingual and culture-specific contexts with human preferences. 
In contrast, CARE provides human preferences on culturally grounded topics across multiple languages.
We further examine response lengths across datasets. The LLM-generated panels plot responses from CARE, UltraFeedback, and OpenOrca; the human-written panels plot CARE human answers and BLEND gold answers. As shown in Figure~\ref{fig:length_dis-compare}, human answers are highly concise, whereas LLM responses are longer and heavy-tailed.

\section{Implementation Details}
\label{appendix:implementation-hyperparameters}

\paragraph{SFT Tuning.} 
We conduct full-parameter SFT on the instruction data only, using the questions and human-written reference responses in CARE. 
We tune the learning rate in the range $\{1e^{-5}, 2e^{-5}\}$, and train with a batch size of 128 on 4–8 NVIDIA A40 GPUs.

\paragraph{Preference Tuning.} 
We perform full-parameter preference optimization individually for each language using DPO, KTO, and SimPO for 3 epochs until the loss converges. 
Training is done with a batch size of 128 on 4-8 NVIDIA A40 GPUs. 
We tune the learning rate in the range $\{3e^{-7}, 5e^{-7}, 7e^{-7}\}$ and set beta as 0.1. 
The training involves full fine-tuning with 5 warmup steps and employs a linear learning rate scheduler.

\paragraph{MAPO Tuning.} 
We use the same base LLM to be tuned as the rollout model and use \texttt{NLLB-600M-distilled} \cite{costa2022no} to calculate the alignment score as suggested in the original paper \cite{she2024mapo}. We conduct DPO tuning on the generated preference pairs and report the results of the first iteration.

\paragraph{LM Inference.} 
For open-sourced LMs, we run inference on one NVIDIA A40 GPU with the vLLM library\footnote{\url{https://docs.vllm.ai}} \cite{kwon2023efficient}.  We perform decoding by setting the following parameters \{\texttt{temperature=0.7}, \texttt{top\_p=1}\}.  
We limit the context length by setting \{\texttt{max\_model\_len=2048}\}. We also limit the number of generated tokens by the models by setting \{\texttt{max\_tokens=1024}\}. For the closed-source \texttt{GPT-4o} LM, we run inference with Azure OpenAI API.



\paragraph{Baseline preference datasets.}
Table~\ref{tab:baseline_preferencedata_url} lists the URLs of the multilingual adaptations of the general-domain preference datasets used in our experiments (\S\ref{subsec:main-res}). \textbf{HelpSteer3} is natively multilingual, with prompts and responses originally written in various languages and manually annotated preference labels. In contrast, the multilingual versions of \textbf{UltraFeedback} and \textbf{OpenOrca} were generated by automatically translating the English response pairs into the target languages, followed by automatic quality control to remove poorly translated samples. Their preference signals are also synthetic. UltraFeedback relies on GPT-4 to rank the responses, while OpenOrca assumes GPT-4’s responses are always superior to Llama’s when constructing preference pairs ~\cite{wang2025more}.

\begin{table}[t]
\scriptsize             
\setlength{\tabcolsep}{2pt}      
\renewcommand{\arraystretch}{1.05} 
\begin{tabularx}{\linewidth}{c c c X}
\toprule
\textbf{Dataset} & \textbf{Language} & \textbf{Source} & \textbf{HuggingFace Repository} \\ 
\midrule
\multirow{3}{*}{{OpenOrca}}
  & ar & translation & \href{https://huggingface.co/datasets/2A2I/argilla-dpo-mix-7k-arabic}{2A2I/argilla-dpo-mix-7k-arabic} \\
  & ja & translation & \href{https://huggingface.co/datasets/yongtae-jp/orca_dpo_pairs_ja}{yongtae-jp/orca\_dpo\_pairs\_ja} \\
  & zh & translation & \href{https://huggingface.co/datasets/wenbopan/Chinese-dpo-pairs}{wenbopan/Chinese-dpo-pairs} \\
\midrule
\multirow{2}{*}{{UltraFeedback}}
  & ar & translation & \href{https://huggingface.co/datasets/2A2I/argilla-dpo-mix-7k-arabic}{2A2I/argilla-dpo-mix-7k-arabic} \\
  & zh & translation & \href{https://huggingface.co/datasets/wenbopan/Chinese-dpo-pairs}{wenbopan/Chinese-dpo-pairs} \\
\midrule
{HelpSteer3} & zh, ja & native & \href{https://huggingface.co/datasets/nvidia/HelpSteer3}{nvidia/HelpSteer3} \\
\bottomrule
\end{tabularx}
\caption{Multilingual adaptations of baseline preference datasets. ``Source'' indicates whether the non-English data are natively collected or derived via automatic translation.}
\label{tab:baseline_preferencedata_url}
\end{table}


\paragraph{Baselines Prompt Templates.}

The prompt templates used for our prompting-based baselines (\S\ref{subsec:main-res}) are provided in Figure~\ref{fig:roleplay-prompt} (role-play prompting) and Figure~\ref{fig:cot-prompt} (CoT prompting).

\begin{figure}[t]
\centering
\begin{tcolorbox}
[colback=black!5!white,colframe=gray!75!black,title=Role-play Prompt Template]
\scriptsize
\begin{verbatim}
You are a native [Chinese/Arab/Japanese] person, familiar 
with [Chinese/Arab/Japanese] culture and traditions.

{Question}
\end{verbatim}
\end{tcolorbox}
\caption{Prompt template for role-play inference. The LM is told to take on the persona of a native \texttt{[Chinese/Arab/Japanese]} person who is familiar with the culture's traditions, and then asked the culture-specific \texttt{\{Question\}}.}
\label{fig:roleplay-prompt}
\end{figure}

\begin{figure}[t]
\centering
\begin{tcolorbox}
[colback=black!5!white,colframe=gray!75!black,title=CoT Prompt Template]
\scriptsize
\begin{verbatim}
You are a helpful assistant.

{Question} 

Let's think step by step.
\end{verbatim}
\end{tcolorbox}
\caption{Prompt template for CoT inference. We provide the LM with the test question in \texttt{\{Question\}}, then ask it to think step by step when providing the answer.}
\label{fig:cot-prompt}
\end{figure}













\begin{table*}[t]
\centering
\begin{adjustbox}{width=0.9\linewidth}
\begin{tabular}{@{}lcccccccccccc@{}}
\toprule
 &  \multicolumn{4}{c}{\textbf{Chinese}} & \multicolumn{4}{c}{\textbf{Arabic}} & \multicolumn{4}{c}{\textbf{Japanese}} \\ 
 \cmidrule(lr){2-5} 
 \cmidrule(lr){6-9}
 \cmidrule(lr){10-13}
\textbf{Model} & \textit{Sub-nationwide} & \textit{Nationwide} & \textit{Continent-wide} & \textit{Worldwide} & \textit{Sub-nationwide} & \textit{Nationwide} & \textit{Continent-wide} & \textit{Worldwide} & \textit{Sub-nationwide} & \textit{Nationwide} & \textit{Continent-wide} & \textit{Worldwide} \\ \midrule
{\includegraphics[width=0.4cm] {fig/google-logo.png} Gemma2-27B} & 6.50 & 7.16 & \textbf{8.19} & 6.47 & 4.71 & \textbf{6.73} & 4.88 & 6.29 & 2.50 & 5.89 & \textbf{7.20} & -- \\
{\includegraphics[width=0.4cm] {fig/meta-logo.png} Llama3.3-70B} & 6.20 & 6.92 & \textbf{7.68} & 5.57 & 5.14 & \textbf{6.16} & 4.31 & 5.50 & \textbf{6.75} & 5.77 & 6.40 & -- \\
{\includegraphics[width=0.4cm]{fig/qwen-logo.png} Qwen2.5-72B} & 7.70 & 8.49 & \textbf{8.60} & 6.78 & 7.86 & 7.62 & 5.19 & \textbf{8.58} & 4.25 & 6.57 & \textbf{8.00} & --   \\
{\includegraphics[width=0.4cm] {fig/mistral-logo.png} Mistral-Large} & 7.42 & 8.11 & \textbf{8.73} & 6.70 & 5.29 & 7.24 & 5.31 & \textbf{8.08} & 4.50 & \textbf{6.45} & 6.40 & --  \\
{\includegraphics[width=0.4cm] {fig/openai-logo.png} GPT-4o}  & 8.59 & \textbf{8.74} & 8.72 & 7.11 & 7.43 & 8.15 & 7.79 & \textbf{8.92} & 7.00 & \textbf{8.32} & 8.00 & -- \\ 
\bottomrule
\end{tabular}
\end{adjustbox}
\caption{Performance comparison w.r.t. geographic scope. Scores are computed on the entire CARE data, including both local culture and foreign culture.}
\label{tab:geo-results}
\end{table*}

\section{LM-as-a-Judge}
\label{appendix:llm-judge-prompts}

\subsection{Evaluation Prompts}

We instruct \texttt{GPT-4o} as the judge LM to score a model's response to culture-specific questions in CARE. For each cultural category, we provide the judge LM with a detailed evaluation guideline, the culture-specific question, the generated response, and the human reference response, and ask it to score the response on the 1-10 scale. Our evaluation prompt templates are provided in Figure~\ref{fig:judge-prompt-entities-opinion} (Entities \& Opinion), Figure~\ref{fig:judge-prompt-norms-commonsense} (Norms \& Commonsense), and Figure~\ref{fig:judge-prompt-literacy} (Literacy). 

\subsection{Correlation with Human Ratings}

\begin{figure}[t]
    \centering
    \includegraphics[width=\linewidth]{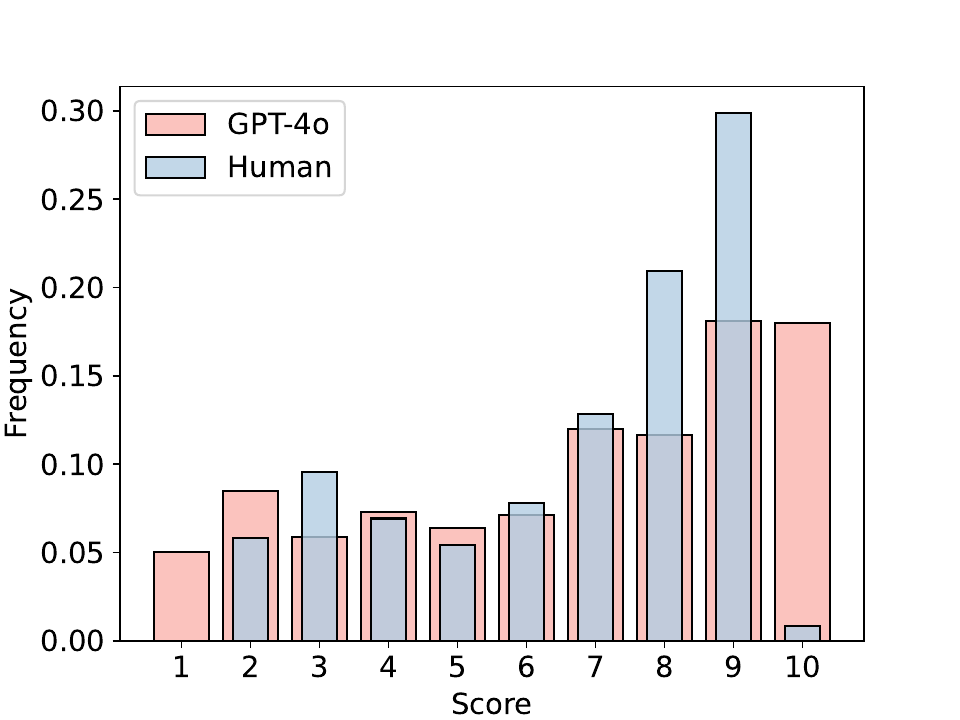}
    \caption{Rating score distributions of the judge LM (\texttt{GPT-4o}) and native human evaluators. The judge LM highly correlates with human ratings.}
    \label{fig:density}
\end{figure}

Fig~\ref{fig:density} shows the rating distribution of \texttt{GPT-4o} and native speakers. We can see a clear correlation between both rating distributions. We also calculate correlation metrics and obtain a Pearson correlation of 0.933, Spearman correlation of 0.901, and Kendall's Tau correlation of 0.733, all indicating high agreement between the judge LM ratings and human ratings. One difference we notice is that humans tend not to assign extreme ratings (1 or 10), yet the LM judge has a higher frequency of those ratings. This is mostly noticeable at the 9-10 rating range where the LM ratings are more equally distributed between a rating of 9 and 10, but the human ratings mostly consist of 9.

\begin{table*}[t]  
  \centering  
  \begin{adjustbox}{width=\linewidth}  
    \scriptsize 
    \setlength{\tabcolsep}{2pt} 
    \begin{tabularx}{\linewidth}{@{}l *{6}{>{\raggedleft\arraybackslash}X} *{6}{>{\raggedleft\arraybackslash}X} @{}}   
      \toprule   
      & \multicolumn{6}{c}{\textbf{Chinese}} & \multicolumn{6}{c}{\textbf{Arabic}} \\   
      \cmidrule(lr){2-7} \cmidrule(l){8-13}    
      \textbf{Approach} & \textit{Entities} & \textit{Opinion} & \textit{Norms} & {\textit{C. sense}} & \textit{Literacy} & \textit{Average} & \textit{Entities} & \textit{Opinion} & \textit{Norms} & {\textit{C. sense}} & \textit{Literacy} & \textit{Average} \\   
      \midrule \\[-10pt] 
      \includegraphics[width=0.3cm] {fig/meta-logo.png} Llama3.1-8B & \colorbox{green4!0}{3.14} & \colorbox{green4!0}{4.16} & \colorbox{green4!0}{4.62} & \colorbox{green4!0}{3.93} & \colorbox{green4!0}{3.03} & \colorbox{green4!0}{3.78} & \colorbox{green4!0}{4.08} & \colorbox{green4!0}{3.62} & \colorbox{green4!0}{2.87} & \colorbox{green4!0}{3.07} & \colorbox{green4!0}{2.07} & \colorbox{green4!0}{3.30} \\
      DPO & \colorbox{green4!10}{+ 0.72} & \colorbox{green4!50}{+ 1.74} & \colorbox{green4!10}{+ 0.74} & \colorbox{green4!50}{+ 1.63} & \colorbox{green4!10}{+ 0.66} & \colorbox{green4!30}{+ 1.10} & \colorbox{green4!0}{+ 0.25} & \colorbox{green4!0}{+ 0.35} & \colorbox{green4!30}{+ 0.83} & \colorbox{green4!50}{+ 1.43} & \colorbox{green4!0}{+ 0.29} & \colorbox{green4!10}{+ 0.56} \\   
      KTO & \colorbox{green4!0}{+ 0.09} & \colorbox{green4!50}{+ 1.54} & \colorbox{green4!10}{+ 0.69} & \colorbox{green4!30}{+ 1.43} & \colorbox{green4!10}{+ 0.83} & \colorbox{green4!10}{+ 0.91} & \colorbox{green4!0}{+ 0.22} & \colorbox{green4!50}{+ 1.61} & \colorbox{green4!30}{+ 1.01} & \colorbox{green4!30}{+ 0.96} & \colorbox{green4!0}{+ 0.03} & \colorbox{green4!10}{+ 0.61}  \\   
      SimPO & \colorbox{green4!10}{+ 0.99} & \colorbox{green4!30}{+ 1.50} & \colorbox{green4!10}{+ 0.70} & \colorbox{green4!30}{+ 1.30} & \colorbox{green4!30}{+ 1.37} & \colorbox{green4!30}{+ 1.16} & \colorbox{green4!0}{+ 0.12} & \colorbox{green4!50}{+ 1.68} & \colorbox{green4!30}{+ 0.95} & \colorbox{green4!30}{+ 1.06} & \colorbox{green4!0}{- 0.07} & \colorbox{green4!10}{+ 0.61}  \\   
      \hdashline[1pt/1pt]
      \noalign{\vskip 0.05cm}
      \includegraphics[width=0.3cm]{fig/qwen-logo.png} Qwen2.5-7B & \colorbox{green4!0}{6.89} & \colorbox{green4!0}{7.86} & \colorbox{green4!0}{7.48} & \colorbox{green4!0}{6.80} & \colorbox{green4!0}{7.37} & \colorbox{green4!0}{7.28} & \colorbox{green4!0}{4.65} & \colorbox{green4!0}{5.84} & \colorbox{green4!0}{5.44} & \colorbox{green4!0}{4.88} & \colorbox{green4!0}{2.84} & \colorbox{green4!0}{4.61} \\
      DPO & \colorbox{green4!30}{+ 0.31} & \colorbox{green4!50}{+ 0.90} & \colorbox{green4!10}{+ 0.18} & \colorbox{green4!0}{+ 0.10} & \colorbox{green4!10}{+ 0.16} & \colorbox{green4!30}{+ 0.33} & \colorbox{green4!0}{- 0.10} & \colorbox{green4!30}{+ 0.56} & \colorbox{green4!0}{+ 0.11} & \colorbox{green4!30}{+ 0.45} & \colorbox{green4!30}{+ 0.51} & \colorbox{green4!30}{+ 0.45} \\   
      KTO & \colorbox{green4!30}{+ 0.54} & \colorbox{green4!50}{+ 0.80} & \colorbox{green4!10}{+ 0.27} & \colorbox{green4!0}{+ 0.13} & \colorbox{green4!0}{- 0.07} & \colorbox{green4!30}{+ 0.33} & \colorbox{green4!0}{- 0.81} & \colorbox{green4!0}{- 0.54} & \colorbox{green4!10}{+ 0.25} & \colorbox{green4!50}{+ 0.83} & \colorbox{green4!50}{+ 0.78} & \colorbox{green4!10}{+ 0.21}  \\   
      SimPO & \colorbox{green4!10}{+ 0.17} & \colorbox{green4!10}{+ 0.54} & \colorbox{green4!0}{+ 0.12} & \colorbox{green4!0}{- 0.17} & \colorbox{green4!0}{+ 0.06} & \colorbox{green4!10}{+ 0.14} & \colorbox{green4!0}{- 0.02} & \colorbox{green4!0}{- 0.04} & \colorbox{green4!0}{+ 0.13} & \colorbox{green4!0}{- 0.16} & \colorbox{green4!10}{+ 0.26} & \colorbox{green4!10}{+ 0.16}  \\   
      \bottomrule    
    \end{tabularx}
  \end{adjustbox}
\caption{Performance comparison w.r.t. different preference learning algorithms on CARE\includegraphics[width=0.5cm]{fig/Hugging.jpg} data. Results show the average score improvements over the vanilla model.}
\label{tab:diff-align-strategy}
\vspace{-.5em}
\end{table*}

\begin{table}[t]
\centering
\begin{adjustbox}{width=\linewidth}
\begin{tabular}{@{}ccccccc@{}}
\toprule
 & \multicolumn{2}{c}{\textbf{Chinese}} & \multicolumn{2}{c}{\textbf{Arabic}} & \multicolumn{2}{c}{\textbf{Japanese}}  \\ 
\cmidrule(lr){2-3} \cmidrule(lr){4-5} \cmidrule(lr){6-7} 
 \textbf{Model} &  {Vanilla} & {Aligned} & {Vanilla} & {Aligned} & {Vanilla} & {Aligned} \\ 
 \midrule
 \textbf{Score} & $6.440 \pm 0.430$ & $6.859 \pm 0.380$ & $5.732 \pm 0.394$ & $6.043 \pm 0.513$ & $5.182 \pm 0.383$ & $5.344 \pm 0.371$ \\
\bottomrule
\end{tabular}
\end{adjustbox}
\caption{Three repeated runs of \texttt{Gemma2-9B} before (Vanilla) and after CARE alignment (Aligned). 
Scores are mean $\pm$ std from LM-as-a-judge evaluation, showing consistent gains after alignment.}
\label{tab:repeated-results}
\end{table}

\begin{table*}[t]
\centering
\begin{adjustbox}{width=0.8\linewidth}
\begin{tabular}{@{}ccccccccc@{}}
\toprule
 & \multicolumn{2}{c}{\textbf{Gemma2-9B}} & \multicolumn{2}{c}{\textbf{Qwen2.5-7B}} & \multicolumn{2}{c}{\textbf{Llama3.1-8B}} & \multicolumn{2}{c}{\textbf{Mistral-7B}}  \\
\cmidrule(lr){2-3} \cmidrule(lr){4-5} \cmidrule(lr){6-7} \cmidrule(lr){8-9} 
 \textbf{Model (w/ preference data)} &  \textit{Arabic} & \textit{Chinese} & \textit{Arabic} & \textit{Chinese} & \textit{Arabic} & \textit{Chinese} & \textit{Arabic} & \textit{Chinese} \\ 
 \midrule
 Vanilla & 43.5 & 53.9 & 39.4 & 55.0 & 28.2 & 37.2 & \textbf{27.1} & 47.1 \\
 DPO (w/ UltraFeedback) & 44.2 & 53.3 & 37.7 & 56.4 & \textbf{30.5} & 38.0 & 26.3 & 39.1 \\
 DPO (w/ HelpSteer3) & -- & 53.1 & -- & \textbf{57.6} & -- & 37.8 & -- & 40.8 \\
 DPO (w/ CARE\includegraphics[width=0.5cm]{fig/Hugging.jpg}) & \textbf{44.6} & \textbf{54.0} & \textbf{39.7} & 56.6 & 30.0 & \textbf{41.4} & 21.0 & \textbf{48.6} \\
\bottomrule
\end{tabular}
\end{adjustbox}
\caption{Out-of-domain evaluation on Blend~\cite{myung2024blend} dataset. For each model family, we compare vanilla with aligned variants using UltraFeedback, HelpSteer3, or CARE, reporting accuracy (\%) on Arabic and Chinese. Helpsteer3 does not offer Arabic data for training.}
\label{tab:Blend-results}
\end{table*}

\subsection{Robustness Analysis via Repeated Runs}

To verify the evaluation reliability, we repeated the inference three times for each prompt on two representative model variants: Gemma2-9B before and after alignment with CARE. The repeated experiments were conducted across all three languages (Arabic, Chinese, and Japanese). The results in Table~\ref{tab:repeated-results} demonstrate that the performance trend observed from the single-run evaluation still holds, confirming its robustness.

\section{Additional Results}
\label{appendix:additional_result}




\subsection{Performance by Preference Learning strategies}
\label{appendix:detail_kto_simpo}

Table~\ref{tab:diff-align-strategy} shows per-category performance when tuned with different preference learning strategies.

\subsection{Performance by Geographic Scope}
\label{appendix:geographic_scope}

Table~\ref{tab:geo-results} shows results from different LMs when stratified by geographic scope. We find that the models struggle the most with sub-nationwide questions. Most models achieve the best scores on continent-wide or worldwide questions.

\subsection{Performance on general NLP tasks}
\label{appendix:general-capabilities}

We then examine whether cultural preference learning impacts the model's overall knowledge and capabilities, using well-established benchmarks for Chinese, Arabic, Japanese, and English: ArabicMMLU \cite{koto2024arabicmmlu}, ChineseMMLU \cite{li2023cmmlu}, the Japanese subset of MMLU-ProX~\cite{xuan2025mmlu}, MMLU \cite{hendrycks2020measuring}, TruthfulQA \cite{lin2021truthfulqa}, and WinoGender \cite{rudinger2018gender}. 
We compare different versions of \texttt{Llama3.1-8B-Instruct}: the vanilla LM, the aligned LM with general human preference from the combined UltraFeedback and OpenOrca datasets, and the culturally aligned LM with cultural human preference from CARE.

The results in Table~\ref{tab:general-capabilities-results} show very small differences between the vanilla LM and its aligned version, on benchmarks in both native language and English, suggesting that cultural preference learning does not hinder the model's overall capabilities.


\subsection{Out-of-Domain Evaluation on Blend}
\label{apendix-ood-blend}

In Figure~\ref{fig:ood-culture-tasks}, we compared models before and after CARE alignment on external cultural benchmarks (Blend, Include). Here, we fix {Blend} as the out-of-domain test set and assess different model variants that are tuned with different preference data. 

As shown in Table~\ref{tab:Blend-results}, CARE-aligned variants are generally competitive or improved relative to vanilla and alternative alignments. The results indicate that a modest amount of carefully curated human-annotated, culture-focused preferences can transfer to out-of-domain tasks—an especially practical point given the scarcity of multilingual human-annotated preference resources (e.g., Open Assistant~\cite{kopf2023openassistant}, HelpSteer3~\cite{wang2025helpsteer3preferenceopenhumanannotatedpreference}, and CARE).

\begin{table*}[t]
\tiny
  \centering
  \begin{adjustbox}{width=\linewidth}
    \begin{tabularx}{\linewidth}{@{}l *{9}{>{\centering\arraybackslash}X}@{}}
      \toprule
       &  \multicolumn{3}{c}{\textbf{Arabic}}
       &  \multicolumn{3}{c}{\textbf{Chinese}}
       &  \multicolumn{3}{c}{\textbf{Japanese}} \\ 
      \cmidrule(lr){2-4} \cmidrule(lr){5-7} \cmidrule(lr){8-10}
      \textbf{Model}
       & \textit{Vanilla} & \textit{DPO (w. Orca+Ultra)} & \textit{DPO (w. CARE)}
       & \textit{Vanilla} & \textit{DPO (w. Orca+Ultra)} & \textit{DPO (w. CARE)}
       & \textit{Vanilla} & \textit{DPO (w. Orca)} & \textit{DPO (w. CARE)} \\
      \midrule
      MMLU (ar/zh/ja)   & 56.4 & 57.1 & 56.5 & 55.5 & 55.5 & 55.2 & 28.8 & 28.4 & 30.2 \\
      MMLU (en)         & 69.1 & 68.8 & 69.1 & 69.1 & 68.9 & 69.1 & 69.1 & 68.8 & 69.0 \\
      TruthfulQA        & 37.8 & 38.7 & 38.6 & 37.8 & 38.8 & 40.8 & 37.8 &	41.1 &	38.7 \\
      WinoGender        & 52.6 & 55.0 & 52.8 & 52.6 & 54.4 & 52.1 & 52.6 &	54.7 &	52.9 \\
      \bottomrule
    \end{tabularx}
  \end{adjustbox}
  \caption{Accuracy of \texttt{Llama3.1-8B-Instruct} variants on general benchmarks assessing overall capabilities in native languages (Chinese, Arabic, and Japanese) and English. Slight differences are observed among the LM variants after applying different preference learning methods.}
  \label{tab:general-capabilities-results}
\end{table*}

\section{Responsible NLP Checklist}
\label{appendix:responsible-checklist}

In our work, we ensure that the use of existing artifacts is consistent with their intended use. 
Specifically, all samples collected for the CARE dataset are sourced from open-sourced datasets, adhering strictly to the original access conditions intended for research purposes. 
We meticulously filter the data to remove any instances of toxic or offensive language and ensure that it does not contain any personally identifiable information. 
For the artifacts we create, we clearly specify their intended use within research contexts, maintaining compatibility with the original access conditions. 
We acknowledge the use of GPT-4 for grammar checking and word polishing.

The annotators for the Chinese and Arabic samples were university-level students fluent in the respective languages. Japanese annotators were Japanese workers with university degrees and cross-cultural experience abroad. Each Chinese and Arabic annotator was paid at \$18 per hour, exceeding the U.S. federal minimum wage. 
All annotators were informed that they were assessing data to be used in a research project focused on ``studying the cultural awareness of various language models''.

\clearpage

\begin{figure*}[htbp]
\centering
\begin{tcolorbox}
[colback=black!5!white,colframe=gray!75!black,title=Cultural entities and opinion evaluation]
\scriptsize
\begin{verbatim}
Please serve as an unbiased evaluator and critically assess the quality of the assistant's response to the user's question 
presented below. When evaluating, focus on the following aspects:

1. **Accuracy**: Is information in the response factually correct and up-to-date?
2. **Depth**: Does the response demonstrate a deep understanding of the topic, providing thorough explanations and context?
3. **Relevance**: Does the response stay focused on the specific question without including unnecessary information?

Begin your evaluation with a brief commentary explaining your judgment on each of these criteria. Aim to be as objective as 
possible in your assessment.

After providing your commentary, assign a numerical rating to the response on a scale from 1 to 10, where:

- **1-2**: Poor — The response is highly inaccurate, lacks detail, contains significant incorrect information, and/or 
includes irrelevant information.

- **3-4**: Below Average — The response is partially accurate, addresses some parts of the question but lacks detail, and 
may include irrelevant information.

- **5-6**: Average — The response is moderately accurate but may contain minor errors, addresses most parts of question with 
adequate detail, and is mostly relevant.

- **7-8**: Good — The response is mostly accurate, addresses all parts of the question with good detail, and is relevant 
with minimal irrelevant information.

- **9-10**: Excellent — The response is highly accurate, provides comprehensive detail, and contains no irrelevant information.

Please format your rating as follows: "Rating: [[number]]". For example: "Rating: [[6]]".

## Question: {question}
## Golden answer: {answer}
## Assistant’s response: {response}
\end{verbatim}
\end{tcolorbox}
\caption{LM-as-a-judge prompt template for cultural entities and opinion questions. }
\label{fig:judge-prompt-entities-opinion}
\end{figure*}

\begin{figure*}[htbp]
\centering
\begin{tcolorbox}
[colback=black!5!white,colframe=gray!75!black,title=Social norms and commonsense evaluation]
\scriptsize
\begin{verbatim}
Please serve as a critical evaluator and rigorously assess the quality of the assistant's response to the user's question 
shown below. When evaluating, prioritize the following stringent criteria:

1. **Accuracy**: Based on the golden answer, is the information factually correct and representative of real-world cultural 
context?
2. **Comprehensiveness**: Does the response provide clear explanations and consider possible nuances or exceptions within 
the cultural context?
3. **Relevance**: Is the response perfectly tailored to the specific cultural context, without any generalizations or 
inaccuracies?

Begin your evaluation with a detailed commentary critically analyzing each of these criteria. Strive to be as objective 
and discerning as possible in your assessment.

After providing your commentary, assign a numerical rating to the response on a scale from 1 to 10, where:

- **1-2**: Poor — The response fails to meet basic expectations for accuracy or relevance, showing major misunderstandings or 
errors.

- **3-4**: Below Average — The response has substantial inaccuracies or omissions, only partially addressing the user's needs.

- **5-6**: Average — The response is fairly accurate and relevant but lacks depth, missing important details or subtleties.

- **7-8**: Good — The response is accurate and covers most aspects well, though it may lack in minor details or perfect 
contextual alignment.

- **9-10**: Excellent — The response is outstanding in all respects; it is precise, detailed, fully relevant, and excellently 
contextualized.

Please format your rating as follows: "Rating: [[number]]". For example: "Rating: [[6]]".

## Question: {question}
## Golden Answer: {answer}
## Assistant’s response: {response}
\end{verbatim}
\end{tcolorbox}
\caption{LM-as-a-judge prompt template for social norms and commonsense questions. }
\label{fig:judge-prompt-norms-commonsense}
\end{figure*}

\begin{figure*}[htbp]
\centering
\begin{tcolorbox}
[colback=black!5!white,colframe=gray!75!black,title=Literacy evaluation]
\scriptsize
\begin{verbatim}
Please serve as a critical evaluator and rigorously assess the quality of the assistant's response to the user's question shown 
below. When evaluating, prioritize the following stringent criteria:

1. **Accuracy**: Is the information in the response factually correct and contextually appropriate?
2. **Interpretation**: Does the response offer insightful and well-supported interpretations of the literary work or topic?
3. **Textual Evidence**: Does the response appropriately reference and analyze specific parts of the text to support its points 
when necessary?
4. **Relevance**: Does the response stay focused on specific question without including unnecessary information?

Begin your evaluation with a detailed commentary critically analyzing each of these criteria. Strive to be as objective and 
discerning as possible in your assessment.

After providing your commentary, assign a numerical rating to the response on a scale from 1 to 10, where:

- **1-2**: Poor — The response fails to meet basic expectations for accuracy or relevance, showing major misunderstandings or 
errors.

- **3-4**: Below Average — The response has substantial inaccuracies or omissions, only partially addressing the user's needs.

- **5-6**: Average — The response is fairly accurate and relevant but lacks depth, missing important details or subtleties.

- **7-8**: Good — The response is accurate and covers most aspects well, though it may lack in minor details or perfect contextual 
alignment.

- **9-10**: Excellent — The response is outstanding in all respects; it is precise, detailed, fully relevant, and excellently 
contextualized.

Please format your rating as follows: "Rating: [[number]]". For example: "Rating: [[6]]".

## Question: {question}
## Reference Answer: {answer}
## Assistant’s response: {response}
\end{verbatim}
\end{tcolorbox}
\caption{LM-as-a-judge prompt template for literacy questions. }
\label{fig:judge-prompt-literacy}
\end{figure*}

\clearpage
\vspace{+10pt}
\noindent\begin{minipage}{\textwidth}
   \small
    \centering
    \fbox{\includegraphics[scale=2, width=\textwidth, page=1, clip, trim=0cm 0cm 0cm 0cm]{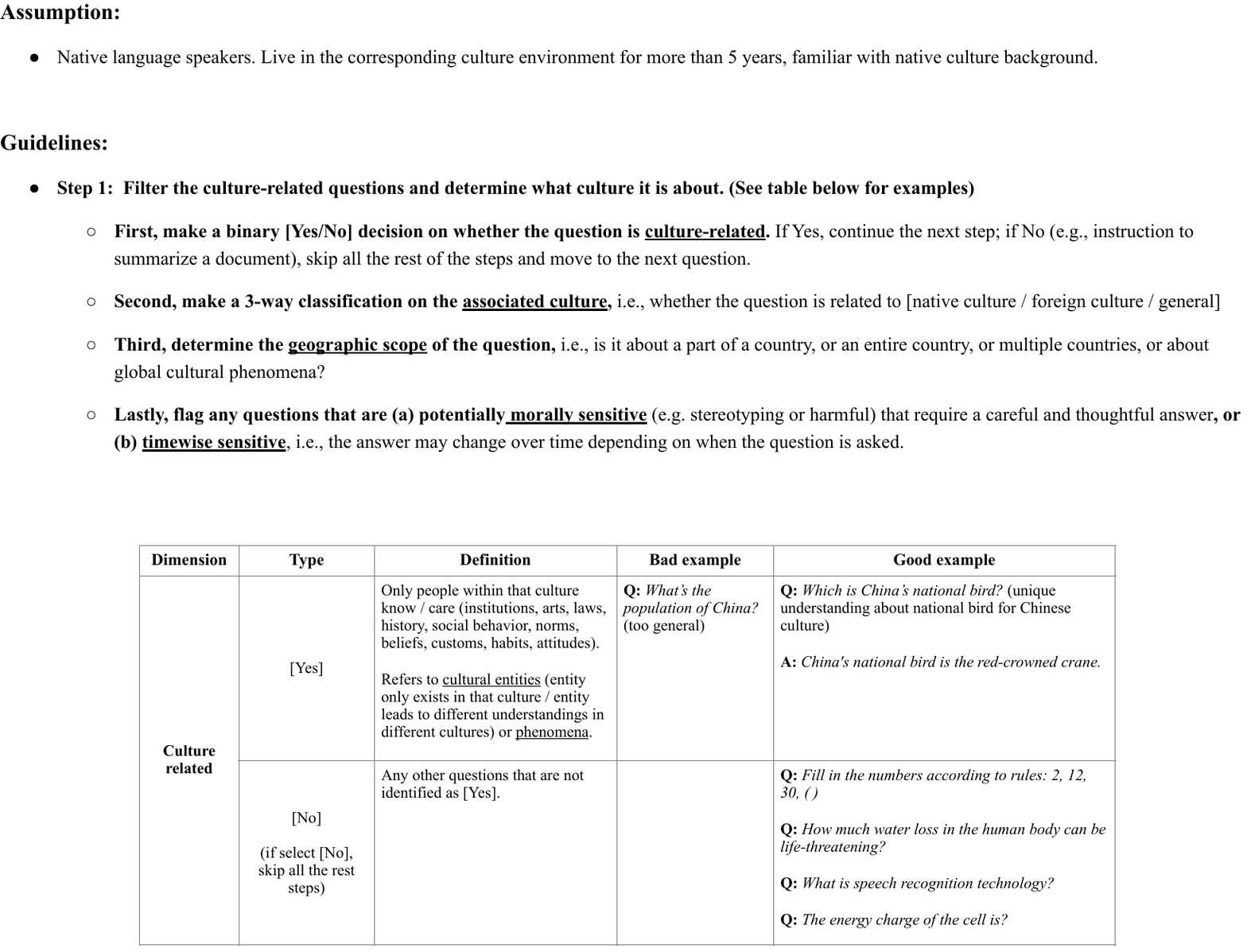}}

    \captionof{figure}{Annotation Guideline (1/5).}
    \label{fig:annotation-guideline-1}
\end{minipage}

\clearpage
\vspace{+10pt}
\noindent\begin{minipage}{\textwidth}
   \small
    \centering
    \fbox{\includegraphics[scale=2, width=\textwidth, page=2, clip, trim=0cm 0cm 0cm 0cm]{fig/annotation_guideline.pdf}}

    \captionof{figure}{Annotation Guideline (2/5).}
\end{minipage}

\clearpage
\vspace{+10pt}
\noindent\begin{minipage}{\textwidth}
   \small
    \centering
    \fbox{\includegraphics[scale=2, width=\textwidth, page=3, clip, trim=0cm 0cm 0cm 0cm]{fig/annotation_guideline.pdf}}

    \captionof{figure}{Annotation Guideline (3/5).}
\end{minipage}

\clearpage
\vspace{+10pt}
\noindent\begin{minipage}{\textwidth}
   \small
    \centering
    \fbox{\includegraphics[scale=2, width=\textwidth, page=4, clip, trim=0cm 0cm 0cm 0cm]{fig/annotation_guideline.pdf}}

    \captionof{figure}{Annotation Guideline (4/5).}
\end{minipage}

\clearpage
\vspace{+10pt}
\noindent\begin{minipage}{\textwidth}
   \small
    \centering
    \fbox{\includegraphics[scale=2, width=\textwidth, page=5, clip, trim=0cm 0cm 0cm 0cm]{fig/annotation_guideline.pdf}}

    \captionof{figure}{Annotation Guideline (5/5).}
    \label{fig:annotation-guideline-5}
\end{minipage}

\end{document}